\documentclass[MS]{macro/neu_msthesis}

\usepackage{amsfonts}			
\usepackage{times}		

\providecommand{\ifno}[1]{}

\usepackage{multirow}
\makeindex
\usepackage{makeidx}
\usepackage{acronym}

\usepackage{url}
\usepackage{graphicx}

\ifnum\pdfoutput>0
\usepackage[pdftex,
bookmarks=true,
bookmarksnumbered=true,
hypertexnames=false,
breaklinks=true           
]{hyperref}
\else
\usepackage[hypertex,
bookmarks=true,
bookmarksnumbered=true,
hypertexnames=false,
breaklinks=true           
]{hyperref}
\fi

\hypersetup{				
pdfauthor = {\authorRef},
pdftitle = {\titleRef},
pdfsubject = {\expandafter{\degreeRef} thesis submitted to Northeastern University},
pdfkeywords = {Multimodal robots,M4,Traversability,3D path planning}
pdfcreator = {LaTeX with hyperref package},
pdfproducer = {dvips + ps2pdf}}

\usepackage[style=ieee,dashed=false,isbn=false,url=false,doi=false]{biblatex}
\DeclareSourcemap{
  \maps{
    \map{
      \pertype{article}
      \step[fieldset=language, null]
      \step[fieldset=url, null]
      \step[fieldset=doi, null]
      \step[fieldset=issn, null]
      \step[fieldset=isbn, null]
      \step[fieldset=note, null]
      \step[fieldset=editor, null]
      \step[fieldset=urldate, null]
      \step[fieldset=file, null]
    }
  }
}

\DeclareSourcemap{
  \maps{
    \map{
      \pertype{inproceedings}
      \step[fieldset=language, null]
      \step[fieldset=url, null]
      \step[fieldset=doi, null]
      \step[fieldset=issn, null]
      \step[fieldset=isbn, null]
      \step[fieldset=note, null]
      \step[fieldset=editor, null]
      \step[fieldset=urldate, null]
      \step[fieldset=file, null]
    }
  }
}
\usepackage{physics}
\usepackage{siunitx}
\usepackage{graphics} 
\usepackage{graphicx} 
\usepackage{epsfig} 
\usepackage{lipsum} 
\usepackage{amsmath, bm} 
\usepackage{amssymb}  
\usepackage{upgreek}
\usepackage{url}
\usepackage{breakurl}
\usepackage{caption}
\usepackage{subcaption}
\usepackage{pgfplots}
\usepackage{floatrow}
\usepackage{float}
\usepackage{multicol}
\usepackage{tikz}
\usepackage{booktabs}
\usepackage{longtable}
\usepackage{algorithm}
\usepackage{algpseudocode}
\usepackage{placeins}
\usepackage{placeins}
\usepackage[htt]{hyphenat}
\usetikzlibrary{shapes, arrows}
\usepackage{amsmath}
\newfloatcommand{capbtabbox}{table}[][\FBwidth]

\title{Contact-Implicit Modeling and Simulation of a Snake Robot on Compliant and Granular Terrain}

\author{Haroon Hublikar}

\dept{Electrical and Computer Engineering}

\degree{Master of Science}

\degreename{Electrical and Computer Engineering}

\field{Electrical and Computer Engineering}

\submitdate{December 2025}

\numberofmembers{2}

\principaladviser{Dr. Alireza Ramezani}
\firstreader{Dr.xxx}

\chairman{Dr. xxx}

\dean{Dr. xxx}

\RequirePackage{filecontents}

\begin{document}

\pdfbookmark[1]{Cover}{cover}

\titlepage

\begin{frontmatter}

\pdfbookmark[1]{Table of Contents}{contents}
\tableofcontents
\listoffigures
\newpage\ssp
\listoftables

\chapter*{List of Acronyms}
\addcontentsline{toc}{chapter}{List of Acronyms}
\markboth{List of Acronyms}{List of Acronyms}

\begin{acronym}
\acro{COBRA}{Crater Observing Bio-inspired Rolling Articulator}.
	A snake-like articulated robotic platform designed for flexible locomotion and object interaction in complex environments.

\acro{GJK}{Gilbert-Johnson-Keerthi}.
	A geometric algorithm used for computing the shortest distance between convex shapes, commonly applied in collision detection and contact estimation.

\acro{IMU}{Inertial Measurement Unit}.
	A sensor module that measures acceleration and angular velocity, typically used for motion tracking and orientation estimation in robotics.

\acro{KD}{k-Dimensional}.
	Refers to data structures such as KD-trees that organize points in k-dimensional space for efficient nearest-neighbor and spatial queries.

\acro{RGB-D}{Red-Green-Blue and Depth}.
	Refers to combined visual and depth data captured by sensors like Intel RealSense, enabling perception of 3D structure and color.

\acro{SLAM}{Simultaneous Localization and Mapping}.
	A technique that allows a robot to build a map of an unknown environment while simultaneously tracking its own location within that map.

\acro{STL}{Standard Tessellation Language}.
	A widely used file format for 3D models that describes surface geometry using triangular facets, common in 3D printing and CAD.

\acro{MPC}{Model Predictive Control}.
	An optimization-based control strategy that predicts future system behavior over a finite horizon and computes optimal control actions by solving a constrained optimization problem at each time step.

\acro{VIO}{Visual-Inertial Odometry}.
	A state estimation technique that fuses visual features tracked across camera frames with inertial measurements to provide robust pose estimation, particularly effective when GPS is unavailable.

\acro{SCM}{Soil Contact Model}.
	Chrono's physics-based deformable terrain model that simulates particle-body interactions using pressure-sinkage relationships and Janosi-Hanamoto shear stress formulation for granular media locomotion.

\acro{ODE}{Ordinary Differential Equations}.
	Mathematical equations describing continuous-time system dynamics where state derivatives depend on current states and inputs, fundamental to modeling rigid-body robot dynamics.

\acro{DAE}{Differential Algebraic Equations}.
	A system combining differential equations with algebraic constraints, essential for modeling mechanical systems with contact forces and kinematic constraints in robotics.

\acro{LiPo}{Lithium Polymer}.
	Rechargeable battery technology using polymer electrolytes, offering high energy density and flexible form factors ideal for mobile robotics applications.

\acro{DEM}{Discrete Element Method}.
    A numerical technique used to model and simulate the motion and interaction of large numbers of particles, commonly applied to granular media, soil mechanics, and contact-rich environments in robotics simulations (e.g., Chrono DEM).
    
\acro{CUDA}{Compute Unified Device Architecture}.
    A parallel computing platform and programming model developed by NVIDIA, enabling general-purpose computation on GPUs for high-performance scientific simulation, robotics, and real-time perception workloads.
    
\acro{URDF}{Unified Robot Description Format}.
    An XML-based representation used to describe the kinematic structure, physical properties, visuals, and collision geometry of robots for simulation and control in frameworks such as ROS and Chrono.

\acro{VSG}{Vulkan Scene Graph}.
    A high-performance graphics rendering framework based on the Vulkan API, used in Chrono for efficient real-time terrain deformation and robot visualization.

\acro{CPU}{Central Processing Unit}.
    The primary compute engine responsible for executing general-purpose instructions in robotic systems.

\acro{GPU}{Graphics Processing Unit}.
    A parallel compute architecture optimized for high-throughput operations such as contact resolution and terrain simulation.

\end{acronym}


\begin{acknowledgements}

I would like to express my heartfelt gratitude to Prof. Alireza Ramezani, my thesis advisor, for his exceptional mentorship, continuous encouragement, and insightful guidance throughout my Master’s research journey. His expertise has shaped my approach to scientific thinking and has been instrumental in the successful completion of this thesis. I would also like to thank all members of the Silicon Synapse Lab at Northeastern University for fostering a supportive and innovative research environment that made this journey both productive and enjoyable. I am especially grateful to Adarsh Salagame, who has been a tremendous source of guidance as both a collaborator and mentor. His leadership, technical contributions, and continuous support have been central to my growth as a researcher. I have learned a great deal from working alongside him, and I truly value the time and effort he has invested in helping me succeed. I extend my sincere appreciation to my thesis committee members, Prof. Seth Hutchinson and Prof. Derya Aksaray, for their constructive feedback and thoughtful discussions during the defense process. Their perspectives greatly strengthened the technical quality and clarity of this work. Finally, I am deeply grateful to my parents and family for their unconditional love, patience, and constant encouragement. Their belief in me has been my greatest source of motivation. This Thesis marks an important milestone in my academic and professional path, and I look forward to contributing further to the field of robotics.

\end{acknowledgements}


\begin{abstract}

This thesis presents a unified modeling and simulation framework for analyzing
sidewinding and tumbling locomotion of the COBRA snake robot across rigid,
compliant, and granular terrains. A contact-implicit formulation is used to
model distributed frictional interactions during sidewinding, and validated
through MATLAB Simscape simulations and physical experiments on rigid ground 
and loose sand. To capture terrain deformation effects, Project Chrono’s
Soil Contact Model (SCM) is integrated with the articulated multibody dynamics,
enabling prediction of slip, sinkage, and load redistribution that reduce
stride efficiency on deformable substrates. For high-energy rolling locomotion on steep slopes, the Chrono DEM Engine is
used to simulate particle-resolved granular interactions, revealing soil failure,
intermittent lift-off, and energy dissipation mechanisms not captured by rigid
models. Together, these methods span real-time control-oriented simulation and
high-fidelity granular physics. Results demonstrate that rigid-ground models provide accurate short-horizon
motion prediction, while continuum and particle-based terrain modeling becomes
necessary for reliable mobility analysis in soft and highly dynamic environments.
This work establishes a hierarchical simulation pipeline that advances robust,
terrain-aware locomotion for robots operating in challenging unstructured
settings.
\end{abstract}

\end{frontmatter}


\pagestyle{headings}


\chapter{Introduction}
\label{chap:intro}

\section{Background and Motivation}

Locomotion in contact-rich environments remains a central challenge in modern robotics. 
Unlike wheeled or legged systems with discrete and typically well-structured contact events, 
snake robots generate propulsion through a continuous distribution of frictional interactions 
that vary along the body. These dense and highly coupled contacts make modeling and control 
nontrivial, yet they enable terrain-adaptive motion and access to confined spaces that are 
infeasible for more conventional platforms. As a result, snake robots have demonstrated 
significant potential in applications such as search and rescue \cite{whitman_snake_2018}, 
industrial inspection, medical procedures \cite{shin_autonomous_2019}, and planetary exploration 
\cite{liu_review_2021}.

Beyond advancing limbless robot design, the study of snake locomotion also offers insights 
relevant to dexterous manipulation. Contact-rich locomotion and multi-fingered manipulation 
share a common requirement: coordinating many simultaneous or intermittent contacts while 
respecting nonpenetration and frictional constraints 
\cite{jiang_contact-aware_2023,chavan-dafle_stable_2018,xiao_one-finger_2024}. 
This duality has motivated research into unified, optimization-based frameworks capable of 
reasoning over large sets of possible contact states without predefined sequencing or timing.

However, most classical analyses of snake locomotion assume rigid terrain and Coulomb friction 
\cite{tesch_parameterized_2009,rollinson_virtual_2012,wiriyacharoensunthorn_analysis_2002}. 
In real-world environments such as sand, soil, mud, or snow, the ground is deformable, compliant, 
and often particulate. In these conditions, terrain deformation and body motion are tightly coupled, 
leading to effects such as sinkage, slip, shear displacement, and resistive drag that rigid ground 
models fail to capture. Accurate modeling of this coupling is essential for enabling reliable 
locomotion in unstructured outdoor and exploratory environments where deformability dominates 
\cite{yoo_traversability-aware_2024,arachchige_tumbling_2024}.

\begin{figure}[ht]
    \centering
    \includegraphics[width=\textwidth]{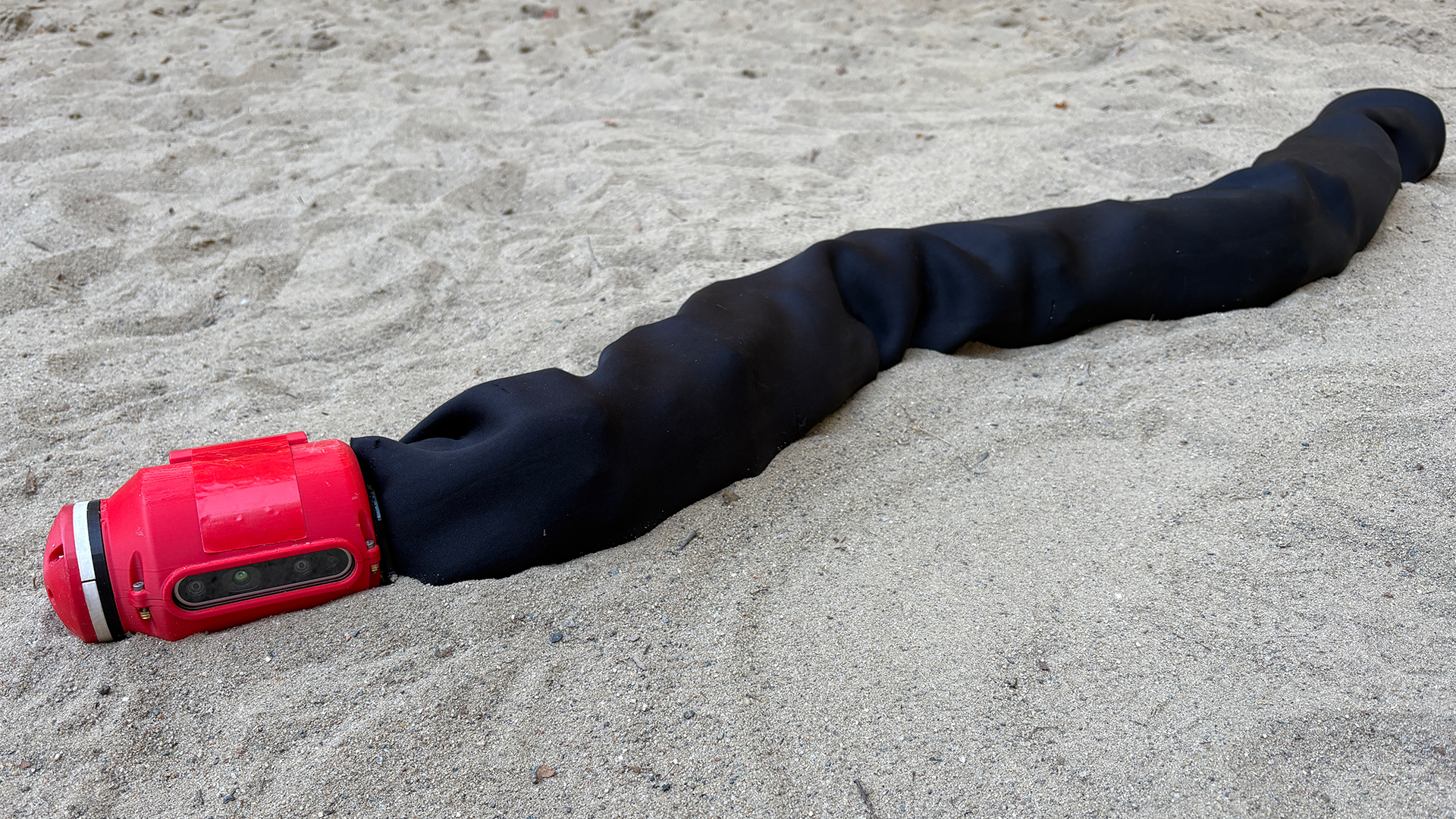}
    \caption{COBRA robot performing sidewinding motion across loose sandy terrain.} 
    \label{fig:cobra_on_sand}
\end{figure}

The physics of granular media occupy a regime between solid and fluid mechanics, where
particle contacts form and break dynamically, producing nonlinear resistance, hysteresis,
and anisotropic force transmission. Modeling this behavior in a computationally tractable
manner requires balancing physical realism with numerical stability. Compliant-contact
formulations address this need by replacing hard complementarity constraints with
spring–damper reactions. These approaches retain differentiability for optimization-based
solvers while capturing first-order deformation effects such as penetration depth, shear
lag, and energy dissipation, making them suitable for both motion analysis and control.

Recent advances in contact-implicit trajectory optimization (CITO) have demonstrated that
contact dynamics can be embedded directly into the optimization problem 
\cite{kim_contact-implicit_2024,wang_contact-implicit_2023,huang_adaptive_2024}. Rather
than representing contact as discrete mode transitions, CITO formulates forces and
kinematics as continuous variables that evolve within constraints representing frictional,
unilateral interaction.

Building on these ideas, our earlier IEEE RA-L 2026 paper (under review), titled
``Contact-Implicit Modeling of Snake Robot Sidewinding on Compliant and Granular Terrain'',
extended Moreau’s time-stepping differential inclusion framework \cite{moreau_unilateral_1988}
to incorporate compliant surface interactions. In that work, normal and tangential contact
forces were modeled using spring–damper relationships, enabling limited ground indentation
(negative gap) and velocity-dependent damping within the contact manifold. This compliant
extension supported simulation of sidewinding propulsion on deformable substrates and
provided a unified structure for modeling locomotion across rigid, compliant, and granular
terrains. Hardware experiments performed with the COBRA modular snake robot confirmed that
even simplified compliant-ground models can recover key locomotion characteristics on sand,
including phase-locked traction patterns and propagation of traveling-wave kinematics.

This thesis investigates two distinct locomotion modes demonstrated by the
\ac{COBRA} robot, each requiring a different terrain modeling approach based on
the level of ground deformation involved:

\begin{itemize}
    \item \textbf{Sidewinding Gait:} analyzed using Simscape Multibody for
    rigid-ground behavior and Chrono \ac{SCM} for compliant surface interaction,
    complemented by hardware experiments performed on both flat laboratory
    flooring and loose sand.

    \item \textbf{Tumbling Gait:} investigated using a MATLAB Simulink rigid-ground formulation together with particle-resolved granular simulations in the Chrono \ac{DEM} Engine, capturing the high-energy, deep-substrate interactions characteristic of deformable particulate terrain.
\end{itemize}

While the compliant model captured critical qualitative behaviors such as sinkage and
reduced stride length, it did not fully represent the distributed deformation and
soil-pressure response modeled by Project Chrono’s Soil Contact Model (SCM). In earlier
work, Chrono-based simulations were used primarily for validation, but the detailed
simulation architecture, solver configuration, contact mechanics, and soil-parameter
choices were not described in depth. This thesis closes that gap by presenting a
complete account of the Chrono SCM setup, including soil formulation, the simulation
pipeline, and the associated data-processing methods, in order to provide a
higher-fidelity continuum-based terrain representation that supports locomotion modeling.

Beyond continuum-style compliance modeling, this thesis incorporates granular
simulation using the Chrono DEM Engine (DEME). While SCM represents soft soil through
a deformable surface mesh and empirical pressure sinkage laws, DEME resolves the
terrain as a collection of discrete particles whose collisions generate frictional and
normal reactions at the grain scale. This makes it possible to analyze sliding,
toppling, and mass rearrangement during high-energy behaviors such as tumbling.

Together, SCM and DEME establish a dual-scale modeling strategy: SCM provides a
computationally efficient approximation for steady sidewinding on deformable
ground, whereas DEME offers high-fidelity evaluation of granular mechanics
critical to tumbling stability and downslope behavior. This combined framework
supports a comprehensive investigation of how substrate mechanics affect
propulsion, contact distribution, and locomotion robustness across both gait
modalities, forming a core foundation for the analyses presented in this thesis.

\section{Overview of the COBRA Platform}

The robotic system studied in this thesis is the \textit{Crater Observing Bioinspired Rolling Articulator (COBRA)}, a modular snake robot composed of twelve interconnected segments, including eleven actuated joints and a head payload module. The modules are arranged in an alternating yaw–pitch sequence, enabling the robot to generate coupled lateral and vertical curvatures along its 1.7\,m structure. This kinematic configuration supports traveling-wave motions that underlie sidewinding and other gaits on rigid and deformable terrains \cite{salagame_crater_2025}.

\subsection{Body Module Design}
Each body module houses a Dynamixel XH540-W270-R servo motor enclosed within a Markforged Onyx composite housing that provides high stiffness and heat resistance. A PLA outer shell encloses the local electronics and facilitates cable routing. The module includes thrust bearings, washers, and a standardized joint-mating connector for mechanical continuity between adjacent links. Power is supplied by a 4S 850\,mAh 75C Li-Po battery regulated through an LM2596 buck converter providing 14\,V to each servo. The power and data lines run along an RS-485 bus, maintaining low-latency actuation and synchronized communication across all modules.

\begin{figure}[H]
    \centering
    \includegraphics[width=\textwidth]{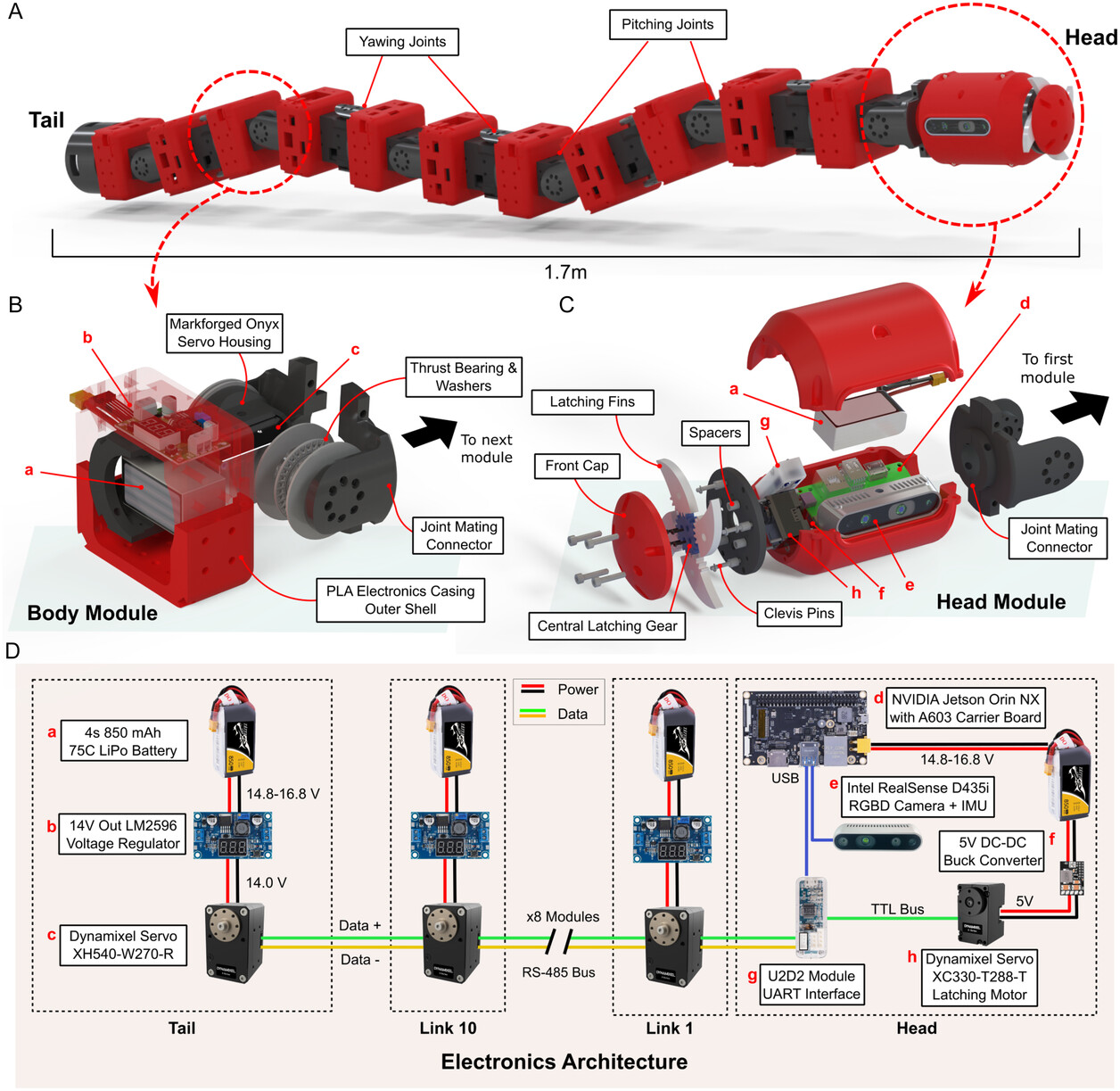}
    \caption{Overview of the COBRA hardware platform. 
    (\textbf{A}) Full-body configuration showing alternating yawing and pitching joints across a 1.7\,m serial chain. 
    (\textbf{B}) Body module assembly including the Dynamixel XH540-W270-R servo, voltage regulator, and battery interface. 
    (\textbf{C}) Exploded view of the head module illustrating the Jetson Orin NX computer, RealSense D435i RGB-D camera, latching mechanism, and electronics layout. 
    (\textbf{D}) System electronics architecture with distributed power and data buses along the RS-485 daisy chain. \textit{Image courtesy of Adarsh Salagame}. \cite{salagame_crater_2025}}
    \label{fig:cobra_hardware}
\end{figure}

\subsection{Head Module and Onboard Computation}
The head module integrates computation, sensing, and an active latching mechanism. An NVIDIA Jetson Orin NX with an A603 carrier board executes high-level perception, planning, and control processes. It interfaces with the RS-485 network via a U2D2 USB-to-TTL interface and receives visual–inertial data from an Intel RealSense D435i camera, which provides RGB-D imagery and IMU readings for 6-DoF pose estimation and environment mapping. Power is distributed through a 5\,V DC–DC buck converter and monitored across both computation and sensing subsystems.

\subsection{Latching Mechanism}
The head incorporates a compact mechanical latching system actuated by a Dynamixel XC330-T288-T servo. The mechanism employs a central latching gear that synchronously drives four latching fins, allowing the head to securely connect or release from other modules or environmental fixtures. This functionality enables modular reconfiguration, anchoring, and loco-manipulation behaviors, such as transitioning between sidewinding and rolling (tumbling) gaits \cite{salagame_crater_2025}.

\subsection{Electronics Architecture}
The distributed electronics system connects all servos and sensors through a unified RS-485 communication bus (data) and a parallel 14\,V power line (power). As shown in Fig.~\ref{fig:cobra_hardware}D, power is provided by a Li-Po source at the tail and regulated locally at each module. The architecture supports real-time operation at control frequencies up to 500\,Hz, with feedback from servo encoders (position, velocity, and torque) and inertial sensors. This configuration enables precise gait control, real-time state estimation, and robust outdoor operation without reliance on external motion-capture systems.

\subsection{Summary}
In summary, the COBRA platform combines modular mechanical design, high-bandwidth distributed control, and integrated perception into a single architecture suitable for both terrestrial and extraterrestrial locomotion studies. Its combination of mechanical adaptability and computational autonomy makes it an effective testbed for developing and validating contact-implicit control and perception-driven planning algorithms on deformable terrains.

\section{Related Work}

The COBRA platform has undergone several generations of research that progressively expanded its capabilities in locomotion, modeling, perception, and terrain interaction. Early work demonstrated open-loop gait generation using coupled sinusoidal patterns across its alternating yaw–pitch joint structure, enabling sidewinding and lateral undulation on flat and moderately uneven terrain \cite{rollinson_virtual_2012,tesch_parameterized_2009,huang_unified_2024}. These studies showed that continuous traveling-wave motions can be produced reliably along a serially actuated body, highlighting the role of phase coordination, distributed contacts, and curvature modulation in achieving stable propulsion on rigid surfaces.

Further investigation extended COBRA to compliant surfaces and sloped environments. Experiments on loose sand and inclined terrain showed that distributed body contacts enable propulsion even when parts of the robot lose support, similar to biological sidewinders \cite{astley_modulation_2015,marvi_sidewinding_2014}. These results emphasized how contact placement and load distribution influence normal forces, revealing that selective lifting of body segments improves traction and reduces slip on soft ground. This motivated the development of modeling frameworks capable of capturing the interplay between body curvature, contact distribution, and terrain yielding.

Research on snake robots more broadly has continued to expand in locomotion modeling, gait synthesis, and control. Foundational work established serpenoid curvature parametrization and coordinated phase-gait generation for hyper-redundant systems \cite{wiriyacharoensunthorn_analysis_2002,wang_dynamic_2009,yang_hierarchical_2012}. These principles were later adapted to obstacle-rich environments and manipulation tasks through methods that leveraged frictional interactions, redundancy resolution, and shape control \cite{reyes_studying_2017,reyes_planar_2014,reyes_using_2015,tanaka_cooperative_2006}. Rhythmic controllers, hybrid CPG–MPC approaches, and learning-based strategies further improved adaptation to uncertainty and environmental variation \cite{nor_cpg-based_2014,chen_bioinspired_2021,yan_bionic_2022,zhu_path_2022,tanev_automated_2005,wang_path_2021,zanon_safe_2021,shin_autonomous_2019}.

Beyond snake robots, substantial work has focused on contact modeling, deformable terrain interaction, and motion planning under uncertainty. Proprioceptive contact estimation has shown strong performance in identifying multi-contact forces \cite{vorndamme_collision_2017}. Efficient collision-geometry computation, including GJK and KD-tree search, supports fast distance queries for articulated bodies \cite{gilbert_fast_1988,bentley_multidimensional_1975,bucki_rectangular_2020}. Research on granular locomotion highlights the effects of yielding terrain on propulsion, with particle-scale DEM studies \cite{maurel_chronodeme_2023,sunday_validation_2019}, tumbling robots on loose ground \cite{arachchige_tumbling_2024}, and planning algorithms aware of ground uncertainty \cite{yoo_traversability-aware_2024}. Perception-driven mobility has advanced through visual–inertial SLAM and onboard estimation frameworks for cluttered or GPS-denied environments \cite{zhang_laservisualinertial_2018,sun_robust_2018,mohta_experiments_2018}. Terramechanics studies introduced continuum-based models such as SCM \cite{krenn_scm_2009,krenn_soft_2011,serban_real-time_2023,buse_scm_2016} while DEM research emphasized grain-scale forces and mass wasting \cite{preclik_ultrascale_2015,tavarez_discrete_2007}. Finally, theoretical progress in non-smooth mechanics and contact time-stepping formed the basis of modern simulation systems used in robotics \cite{moreau_unilateral_1988,studer_numerics_2009,hwangbo_per-contact_2018,castro_unconstrained_2023}.

Work from the Silicon Synapse Lab builds directly on these foundations and extends them into multi-modal, contact-rich locomotion across legged, aerial, and serpentine platforms. Husky Carbon demonstrated posture manipulation and thrust-vectoring control for steep-slope and narrow-path locomotion \cite{krishnamurthy_enabling_2024,wang_quadratic_2025,krishnamurthy_thruster-assisted_2024,krishnamurthy_optimization_2024}. QP-based force regulation and stability controllers further strengthened locomotion robustness \cite{pitroda_capture_2024,pitroda_quadratic_2024,pitroda_enhanced_2024,pitroda_conjugate_2024}. Parallel work on the Harpy platform explored thruster-assisted bipedal locomotion, reduced-order models, and real-time momentum-based disturbance estimation \cite{dangol_control_2021,gupta_conjugate_2024,dangol_thruster-assisted_2020}. 

Aerial robotics research in the Lab introduced Aerobat, a high-DOF morphing-wing vehicle capable of navigating confined environments through unsteady aerodynamic interactions \cite{ramezani_aerobat_2022,sihite_wake-based_2022,sihite_unsteady_2022}. This line of work expanded into neural and physics-informed aerodynamic models, contributing insights into maneuverability in dynamic morphing-wing flight \cite{ghanem_efficient_2021,gupta_banking_2024,gupta_modeling_2024,gupta_estimation_2025}. 

The Lab also advanced articulated and serpentine systems. Vision-guided loco-manipulation and large-scale navigation were demonstrated on snake robots operating in cluttered terrain \cite{salagame_vision-guided_2025,jiang_hierarchical_2024}. Contact-aware optimization frameworks were introduced for non-impulsive loco-manipulation \cite{salagame_non-impulsive_2024,salagame_loco-manipulation_2024}. Learning-based methods improved terrain prediction and dynamic parameter identification \cite{salagame_reinforcement_2024,gherold_self-supervised_2024,ghanem_learning_2024}. COBRA-specific research examined granular physics, tumbling stability, and high-energy ground interaction \cite{salagame_crater_nodate,salagame_validation_2024}. Additional work explored dynamic modeling and control for multi-modal planetary and transformable robots \cite{mandralis_atmo_2025,wang_dynamic_2025,wang_guiding_2025,wang_thruster-enhanced_2025}.

Together, these efforts from the wider research community and the SS Lab form a cohesive body of knowledge that informs the modeling, perception, and control methods used in this thesis.

\begin{figure}[ht]
    \centering
    \includegraphics[width=\textwidth]{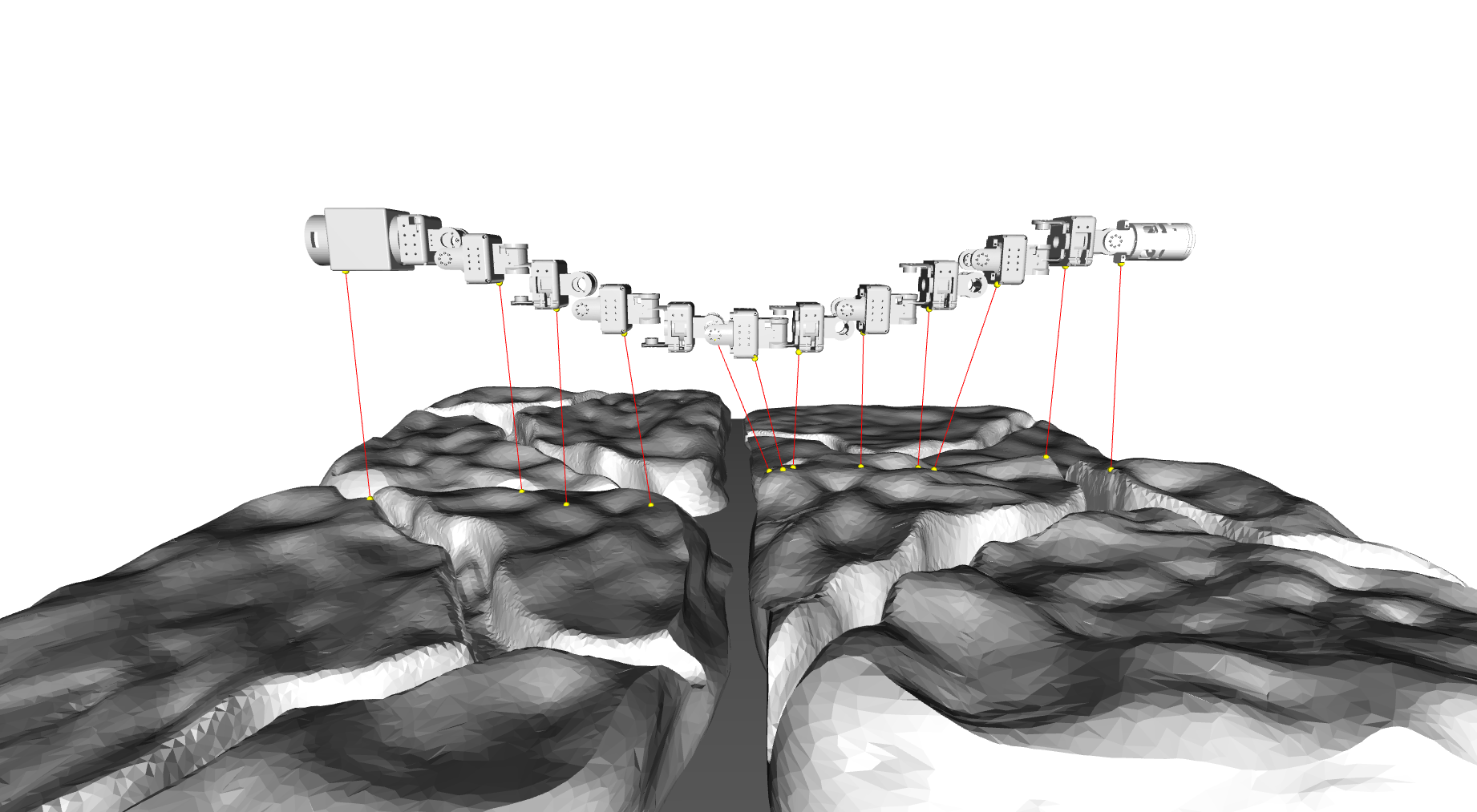}
    \caption{Gap function computation from the Master's Project (April 2025), showing projection lines from each COBRA segment to the terrain surface for real-time distance and contact estimation.}
    \label{fig:cobra_on_sand}
\end{figure}

Finally, COBRA has also been used as a perception-enabled loco-manipulation testbed through the Master’s Project titled \textit{Real-Time Gap Function Computation, Contact Estimation, and Object Pose Estimation for Snake Robots in Loco-Manipulation Tasks}. This system integrated FoundationPose-based tracking, SAM2 segmentation, and a geometry-based contact estimator relying on GJK \cite{gilbert_fast_1988} and KD-tree search \cite{bentley_multidimensional_1975}. Running onboard a Jetson Orin NX with visual–inertial sensing, this framework enabled real-time contact estimation that supports future contact-implicit planning.

Unlike earlier work that emphasized gait synthesis, perception, or hardware validation, this thesis focuses primarily on the modeling and simulation of snake-robot locomotion across deformable terrains. Specifically, it advances:
\begin{itemize}
    \item The contact-implicit modeling framework for snake locomotion introduced in the RA-L 2026 study,
    \item The implementation and parameterization of SCM-based continuum simulations in Project Chrono, and
    \item The integration of particle-resolved DEME simulations for high-fidelity terrain interaction analysis.
\end{itemize}

Together, these components establish a unified computational framework for studying, validating, and extending contact-rich sidewinding locomotion across rigid, compliant, and granular environments.

\section{Thesis Contributions and Outline}

This thesis consolidates and deepens work on deformable-terrain locomotion for modular
snake robots using the COBRA platform. While the core mathematical formulation of the
contact-implicit locomotion model in our \textit{IEEE RA-L} submission was developed by
Adarsh Salagame in collaboration with Professor Alireza Ramezani, my primary contributions
lie in (i) the creation of high-fidelity simulation environments in Project Chrono for
compliant and granular terrain, (ii) extensive hardware data collection for validation, and
(iii) integration of these components into a unified evaluation framework. These enable
direct comparison of analytical modeling assumptions against physically grounded simulation
and real-world performance.

\subsection{Primary Contributions}

\begin{enumerate}
    \item \textbf{Project Chrono Simulation Software Development.}
    I built the Chrono-based simulation environment from the ground up, including custom
    configuration for GPU execution, collision model setup, and a modular simulation loop for
    loading the COBRA URDF, applying gait commands, resolving terrain interaction, and logging
    high-frequency state and contact data.

    \item \textbf{Integration and Calibration of Chrono’s Soil Contact Model (SCM).}
    I incorporated SCM terrain into the COBRA simulation framework, tuned soil parameters
    using hardware sand experiments, and generated the compliant-surface datasets used for
    model comparison in the RA-L study.

    \item \textbf{Simulation of Tumbling Locomotion Using Chrono DEME (AISJ 2025).}
    I implemented the DEM-based tumbling simulation used in the AISJ study, configuring
    granular terrain and slope settings to analyze particle-level resistance forces, sinkage,
    and flow behavior during multimodal locomotion.

    \item \textbf{Hardware Data Collection and Simulation Validation.}
    I led the collection of experimental datasets used for validating the simulation models,
    including RGB-D + IMU data from COBRA on loose sand and motion-capture data on rigid
    ground. I also produced Chrono-based simulation datasets that were used to evaluate ground
    stiffness, contact patterns, and locomotion performance.
\end{enumerate}

Together, these contributions provide the practical computational and experimental framework
required to evaluate deformable-surface locomotion, forming the core foundation of this thesis.

\subsection{Thesis Outline}

The remainder of this thesis is structured as follows:

\begin{itemize}
    \item \textbf{Chapter~1: Introduction}  
    Presents the motivation behind studying snake-robot locomotion on diverse terrain types, provides an overview of the COBRA platform, reviews related work, and summarizes the thesis contributions and structure.

    \item \textbf{Chapter~2: Literature Review}  
    Summarizes prior work on contact modeling, deformable terrain simulation, and relevant techniques in locomotion analysis for hyper-redundant robots.

    \item \textbf{Chapter~3: Methodology}  
    Details the modeling and simulation frameworks used in this thesis. This includes:
    \begin{itemize}
        \item the contact-implicit locomotion model used for sidewinding analysis,
        \item MATLAB Simulink rigid-ground simulation for sidewinding,
        \item Project Chrono SCM deformable-terrain simulation pipeline,
        \item hardware experiment setup for validation,
        \item MATLAB Simulink modeling for tumbling,
        \item Chrono DEM-Engine granular simulation, including numerical contact mechanics and particle-based terrain generation.
    \end{itemize}

    \item \textbf{Chapter~4: Results}  
    Provides qualitative and quantitative comparison of simulation and hardware performance for sidewinding and tumbling gaits. Contact forces, slip behavior, sinkage, and locomotion efficiency are analyzed across rigid, compliant, and granular terrains.

    \item \textbf{Chapter~5: Conclusion}  
    Summarizes key findings, discusses the relevance of the developed simulation frameworks, and outlines future directions including perception-enhanced control, adaptive terrain reasoning, and planetary field deployment.
\end{itemize}

\chapter{Literature Review}
\label{chap:Literature Review}

The development of sophisticated control systems for snake robots has progressed substantially through the integration of biologically inspired learning mechanisms, rhythmic control architectures, and modern optimization-based frameworks. Early research drawing inspiration from cerebellar learning demonstrated that adaptive internal models can improve serpentine coordination by continuously refining motion patterns in response to unexpected frictional variations or disturbances, thereby offering a foundation for agile and resilient locomotion \cite{ouyang_motion_2018}. Building on this perspective, Central Pattern Generator (CPG) frameworks introduced compact oscillatory structures capable of producing rich families of 3D slithering motions, and their tunability enabled smooth transitions between gaits along with real-time modulation of locomotion speed and direction \cite{bing_towards_2017}. As the community sought greater predictability and constraint awareness, simplified averaged representations of snake robot dynamics enabled Model Predictive Control (MPC) approaches capable of respecting inequality constraints, environmental interactions, and path-following requirements while still maintaining computational tractability \cite{fukushima_model_2021}. More recent optimal control and data-driven methods have further pushed this direction by incorporating friction uncertainty estimation and efficient numerical solvers to achieve robust trajectory tracking even under unknown surface conditions \cite{li_design_2022}. Parallel efforts in amphibious robotics have shown that gradient-free online optimization can autonomously discover high-performance gaits in both swimming and crawling modes without detailed analytical models, providing strong evidence that learning and optimization can complement biological inspiration \cite{crespi_online_2008}. Nonlinear MPC (NMPC) studies have emphasized the importance of computational feasibility, revealing how constraint-handling frequency and update rates must align with the robot's intrinsic dynamics to ensure stable operation \cite{liljeback_review_2012}. Reinforcement learning formulations using path integrals have offered an alternative means of generating smooth curvature sequences for crawler locomotion, yielding robust performance across diverse terrains \cite{wang_path_2021}. Lastly, the integration of safe reinforcement learning with robust MPC has shown that high-performance adaptive controllers can be equipped with formal safety guarantees, helping advance snake robot locomotion toward real-world application domains \cite{zanon_safe_2021}. Collectively, these studies trace a clear trajectory from biologically inspired control principles toward predictive and learning-based optimization frameworks that support adaptive and constraint-aware snake robot control.

The progress made in snake robot control also connects with a growing body of work focused on extraterrestrial mobility, where robots must navigate highly irregular and extreme environments. The frontier of planetary exploration has been significantly shaped by robotic systems capable of traversing complex terrains such as lunar craters, Martian regolith, and subsurface caves. A prominent example is the Crater Observing Bioinspired Rolling Articulator (COBRA), a multimodal snake robot designed for variable lunar landscapes, where its ability to transition between slithering and tumbling locomotion provides unique mobility compared with wheeled platforms on steep crater slopes and loose sediments \cite{salagame_crater_2025}. Complementary research on thruster-assisted legged mobility for Mars has demonstrated that small, strategically applied thrust forces can help maintain balance and enforce safety constraints in low-gravity environments, with reference governors ensuring reliable control during dynamic maneuvers \cite{dangol_thruster-assisted_2020}. The growing interest in aerial reconnaissance has led to parametric studies of Mars helicopters that clarify aerodynamic, mass, and geometric constraints necessary for stable flight in the thin Martian atmosphere \cite{fujita_parametric_nodate}. Broad surveys of rotorcraft designs further highlight their value for high-resolution mapping, terrain scouting, and scientific support operations \cite{radotich_study_nodate}. At the aerodynamic level, detailed analyses of coaxial rotors under Martian-relevant conditions have revealed how rotor spacing, interference mechanisms, and induced flow behavior influence lift and efficiency \cite{zhao_investigation_2023}. Earlier foundational work also provided critical insights into ultralight rotorcraft suitable for Martian exploration long before current missions \cite{young_rotorcraft_2002}. Supporting geological studies have identified ridge-like Martian lava tubes with notable internal stability, highlighting their relevance for future exploration and habitation \cite{zhao_ridge-like_2017}. Along similar lines, numerical modeling of ice cave stability has revealed the persistence of buried ice reservoirs, guiding scientific interest toward subsurface environments with astrobiological significance \cite{williams_ice_2010}. Together, these contributions show a research landscape where planetary mobility increasingly relies on biologically inspired systems, hybrid locomotion strategies, and aerial capabilities that address extraterrestrial constraints.

Alongside these developments, machine learning has steadily reshaped how robots acquire adaptive behaviors, making it possible for controllers to improve performance directly from interaction data. Machine learning has introduced a transformative paradigm to robot control by enabling systems to learn robust behaviors directly from data and to adapt to unknown dynamics without relying only on handcrafted models. Online-learning-augmented MPC has shown that combining predictive modeling with policy refinement allows robots to preserve baseline stability while improving performance through real-world feedback \cite{bellegarda_online_2020}. In manipulation, Sim2Real strategies using tactile-based reinforcement learning have demonstrated that training on diverse simulated objects and contact scenarios can produce manipulation skills that transfer reliably to hardware, closing long-standing gaps between simulation and physical systems \cite{su_sim2real_2024}. End-to-end learning applied to quadrupedal robotics has revealed that locomotion and local navigation can be jointly optimized in unified frameworks, allowing robots to traverse unstructured terrain using energy-efficient learned gaits that require minimal tuning \cite{rudin_advanced_2022}. For snake robots, Koopman operator-based MPC has shown that linear embeddings of nonlinear dynamics can support effective and model-free path following, extending predictive control methods to systems with complex body-terrain interactions \cite{zhu_path_2022}. Reinforcement learning integrated into scenario-tree MPC has allowed autonomous surface vehicles to reliably handle uncertainty in obstacle avoidance and pursuit tasks, illustrating the broader value of combining learning with predictive planning \cite{kordabad_reinforcement_2021}. In surgical robotics, learning-based MPC approaches comparing reinforcement learning and imitation learning have highlighted the potential of data-driven control in delicate tissue manipulation tasks that demand safe but precise motions \cite{shin_autonomous_2019}. Collectively, these studies show that learning-enabled controllers can effectively complement predictive frameworks in robots operating in uncertain and dynamic environments.

Reliable autonomy depends not only on control and learning but also on the ability to understand and map the surrounding environment, which has driven rapid advances in SLAM and perception. The evolution of Simultaneous Localization and Mapping (SLAM) has been shaped by progress in semantic representation, perception-driven planning, and multisensor fusion. Real-time semantic SLAM for large-scale autonomous flight has shown that identifying high-level structures such as trunk clusters, canopy density, and ground planes improves navigation robustness in forested environments with poor visibility \cite{liu_large-scale_2022}. Visual-inertial SLAM frameworks using submapping approaches have extended this capability by supporting long-duration flight in GPS-denied environments through continuous re-localization and map refinement \cite{laina_scalable_2024}. In high-speed navigation, adaptive search-based replanning has enabled drones to refine sampling resolution and update trajectories based on environmental complexity, which increases reliability during aggressive maneuvers \cite{jarin-lipschitz_experiments_2022}. For snake robots, perception-aware locomotion has employed onboard LiDAR to reconstruct local environments and support planning in confined and low-visibility conditions \cite{yang_perception-aware_2020}. Classic work introducing the virtual chassis concept provided a simplified representation for snake robot deformation and control by defining a representative frame based on averaged link orientations, which enabled more interpretable and stable state estimation \cite{rollinson_virtual_2012}. Laser-visual-inertial odometry has achieved extremely low drift over multi-kilometer trajectories, demonstrating the capability of tightly coupled multisensor fusion for precise navigation \cite{zhang_laservisualinertial_2018}. Stereo visual-inertial odometry using MSCKF methods has further shown that lightweight perception pipelines can support stable flight at speeds exceeding 17 meters per second \cite{sun_robust_2018}. The RAPPIDS planning framework added to this ecosystem by enabling efficient collision checks and trajectory generation from a single depth image, providing a practical tool for fast navigation through clutter \cite{lee_autonomous_2021}. Together, these works show a steady progression toward SLAM systems that are increasingly semantic, adaptive, and tightly integrated with planning.

Insights from biology further broaden the design space of robotics, showing how natural systems achieve efficient and adaptive motion across air, land, and complex terrain. The field of bio-inspired robotics has expanded significantly through detailed study of how animals achieve efficient, adaptive, and versatile locomotion. A notable example is the Bat Bot (B2), which demonstrated that articulated morphing wings with only a small number of actuators can reproduce the aeroelastic behavior necessary for stable and maneuverable bat-like flight \cite{ramezani_biomimetic_2017}. Earlier biomechanical studies showed how bat wings feature coupled interactions between compliant skin, skeletal flexibility, and aerodynamic loading, suggesting that mechanical intelligence arises naturally from the synergy between morphology and actuation \cite{ramezani_bat_2016}. Principal component analysis applied to bat kinematic data identified dominant modes of biological motion and enabled robotic designs that replicate key aerodynamic effects while reducing actuation complexity \cite{hoff_optimizing_2018}. Additional computational structure designs introduced armwing mechanisms capable of producing biologically meaningful flapping motions with relatively little mechanical overhead \cite{ramezani_bat_nodate}. Inspired by reptiles, research on rolling locomotion has shown that manipulating the center of gravity can produce energy-efficient rolling motion, which is a promising complement to traditional undulatory gaits for snake-like robots \cite{yamano_efficient_2023}. Studies of helical gaits have shown that combining rolling and bending can enhance end-effector dexterity, allowing snake robots to perform complex manipulation tasks alongside locomotion \cite{elsayed_mobile_2022}. Modular snake robots equipped with docking mechanisms have expanded operational capabilities by allowing robots to connect, extend their body, and increase available degrees of freedom \cite{qin_design_2022}. Comparative dynamic analyses have shown that sinus-lifting and sidewinding gaits often provide better energy efficiency relative to lateral undulation, which informs gait selection for real-world applications \cite{ariizumi_dynamic_2017}. Collectively, these works demonstrate how studying biological systems continues to inspire new forms of mechanical intelligence and multifunctional locomotion.

Translating biological inspiration into physical machines requires strong attention to structure and mechanical design, especially for robots expected to operate in demanding or unfamiliar environments. Advanced structural analysis and mechanical design continue to shape modern robotic systems, particularly those intended for use in complex terrains. Geological investigations of lunar and Martian lava tubes have shown that large-scale voids, sometimes exceeding one kilometer in diameter, may remain structurally stable, which emphasizes the need for robotic platforms that can navigate subsurface structures. In aerial robotics, synergistic mechanical designs have combined biological insights with multi-body dynamic modeling to produce flapping-wing mechanisms that reproduce key aspects of bat flight kinematics and control \cite{hoff_synergistic_2016}. Optimization-based frameworks have identified kinematic synergies that reduce the number of required actuators while preserving aerodynamic functionality, blending computational design with biologically inspired principles. In legged and hybrid robots, generative design methods have addressed the Mobility Value of Added Mass problem by optimizing morphology and structure to reduce transport cost while maintaining mechanical robustness \cite{ramezani_generative_2021}. The redesigned Serpens robot employs series elastic actuators and modular 3D-printed components to achieve compliance, robustness, and ease of assembly, representing a practical approach to scalable snake robot design \cite{duivon_redesigned_2022}. Broader discussions of self-reconfigurable modular robots have highlighted rolling mechanisms inspired by the Golden Wheel Spider, which offer promising strategies for developing scalable and shape-changing planetary robots \cite{western_golden_2023}. Multi-loop linkage robots have shown that closed-chain mechanical designs can achieve rolling and direction-switching capabilities without the need for sophisticated actuation or control \cite{tian_multi-loop_2020}. Research on series elastic snake robots has also explored improved sensing, advanced actuation strategies, and the potential use of augmented reality to enhance teleoperation and situational awareness \cite{seetohul_snake_2022}. Together, these studies show how structural insight, mechanical intelligence, and modularity are shaping the next generation of resilient robotic platforms.

In parallel with ground and subterranean systems, aerial robots have seen major progress through improved dynamics modeling, real-time planning, and mechanically intelligent designs. The domain of autonomous aerial systems has advanced rapidly through improvements in perception, dynamic modeling, and real-time trajectory generation. The MIMIC framework showed that integrating mechanical intelligence into flapping-wing structures allows bat-like robots to perform agile and stable mid-air maneuvers, illustrating how embodied morphology can reduce control complexity \cite{delaune_mid-air_2022}. Direct-collocation-based trajectory planning methods have enabled bat-inspired robots to execute feasible transitions and aerobatic motions while respecting aerodynamic and structural constraints \cite{hoff_trajectory_2019}. In legged locomotion, thruster-assisted balancing has shown that small controlled thrust pulses can prevent falls, enable traversal of challenging obstacles, and support high-dynamics maneuvers that would otherwise be infeasible \cite{dangol_control_2021}. Safe corridor planning for quadrotors has reframed trajectory generation as a constrained quadratic program, which provides real-time navigation through cluttered spaces with formal safety guarantees \cite{liu_planning_2017}. High-speed autonomous flight systems have demonstrated navigation at speeds above 18 meters per second in GPS-denied indoor and outdoor settings, supported entirely by onboard perception and computation \cite{mohta_experiments_2018}. RAPPIDS contributed a lightweight planning method using single-depth-image collision checking, making it well suited for compact, resource-limited aerial platforms \cite{bucki_rectangular_2020}. Aerodynamic studies have clarified how coaxial rotor performance depends on rotor separation and induced interactions under Martian atmospheric conditions, guiding the design of next-generation extraterrestrial drones. Long-term surveys of Mars rotorcraft configurations have synthesized research innovations across two decades, offering a clear roadmap for future planetary aerial exploration \cite{saez_mars_2022}. Collectively, these developments illustrate how flapping-wing intelligence, thrust-assisted mobility, and safe real-time planning are advancing aerial autonomy.

Research in robotic manipulation highlights another dimension of mobility, emphasizing how robots interact physically with objects and surfaces in controlled yet flexible ways. The development of advanced robotic manipulation capabilities has benefited significantly from innovations in tactile sensing, compliant actuation, and contact-aware planning. Biologically inspired flapping-wing robots using gear-driven and linkage-based mechanisms have shown how robotic systems can replicate multi-degree-of-freedom wing kinematics, providing insights that extend beyond aerial locomotion into compliant mechanical design \cite{lessieur_mechanical_2021}. One-finger manipulation research has demonstrated that carefully planned pushing actions can achieve complex object repositioning tasks in three dimensions, highlighting the capabilities of minimal-contact manipulation strategies \cite{xiao_one-finger_2024}. Soft robotic systems equipped with optical fiber-based tactile sensor sleeves have achieved high-resolution distributed sensing, enabling manipulators to detect subtle contact forces and surface deformations during delicate interactions \cite{sareh_bio-inspired_2014}. Probabilistic contact estimation methods for quadrupeds have enabled reliable detection of ground contact events and impacts without dedicated sensors, improving robustness during dynamic motion \cite{camurri_probabilistic_2017}. In snake robots, wheel-mounted three-axis force sensors have enhanced environmental awareness by providing direct feedback on surface interactions \cite{taal_3_2009}. Contact-aware non-prehensile manipulation techniques have shown that cluttered or confined spaces can be navigated by leveraging controlled interactions with the environment, reducing reliance on grasping alone \cite{jiang_contact-aware_2023}. Methods for prehensile pushing that exploit alternating sticking and slipping phases have enabled stable regrasping and in-hand object repositioning, extending manipulation capabilities with simple grippers \cite{chavan-dafie_regrasping_2018}. Together, these efforts show that tactile intelligence and contact-based reasoning play a central role in developing dexterous robotic manipulation.

Robust locomotion and manipulation also rely on accurate computational models, and advances in simulation and numerical optimization have become essential tools for modern robotics research. The advancement of computational methods has reshaped how robotic systems are modeled, optimized, and simulated across many physical domains. Neural-network-based surrogate models combined with cubature techniques have enabled efficient approximation of high-dimensional morphing-wing dynamics, which makes real-time control and optimization feasible for systems with complex aeroelastic behavior \cite{ghanem_efficient_2021}. In locomotion on structured terrain, neural scene representation methods have allowed model predictive controllers to incorporate richer environmental features directly into decision-making processes, improving performance in settings with predictable structure \cite{wanner_model_2022}. Evolutionary computation studies using genetic programming have shown that coordinated sidewinding behaviors can emerge from simple segmented robots when subject to selection pressures, providing insight into both biological evolution and automated gait discovery \cite{tanev_automated_2005}. Gauss-principle-based dynamic formulations have provided efficient methods for computing whole-body dynamics of free-floating humanoids, supporting advanced simulation and control for space robotics and microgravity environments \cite{bouyarmane_dynamics_2012}. NeuFlow has demonstrated that high-fidelity optical flow can be computed in real time on resource-limited embedded platforms, which enables robust perception pipelines for small aerial and mobile robots \cite{zhang_neuflow_2024}. Sensor-based control in modular-loop robots has produced impressive rolling speeds of up to 26 module lengths per second, demonstrating how mechanical design and feedback control can jointly enable extreme dynamic performance \cite{sastra_dynamic_2009}. The Chrono multi-physics engine supports unified simulation of rigid bodies, granular media, deformable terrain, and fluid-structure interactions, making it a critical tool for researchers who require accurate multi-domain simulation \cite{tasora_chrono_2016}. Real-time deformable-ground simulation methods have extended these capabilities by allowing vehicles and legged robots to train, test, and evaluate performance on soft terrain without prohibitive computational cost \cite{serban_real-time_2023}. Together, these contributions demonstrate how computational innovation is driving advances in simulation fidelity, design optimization, and real-time control.

Recent work has started to unify these developments, pushing toward robotic systems that combine multiple forms of mobility, share information across teams, and make decisions based on rich perception. The robotics field continues to evolve through the integration of new mobility strategies, exploration paradigms, and perception-driven decision making. Geological analyses of Martian lava tube systems have revealed structurally stable subterranean networks with significant scientific and habitation potential, which motivates robotic systems capable of subterranean access \cite{de_oliveira_thruster-assisted_2020}. Hybrid locomotion planning has shown that combining legged, rolling, and flying capabilities within a single platform enables safe and versatile navigation across difficult terrains when coordinated using probabilistic roadmaps and reference governors that enforce constraints \cite{sihite_efficient_2022}. Distributed flight arrays composed of autonomous rotor units have demonstrated the feasibility of modular aerial platforms that can dock, separate, and fly cooperatively, suggesting innovative directions for swarm-based planetary exploration \cite{oung_feasibility_2009}. Surveys of perception-driven obstacle-aided locomotion for snake robots have emphasized the importance of environmental interaction, showing how deliberate contact with features such as walls and rocks can improve stability and traction \cite{sanfilippo_perception-driven_2017}. Multi-robot coordination strategies have employed error-constrained formation-following rules to maintain synchronized motion among multiple snake robots, enabling cooperative exploration and manipulation \cite{li_error_2023}. Traversability-aware planning frameworks have demonstrated that fusing robot internal states with environmental difficulty metrics provides safer trajectory selection in mountainous or uneven terrain \cite{yoo_traversability-aware_2024}. The SenSnake platform, with its 3D bending capabilities and distributed force sensors, has also created new opportunities for controlled experimental studies of contact-rich locomotion strategies, offering a benchmark system for future research \cite{ramesh_sensnake_2022}. Together, these developments highlight a fast-moving trend toward multimodal, perception-driven, and cooperative robotic systems that are capable of operating in the most demanding environments.

\chapter{Methodology}
\label{chap:methodology}

This chapter presents the complete methodological framework used to study
sidewinding and tumbling locomotion of the COBRA robot across rigid,
compliant, and granular terrains. The material is organized to provide a
coherent progression from analytical modeling to simulation tools that
capture different modes of contact and terrain behavior.

\textbf{Sidewinding Gait.}
Section~\ref{sec:model} introduces the contact-implicit modeling formulation
that establishes the compliant contact dynamics and the continuous-time
equations of motion used throughout this thesis. This formulation serves as
the analytical foundation for the subsequent simulation environments.

Section~\ref{sec:simulink} describes the MATLAB Simulink implementation
developed for sidewinding locomotion on rigid ground. In this setting, the
robot’s collision geometry, frictional interactions, and distributed body
contacts are handled through Level-2 S-functions that compute geometric
relations and contact quantities in real time.

Section~\ref{subsec:scm_terrain} then extends the analysis to compliant soil
interaction by introducing the Project Chrono Soil Contact Model (SCM)
pipeline. This simulation environment captures continuum-style deformation
and enables the study of sidewinding on yielding terrain without relying on
particle-resolved granular physics.

\textbf{Tumbling Gait.}
To broaden the study from serpentine locomotion to multi-modal rolling, this
chapter also includes a tumbling simulation developed in MATLAB Simulink.
Here, the terrain is rigid and high-energy impacts are resolved through a
compliant spring–damper contact law with velocity-dependent friction. This
model provides a controlled reference case for evaluating the effects of
terrain compliance.

Finally, Section~\ref{sec:chrono_dem_intro} introduces the Chrono DEM Engine
framework used for particle-resolved granular terrain simulation. This includes
the procedures for generating the granular bed, configuring slope geometry,
and coupling the robot’s kinematics with the dynamics of individual particles.
Together, these tools allow detailed investigation of how rolling locomotion
interacts with fully deformable terrain.

Taken together, these components form a unified multi-resolution analysis
pipeline that spans analytical modeling, rigid-ground simulation, compliant
terrain modeling, and granular physics. This layered structure enables a
systematic comparison across terrain types while maintaining a consistent
treatment of the robot's motion and contact interactions.

\section{Modeling}
\label{sec:model}

\begin{figure}[ht]
    \centering
    \includegraphics[width=1\linewidth]{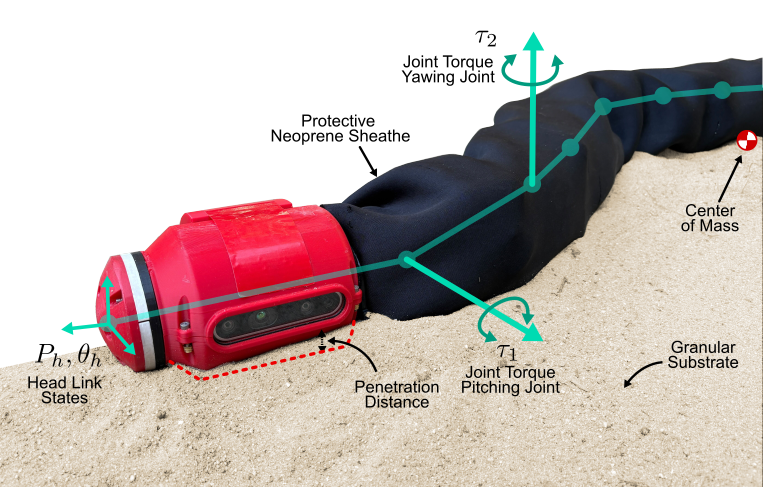}
    \caption{Closeup view of robot sand surface interactions. \textit{Image courtesy of Adarsh Salagame}}
    \label{fig:sand-closed-up}
\end{figure}

\newcommand{\R}{\mathbb{R}}

The modeling framework presented in this section follows the full contact-implicit formulation developed in the IEEE RA-L 2026 study. The goal is to capture how distributed body segments of the \ac{COBRA} robot interact with deformable ground through compliant normal and tangential forces, while maintaining an explicit and differentiable structure suitable for simulation and comparison with real terrain behavior. The formulation retains all geometric, kinematic, and force components required for evaluating sidewinding locomotion on rigid and soft substrates. For completeness, the full continuous-time equations and the semi-implicit integration scheme are included.

\paragraph{Configuration and dynamics without contact.}
We begin by describing the rigid-body dynamics of the snake robot in the absence of terrain interaction. Let $q\in\R^{n}$ denote the generalized coordinates, $v=\dot{q}$ the generalized velocities, and $\tau\in\R^{m_\tau}$ the actuator torques applied at the robot's joints. The unconstrained equations of motion are
\begin{align}
M(q)\,\dot v + C(q,v)\,v + G(q) &= B(q)\,\tau + \sum_{i=1}^{m_c} J_i(q)^{\top} f_i, \label{eq:base1}\\
\dot q &= v. \label{eq:base2}
\end{align}
The matrices $M$, $C$, and $G$ represent the standard inertial, Coriolis, and gravitational contributions. The matrix $B(q)$ maps commanded joint torques to generalized coordinates. Each potential ground contact point is indexed by $i\in\{1,\dots,m_c\}$, with $J_i$ denoting the corresponding geometric Jacobian and $f_i$ the spatial force imparted by the terrain. These forces are not known a priori; rather, they are determined through the compliant contact laws introduced below.

\paragraph{Signed distance, frames, and relative velocity.}
To compute ground interaction forces, the model relies on a smooth signed distance function $g_i(q)$ that measures separation or penetration between each contact point and the terrain surface. A positive signed distance indicates separation, zero indicates contact, and a negative value implies virtual penetration used for penalty-based compliance. At each contact location, an orthonormal triad is defined using the outward terrain normal $\hat n_i$ and two orthogonal tangent vectors $\hat t_{i,1}$ and $\hat t_{i,2}$. These vectors form the rotation matrix
\[
R_i(q) = \begin{bmatrix} \hat n_i & \hat t_{i,1} & \hat t_{i,2} \end{bmatrix},
\qquad R_i^\top R_i = I.
\]
Projecting the Cartesian relative velocity $u_i = J_i(q)\,v$ into this basis gives
\[
\tilde u_i := R_i^\top u_i
= \begin{bmatrix} u_{N,i} \\ u_{T,i} \end{bmatrix},
\]
where $u_{N,i}$ is the normal approach speed and $u_{T,i}$ is the two-dimensional tangential velocity. The contact force is similarly decomposed as $\tilde f_i = R_i^\top f_i$.

This separation of normal and tangential components allows independent modeling of compression, shear, and friction.

\paragraph{Compliant normal force (penalty + damping).}
The normal interaction between the robot and the terrain is modeled using a linear spring and damper acting only during compression. Let the penetration depth be
\[
\delta_{N,i} := \max\{0,\,-g_i(q)\},
\]
with penetration rate
\[
\dot \delta_{N,i} := -\dot g_i(q) = -\frac{\partial g_i}{\partial q}(q)\,v = - W_i(q)^\top v.
\]
The normal force is then defined as
\begin{equation}
f_{N,i} \;=\; 
\begin{cases}
k_{n,i}\,\delta_{N,i} + d_{n,i}\,\dot \delta_{N,i}, & \delta_{N,i} > 0,\\[3pt]
0, & \delta_{N,i} = 0 \text{ and } \dot\delta_{N,i}\ge 0,
\end{cases}
\label{eq:normal}
\end{equation}
where $k_{n,i}$ and $d_{n,i}$ are stiffness and damping parameters. This penalty-based force ensures that contact is one-sided (compression only) and that the reaction grows with penetration depth and impact velocity.

\paragraph{Compliant tangential force with Coulomb cap.}
To capture shear deformation at the contact interface, a tangential spring state $\sigma_i\in\R^{2}$ is introduced. This state accumulates tangential displacement when the contact is in a sticking regime, governed by
\begin{equation}
\dot\sigma_i = u_{T,i}.
\label{eq:sigmadot}
\end{equation}
A linear spring and damper provide a proposed tangential force
\begin{equation}
\tilde f_{T,i}^{\,\text{prop}} = -\,k_{t,i}\,\sigma_i - d_{t,i}\,u_{T,i},
\label{eq:ftprop}
\end{equation}
but this force must remain consistent with Coulomb friction. The final applied tangential force is
\begin{equation}
f_{T,i} =
\begin{cases}
\tilde f_{T,i}^{\,\text{prop}}, & \|\tilde f_{T,i}^{\,\text{prop}}\| \le \mu_i f_{N,i}, \\[4pt]
-\mu_i f_{N,i}\,\dfrac{u_{T,i}}{\max\{\|u_{T,i}\|,\,\epsilon_v\}}, & \|\tilde f_{T,i}^{\,\text{prop}}\| > \mu_i f_{N,i} ,
\end{cases}
\label{eq:friction}
\end{equation}
which smoothly transitions between stick and slip. During slip, the shear state is frozen or decayed to avoid artificial energy buildup. This formulation captures shear hysteresis that is characteristic of locomotion on sand and other yielding substrates.

\paragraph{Contact force mapping to generalized coordinates.}
Reassembling the forces into the world frame gives
\begin{align}
f_i &= R_i\,\tilde f_i \\
    &= \hat n_i\,f_{N,i}
       + \hat t_{i,1}\,[f_{T,i}]_1
       + \hat t_{i,2}\,[f_{T,i}]_2 ,
\label{eq:contact_force}
\end{align}
and the total contribution to the generalized coordinates is $J_i(q)^{\top} f_i$. Substitution into \eqref{eq:base1} and \eqref{eq:base2} yields the complete compliant-contact dynamics.

\paragraph{Compact closed-form equations of motion (continuous time).}
For convenience, define the activation indicator $\chi_i(q)$ to identify active contacts. The full closed-form system is
\begin{align}
M\,\dot v + H &=
   B(q)\,\tau
   + \sum_{i=1}^{m_c} J_i^{\top} R_i
   \begin{bmatrix}
      f_{N,i}(q,v) \\ f_{T,i}(q,v,\sigma_i)
   \end{bmatrix}, 
   \label{eq:eom_cont} \\[3pt]
\dot q &= v, 
   \label{eq:qdot_cont} \\[3pt]
\dot \sigma_i &=
   \begin{cases}
      u_{T,i}(q,v), &
         \chi_i=1 \text{ and }
         \|\tilde f_{T,i}^{\,\text{prop}}\|\le \mu_i f_{N,i}, \\[2pt]
      -\alpha_i\,\sigma_i, &
         \chi_i=1 \text{ and }
         \|\tilde f_{T,i}^{\,\text{prop}}\|> \mu_i f_{N,i}, \\[2pt]
      -\alpha_i\,\sigma_i, & \chi_i=0 ,
   \end{cases}
   \label{eq:sigma_cont}
\end{align}
with $H=C\,v + G$. These equations describe a coupled set of rigid-body dynamics and compliance-induced deformation modes across many distributed contacts.

\paragraph{Energetics and passivity note.}
The combination of normal and tangential damping, along with the Coulomb projection, ensures that the ground interaction does not introduce artificial energy into the system. The total power transferred through all contacts satisfies
\[
\sum_i \tilde f_i^\top \tilde u_i \;=\; \sum_i \big(f_{N,i}\,u_{N,i} + f_{T,i}^\top u_{T,i}\big) \;\le\; 0,
\]
aside from the work done by actuators and storage in the elastic components. This passivity property plays an important role in maintaining numerical stability during simulation.

\paragraph{Semi-implicit (symplectic) Euler step (optional).}
To simulate the dynamics efficiently, a semi-implicit time-stepping scheme may be used. Given $(q_k, v_k, \{\sigma_{i,k}\})$ and step size $h$:

\begin{equation*}
\small
\begin{aligned}
q_{k+1} &= q_k + h\,v_{k+1}, \\[3pt]
\delta_{N,i,k+1} &= \max\{0,\,-g_i(q_{k+1})\}, \\[3pt]
u_{i,k+1} &= J_i(q_{k+1})\,v_{k+1}, \\[3pt]
f_{N,i,k+1} &= \big[k_{n,i}\,\delta_{N,i,k+1} 
   + d_{n,i}\,(-W_i(q_{k+1})^\top v_{k+1})\big]_+, \\[3pt]
\tilde f_{T,i,k+1}^{\,\text{prop}} 
   &= -k_{t,i}\,\sigma_{i,k} - d_{t,i}\,u_{T,i,k+1}, \\[3pt]
f_{T,i,k+1} &= 
   \mathrm{Proj}_{\|\cdot\|\le \mu_i f_{N,i,k+1}}
   \!\big(\tilde f_{T,i,k+1}^{\,\text{prop}}\big), \\[4pt]
v_{k+1} &= v_k 
   + h\,M(q_{k+1})^{-1} \Big(
   B(q_{k+1})\tau_k
   - C(q_{k+1},v_{k+1})v_{k+1}
   - G(q_{k+1}) \\[-1pt]
   &\hspace{5.1cm}
   + \sum_i J_i(q_{k+1})^\top R_i(q_{k+1})
   \begin{bmatrix}f_{N,i,k+1}\\[2pt] f_{T,i,k+1}\end{bmatrix}
   \Big), \\[5pt]
\sigma_{i,k+1} &= 
   \begin{cases}
      \sigma_{i,k} + h\,u_{T,i,k+1}, &
         \|\tilde f_{T,i,k+1}^{\,\text{prop}}\|
         \le \mu_i f_{N,i,k+1}, \\[2pt]
      (1-h\alpha_i)\,\sigma_{i,k}, &
         \text{otherwise}\ \delta_{N,i,k+1}=0 .
   \end{cases}
\end{aligned}
\end{equation*}

The scheme solves a small nonlinear system at each step and is well suited for stiff ground-contact conditions. It is the basis for all compliant-contact simulations presented in later chapters.

\section{Sidewinding Gait Analysis}
\label{sec:sidewinding_intro}

Sidewinding is the primary mode of locomotion examined in this thesis and serves
as the basis for understanding efficient propulsion over flat and moderately
compliant terrain. The gait emerges from coordinated pitching and yawing waves
that travel along the length of the robot, creating controlled regions of
contact while keeping non-supporting links elevated to reduce drag. Through
these alternating patterns of contact and lift, the robot generates directional
reaction forces that support stable and efficient motion.

To study this behavior in a consistent manner, three complementary platforms
are used throughout the analysis. Rigid-ground modeling is carried out in
Simscape Multibody, which offers a control-oriented environment for evaluating
distributed contacts and wave coordination. Compliant terrain behavior is then
investigated using Chrono SCM, allowing soil deformation to be incorporated in
a continuum-style manner and enabling the evaluation of sinkage and load
redistribution effects. Finally, hardware experiments on laboratory flooring and
loose sand provide physical validation, with trajectory estimation obtained from
the Intel RealSense D435i through visual inertial odometry.

These combined studies enable a direct comparison of sidewinding performance
across rigid, compliant, and granular terrain conditions. The results highlight
how the gait responds to reduced support stiffness, increased slip, and terrain
yielding, while also demonstrating that rigid-ground models remain sufficiently
predictive for short-horizon planning tasks that rely on estimating the net
direction and magnitude of motion.

\section{MATLAB Simulink Simulation for Rigid Ground}
\label{sec:simulink}

To evaluate sidewinding locomotion under idealized terrain conditions, a rigid
ground model is first implemented using MATLAB Simscape Multibody. This
environment provides a computationally efficient setting for validating the
baseline dynamics of the COBRA platform without the additional complexity
associated with soil deformation. The insights obtained here establish a useful
reference when the compliant and granular terrain models are introduced in
Sections~\ref{subsec:scm_terrain} and \ref{sec:chrono_dem_intro}.

\begin{figure}[ht]
    \centering
    \includegraphics[width=0.80\textwidth]{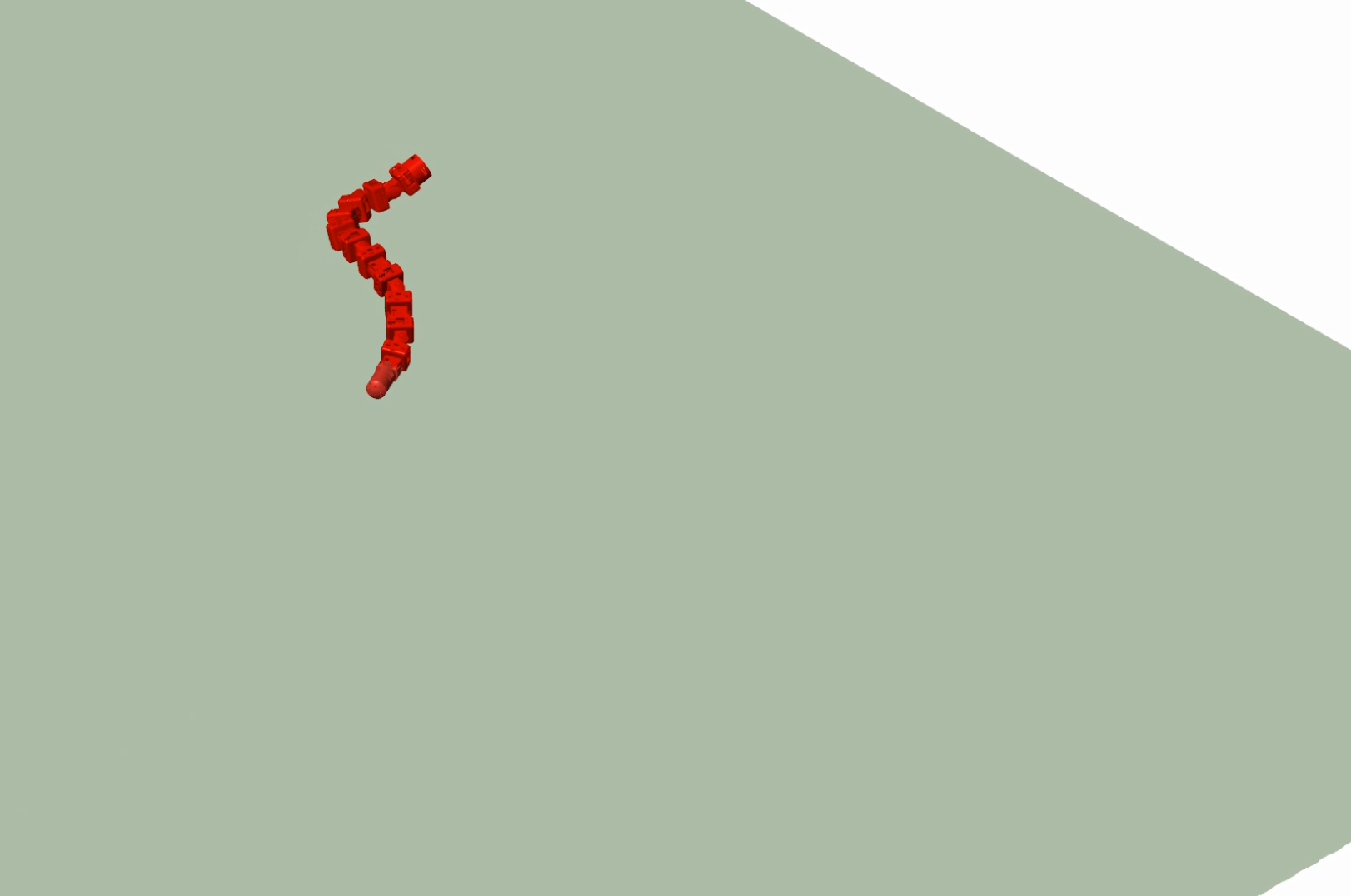}
    \caption{Hard-ground contact simulation results in MATLAB: normal and tangential force behavior during rigid surface interaction.}
    \label{fig:matlab_hard_ground}
\end{figure}

Rigid ground simulations are particularly valuable when designing control
frameworks that require real-time feasibility, such as receding horizon
optimization. In these settings, the model must remain lightweight yet capable
of capturing the essential translational behavior produced by distributed
frictional contacts along the body. As a result, the rigid-ground environment
serves as an efficient platform for prototyping gait strategies and verifying
expected motion direction and displacement during sidewinding.

\subsection*{Ground Contact Model in Simscape Multibody}

Contact interactions are modeled through a nonlinear spring–damper formulation
for both normal and tangential directions. The normal reaction force is expressed as

\begin{align}
    f_n &= s(d,w)\,(k\,d + b\,\dot d), \label{eq:simscape_normal}
\end{align}

where $d$ denotes the penetration depth of the link into the ground plane and
$\dot{d}$ is its rate of change. The stiffness parameter $k$ determines the
magnitude of the elastic restoring force, while the damping coefficient $b$
regulates the rate-dependent component that absorbs energy during impact. The
smoothing function $s(d,w)$ introduces a gradual transition into contact over a
width $w$, ensuring that forces increase continuously rather than appearing
abruptly at $d = 0$. This avoids numerical instabilities while still imposing
strong resistance against penetration.

Tangential contact behavior is governed by a friction model defined through

\begin{align}
    \|f_t\| &= \mu_\mathrm{eff}(u_t)\,\|f_n\|, \label{eq:simscape_friction}
\end{align}

where $u_t$ is the tangential slip velocity at the contact point. The quantity
$\mu_\mathrm{eff}(u_t)$ represents the effective friction coefficient, which
varies smoothly between the static value $\mu_s$ at very low slip speeds and the
dynamic value $\mu_d$ at larger slip speeds. This dependency reflects the
physical tendency of surfaces to resist motion strongly when nearly stationary,
followed by reduced resistance once slipping occurs. Introducing a critical slip
velocity creates a smooth transition between these regimes and prevents abrupt
switching between sticking and sliding.

\subsection*{Simulation Configuration}

To produce a clean baseline for rigid terrain, the contact model parameters are
selected to approximate a stiff, non-compliant surface. The normal stiffness is
set to $k = 10^{4}$ and the damping to $b = 10^{3}$, providing strong resistance
to penetration with moderate dissipation of impact energy. The smoothing width
is chosen as $w = 10^{-3}\,\text{m}$ to maintain continuous contact transitions.
The friction coefficients are assigned values of $\mu_s = 0.5$ and $\mu_d =
0.3$, while the critical slip speed is set to $10^{-3}\,\text{m/s}$ so that the
transition between static and dynamic friction occurs smoothly.

These values produce a terrain model that behaves effectively rigid during
sidewinding and prevents unrealistic sinkage or deformation, which is essential
when isolating the locomotion mechanics associated purely with frictional
contact.

\subsection*{Simulation Outputs and Usage}

The Simscape implementation provides time-synchronized measurements that include
joint states, actuator torques, and estimated ground reaction forces at each
link, along with the motion of the robot’s center of mass. These outputs enable
assessment of gait efficiency under idealized ground conditions and form the
basis for comparisons with the compliant and granular terrain simulations
discussed later. They also support early controller validation by allowing
sidewinding wave parameters to be tested in a setting where the terrain
introduces no additional complexity.

Although this model captures the principal frictional interactions that drive
sidewinding, it does not reproduce effects such as sinkage, shear failure, or
redistribution of support forces caused by soil yielding. These behaviors are
instead addressed in higher-fidelity simulations in subsequent sections. As a
result, the rigid-ground Simscape environment functions as an essential
reference against which the influence of terrain deformability on COBRA
locomotion can be quantified.

\section{Project Chrono}
\label{sec:project_chrono}

Project Chrono is an open-source multi-physics simulation framework designed
for high-fidelity modeling of articulated mechanisms, contact interactions,
deformable media, and large-scale multi-body dynamics \cite{tasora_chrono_2016}.
Its modular architecture allows different physical domains to be incorporated
within a single environment, ranging from rigid-body dynamics and soft material
models to granular flow, fluid interaction, and external cosimulation.

A key feature of Chrono is its comprehensive terrain modeling library, which
supports several distinct representations of ground behavior. These models span
a spectrum of complexity, beginning with simple idealized planes and extending
to fully deformable particle-based media. Flat terrain provides an infinite,
perfectly rigid surface with uniform elevation and is appropriate for
idealized locomotion validation. Rigid terrain introduces arbitrary geometric
variations through height maps, enabling controlled testing over structured
surfaces. CRG terrain extends this idea further by supporting curvilinear road
profiles that are commonly used in automotive and structured environment
simulations.

For deformable ground modeling, Chrono offers multiple options. The Soil
Contact Model (SCM) approximates continuum-style soft soil by allowing vertical
sinkage and permanent deformation under load, creating a realistic intermediate
representation of compliant terrain. Granular terrain is captured through the
Discrete Element Method, where individual particles interact through
collision-based forces and produce complex behaviors such as shear bands, slip
layers, and localized compaction. At the highest fidelity, Chrono includes
finite element terrain models based on the Absolute Nodal Coordinate
Formulation, which allow deformable solids to be represented with detailed
elastic and plastic responses.

This breadth of terrain models enables the selection of an appropriate level of
fidelity for each experiment in this thesis. Rigid-ground validation is carried
out in MATLAB Simulink, soft-terrain behavior is studied through the SCM
framework, and high-resolution granular effects are analyzed using DEM. Chrono
therefore forms the foundation of the multi-resolution simulation pipeline
developed in later sections and allows terrain complexity to be increased in a
controlled and systematic manner.

\section{Setting up Project Chrono}
\label{sec:chrono_setup}

A custom Chrono build environment was prepared for this thesis and published
internally for the Silicon Synapse Lab. The repository provides a validated
configuration of Chrono v9.0.1 with all required dependencies, module settings,
and source adjustments needed for GPU acceleration, Multicore physics, and both
SCM and DEM terrain models \cite{tasora_chrono_2016}. The complete setup is
available at:

\begin{center}
\url{https://github.com/SS-Lab-at-NU/chrono-custom}
\end{center}

To support deformable terrain simulation for snake-robot locomotion, Chrono was
built with the Vehicle module for soil interaction, the Multicore and GPU
modules for parallelized and CUDA-accelerated computation, and the VSG
visualization stack for high-performance rendering. Parsers were enabled for
mesh import, and the Irrlicht demos were included to assist with quick
validation runs. Several small source-level updates were required to ensure compatibility with
recent hardware. Chrono’s default CUDA version checks were extended so that the
RTX 4050 and CUDA~12.4 toolchain would be recognized, and Thrust version
validation was broadened to accommodate newer releases used in Multicore builds.
These updates prevent the build system from incorrectly disabling GPU features.

The VSG visualization stack was compiled manually because upstream scripts
contained version conflicts. All VSG packages were installed under a dedicated
local prefix, and a stable release of vsgImGui was checked out to guarantee
reliable GUI operation on Ubuntu 22.04 with CUDA~12.4. A validated CMake configuration was then generated, enabling both GPU and
Multicore physics, specifying architecture 89 for Ada-Lovelace GPUs, and
explicitly referencing Blaze and Thrust to avoid auto-detection errors. This
profile served as the basis for all SCM and DEM simulations presented later in
Sections~\ref{subsec:scm_terrain} and \ref{sec:chrono_dem_intro}. All required patches, instructions, and dependency scripts are included in the \texttt{chrono-custom} repository, ensuring that future researchers can rebuild the environment without configuration issues and maintain a consistent simulation platform across the lab.

\section{Deformable SCM (Soil Contact Model)}
\label{subsec:scm_terrain}

The SCM implementation in Project Chrono provides a deformable soil model that
captures terramechanical effects important for locomotion on loose terrain. In
contrast to rigid or penalty-based compliant surfaces, SCM explicitly models
sinkage, plastic yield, shear deformation, and bulldozing flow, while remaining
computationally efficient for multi-body simulation \cite{tasora_chrono_2016}.
This enables realistic soil–robot interaction during sidewinding, where ground
compliance strongly influences propulsion, stability, and body shaping
throughout the gait.

The terrain is represented as a Cartesian grid defined over the horizontal
coordinates of a local frame. Each grid node stores its undeformed elevation
$z_0(x,y)$, the current vertical deformation $z(x,y)$, contributions from
elastic and plastic sinkage, the accumulated shear state, and the associated
normal pressure and shear stress. To avoid allocating the full grid, Chrono
creates nodes only when the robot interacts with the surface, maintaining a
sparse hash-mapped structure that scales with the area of deformation rather
than the total terrain size. This allows large domains to be simulated without
excessive memory cost.

At every time step, vertical rays are cast from active terrain nodes toward the
robot. When a ray intersects the geometry, penetration depth is computed and
contact forces are applied. Ray casting is multi-threaded for speed and is
restricted to the oriented bounding box of the robot so that empty regions are
not tested unnecessarily.

\begin{figure}[ht]
    \centering
    \includegraphics[width=0.95\linewidth]{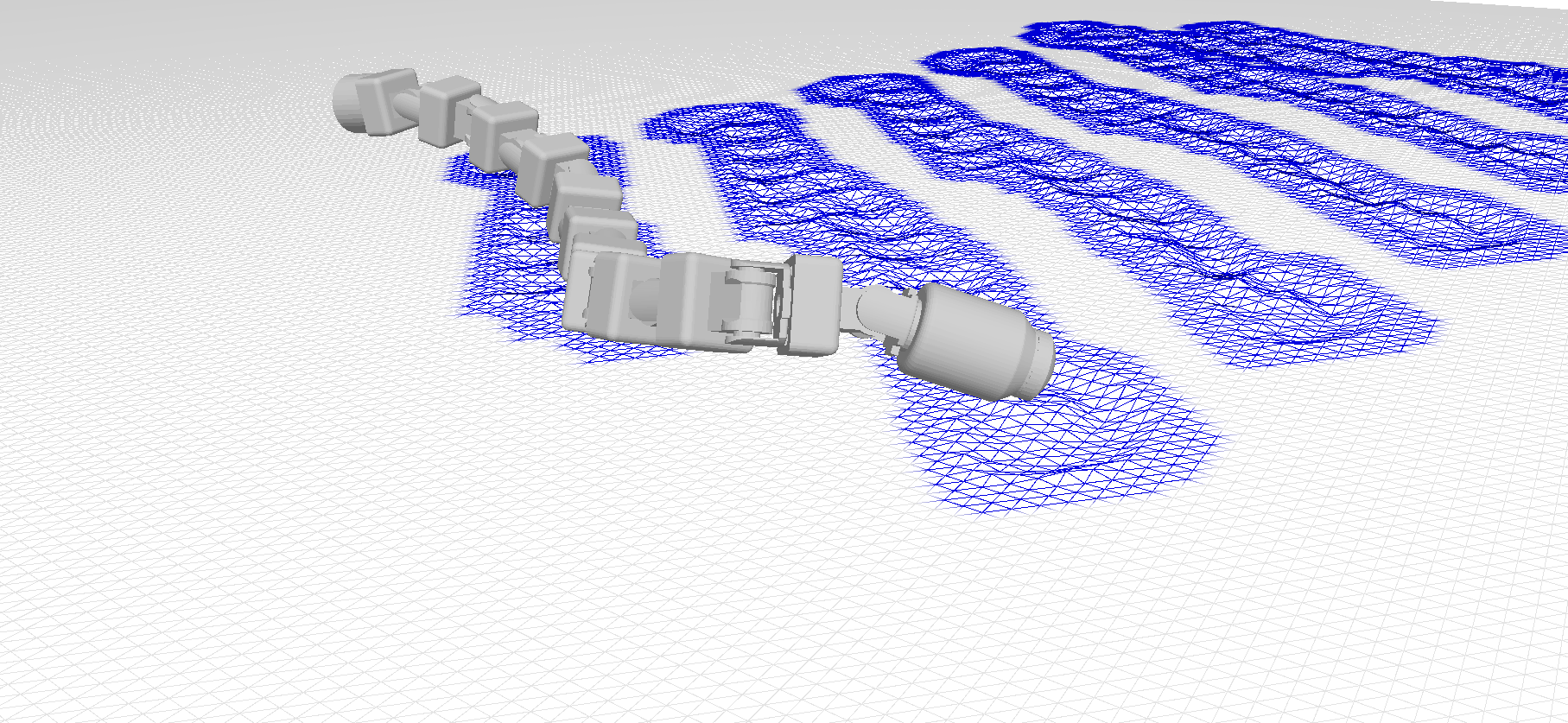}
    \caption{SCM simulation showing vertical deformation of terrain nodes under
    the COBRA robot during sidewinding. The mesh illustrates permanent soil
    depressions, shear displacement, and berm accumulation across multiple gait
    cycles.}
    \label{fig:scm_mesh_refinement}
\end{figure}

Vertical soil response is governed by the Bekker–Wong relationship:
\begin{equation}
    \sigma = (k_c\,b + k_\phi)\,y^n,
    \label{eq:bekker}
\end{equation}
where $\sigma$ is the normal pressure, $y$ the sinkage, $k_c$ and $k_\phi$ the
cohesive and frictional stiffness parameters, $n$ a compaction-dependent
exponent, and $b$ a characteristic contact width. For non-rectangular contact
patches, $b$ is approximated using the area $A$ and perimeter $L$ of the
contact region:
\[
b \approx \frac{2A}{L}.
\]

Shear deformation contributes to traction through a Janosi–Hanamoto law:
\[
\tau = \bigl( c + \sigma \tan\varphi \bigr)\left(1 - e^{-\frac{j}{K}}\right),
\]
where $c$ is cohesion, $\varphi$ the soil friction angle, $j$ the accumulated
shear displacement, and $K$ the characteristic shear deformation length. The
model captures the transition from elastic shear to plastic flow as the
traction limit is approached and exceeded.

SCM tracks permanent deformation produced when the robot overruns a region of
soil. Material is pushed laterally, forming berms and ruts that influence
future contacts. Parameters such as erosion angle, lateral flow factor, and the
number of refinement steps per update can be adjusted to match different soil
types.

Viscous damping contributes to the stability of terrain response through
\[
f_d = -R \cdot \dot{y},
\]
where $R$ is the vertical damping per unit area. This term is particularly
important for fast motions or aggressive gaits, where numerical stability may
otherwise degrade.

\begin{sloppypar}
The deformable terrain is implemented through the class
\texttt{chrono::vehicle::SCMTerrain}. It supports initialization from flat
surfaces, heightmaps, or input triangle meshes. Soil parameters can be assigned
globally or spatially, and detailed contact information can be queried for
either individual bodies or specific nodes. The API also provides access to all
modified grid nodes for visualization and to overlays that display sinkage,
pressure, yield state, and bulldozing effects. A co-simulation mode is also
available when terrain forces need to be exported without being applied
internally.\\
SCM achieves near real-time performance by storing only active nodes, limiting
collision checks to the robot’s immediate vicinity, parallelizing ray casting,
and updating visualization incrementally. When Chrono is built with CUDA,
additional GPU acceleration is available for certain components.
\end{sloppypar}

SCM provides a mid-fidelity terrain model situated between rigid or penalty
ground, which cannot capture soil deformation, and particle-based DEM, which
offers high predictive accuracy at significantly greater computational cost.
This middle ground enables systematic exploration of gait behavior under
varying soil strengths and supports direct comparison with experimental data on
loose sand.

\section{SCM Terrain Mathematical Formulation}
\label{subsec:scm_mathematical_formulation}

Chrono’s SCM implementation is built on semi-empirical soil mechanics models
originally developed by the German Aerospace Center (DLR)
\cite{krenn_scm_2009,krenn_soft_2011} and later generalized to support arbitrary
rigid geometries in real-time simulation environments \cite{serban_real-time_2023}.
The underlying principle is to relate the deformation of the soil at each
terrain node to the normal and tangential reaction forces that oppose penalties
to the surface and resist shear motion. In this manner, SCM captures the core
terramechanical behavior that governs locomotion over loose or deformable
terrain.

\textbf{Normal pressure–sinkage relationship.}
SCM treats the soil as a vertically deformable elastic–plastic medium. At each
terrain node lying beneath the contact footprint, the normal stress is computed
using the Bekker–Wong pressure–sinkage equation:
\begin{align}
    p &= \left( \frac{K_c}{b} + K_{\phi} \right) z^n,
    \label{eq:bekker_scm}
\end{align}
which relates the sinkage depth $z$ to the normal pressure $p$ generated by the
terrain. The quantity $z$ is the vertical displacement of the soil relative to
its undeformed state and measures how far the surface has been compressed by the
robot. As $z$ increases, the material compresses more densely, resulting in
higher terrain resistance.

The parameter $K_c$ is the cohesive modulus and captures the contribution of
soil cohesion to the resulting pressure. Cohesion represents intermolecular
forces or cementation in the soil grains and produces resistance even when the
normal load is small. The term $K_{\phi}$ is the frictional modulus and
incorporates resistance due to internal friction of the soil. Soils with higher
friction exhibit greater stiffness when compressed.

The exponent $n$ controls the degree of nonlinearity in the pressure–sinkage
law. A value of $n < 1$ corresponds to a material that stiffens slowly, while
$n > 1$ corresponds to a material that rapidly becomes rigid with penetration.
Loose dry sand typically exhibits $n$ between 0.5 and 1.0, giving the terrain a
moderately nonlinear response.

The patch size $b$ introduces the effect of footprint geometry into the
pressure–sinkage relationship. Smaller contact widths concentrate force over a
narrower area and therefore increase pressure for the same sinkage depth. Since
snake robots do not form rectangular footprints, $b$ is estimated from the area
$A$ and perimeter $L$ of the contact region:
\begin{align}
    b \approx \frac{L}{2A}.
\end{align}
This expression arises from Bekker’s adaptation for arbitrary geometries and
produces an effective width consistent with the observed terramechanics of
complex footprints.

The form of \eqref{eq:bekker_scm} ensures that pressure rises sharply as the
sinkage grows, in agreement with field observations that sand becomes markedly
stiffer once particles compact and interlock under deeper penetration.

\textbf{Tangential shear resistance and yield.}
SCM computes tangential shear stress generated by horizontal slip using the
Janosi–Hanamoto law:
\begin{align}
    \tau &= \tau_\text{max}\,\bigl( 1 - e^{-j/k} \bigr),
    \label{eq:jana_scm}
\end{align}
where $\tau$ is the shear stress opposing motion along the soil surface. The
variable $j$ denotes the accumulated shear displacement and measures how far the
soil at a node has been dragged relative to its original position. The parameter
$k$ is a characteristic shear deformation length scale, determining how quickly
soil transitions from elastic response to plastic yield.

The maximum attainable shear stress,
\begin{align}
    \tau_\text{max} = c + p\,\tan(\phi),
\end{align}
follows a Mohr–Coulomb relationship. The cohesion $c$ captures the inherent
shear strength of the soil in the absence of normal pressure. The term $p\tan\phi$
adds a pressure-dependent contribution due to friction, where $p$ is the normal
pressure and $\phi$ is the internal friction angle. This angle describes the
resistance of soil grains to sliding past each other and grows with particle
roughness and angularity.

At small values of $j$, the exponential term in \eqref{eq:jana_scm} is close to
zero, and $\tau$ increases almost linearly with shear displacement. In this
regime, the soil behaves elastically: deformation is reversible, and grains
undergo small displacements relative to neighbors. As $j$ becomes large, the
exponential term approaches one, and $\tau$ saturates at $\tau_\text{max}$,
indicating plastic failure. Once the soil yields, particles rearrange
irreversibly, developing ruts and displaced material. This transition is
fundamental to sidewinding traction, as it governs how effectively the robot
uses lateral force generation to propel itself.

\textbf{Grid-based vertical deformation model.}
SCM restricts deformation to the vertical axis, assuming that lateral
displacement of soil appears primarily through mound and berm formation rather
than through large-scale horizontal compression. Under this assumption, each
terrain node stores the net vertical deflection, separated into elastic and
plastic components. Plastic deformation persists after the load is removed,
preserving ruts generated by the robot and enabling cumulative deformation
across multiple gait cycles. Elastic recovery appears only in the decompression
phase when the robot unloads the soil.

This vertical-only deformation model efficiently captures the principal effects
of sinkage and berm growth, which play dominant roles in locomotion over dry
granular media.

\textbf{Contact identification through ray casting.}
Terrain nodes cast vertical rays toward the robot’s geometry to detect contact.
If a ray intersects the robot, the penetration depth is calculated, and the
node’s deformation state is updated accordingly. The accumulated sinkage and
shear history are then used to compute normal and tangential forces, which are
mapped back to the robot through standard rigid-body force accumulation.
Ray-casting is limited to nodes inside the oriented bounding box of the robot
to avoid unnecessary geometric queries, and the procedure is parallelized to
maintain efficiency.

\textbf{Parameter selection for loose sand sidewinding.}
The SCM parameters used in this thesis are calibrated from experiments on dry
natural sand:
\begin{align*}
    n &= 0.6, &
    K_c &= 25~\mathrm{N/m^{(1+n)}}, &
    K_\phi &= 5\times10^4~\mathrm{N/m^{(2+n)}}, \\
    k &= 0.04~\mathrm{m}, &
    c &= 100~\mathrm{Pa}, &
    \phi &= 28^\circ.
\end{align*}
The exponent $n = 0.6$ reflects the moderate nonlinear hardening typical of
granular soil. The cohesive modulus $K_c$ is small relative to the frictional
modulus $K_\phi$, consistent with cohesionless media such as dry sand. The
shear deformation length $k = 0.04\,\mathrm{m}$ sets the scale over which shear
stress builds up before yielding. The cohesion $c = 100\,\mathrm{Pa}$ and
internal friction angle $\phi = 28^\circ$ align with conventional terramechanics
data for loose sand \cite{janosi_analytical_1961} and are validated for
snake-robot simulation in recent Chrono studies \cite{serban_real-time_2023}.
These parameters produce realistic sinkage depths, slip patterns, and pressure
fields consistent with experimental sidewinding trials.

\textbf{Locomotion parameterization for SCM simulations.}
The robot’s motion is driven by prescribed pitching and yawing waves along the
backbone. The joint trajectories for vertical and horizontal articulation are:
\begin{align}
    q_\text{ver}(t) &= A_\text{ver}\,\sin\big(2\pi f t + \tfrac{\pi}{6}i\big),
    & i &= 0,2,4,\dots,10 \\
    q_\text{hor}(t) &= A_\text{hor}\,\sin\big(2\pi f t + \tfrac{\pi}{6}j\big),
    & j &= 1,3,5,\dots,11
\end{align}
where $A_\text{ver}$ and $A_\text{hor}$ are the vertical and horizontal
amplitudes, and $f$ is the gait frequency. Two gait patterns are used:
\[
\text{Gait 1:}\;
A_\text{ver}=40^\circ,\; A_\text{hor}=20^\circ
\qquad
\text{Gait 2:}\;
A_\text{ver}=60^\circ,\; A_\text{hor}=30^\circ.
\]
Larger amplitudes increase the lift of non-contacting segments and concentrate
forces on the subset of links in contact. This modulation of load distribution
creates varying sinkage depths, shear displacement patterns, and bulldozing
behavior that allow SCM to reveal how terrain strength influences propulsion.

\section{Empirical Parameters from Simulation Code}
\label{sec:scm_empirical_params}

The Chrono SCM simulations in this thesis rely on a specific set of numerical,
material, and soil parameters that were tuned to reproduce the locomotion
behavior observed in loose sand experiments. All values reported in this
section are taken directly from the simulation implementation and correspond
exactly to the configuration used during terrain testing. The purpose of this
section is to document these parameters and to explain the physical role of
each, particularly those that are not standard continuum mechanics quantities.

\subsection{SCM Terrain and Soil Properties}

The SCM terrain is initialized as a flat patch whose dimensions are chosen to
comfortably contain several full sidewinding cycles of the COBRA robot. The
terrain extent is set to $9\,\text{m} \times 8\,\text{m}$ so that the robot can
execute multiple gait cycles without reaching the boundary or reusing already
disturbed terrain in the primary analysis window. The reference level of the
terrain is placed at $z = -0.5\,\text{m}$, which provides sufficient vertical
clearance for visualization and ensures that terrain deformation remains within
a well-behaved numerical range. The grid spacing is chosen as
$\Delta = 0.02\,\text{m}$, which offers a balance between spatial resolution
and computational cost: this resolution is fine enough to capture contact
patches under individual links while still allowing simulations to run at a
reasonable speed.

The soil response parameters used in the SCM formulation are summarized in
Table~\ref{tab:scm_params}. Each of these plays a distinct role in shaping how
the terrain reacts to loading from the robot.

\begin{table}[h!]
    \centering
    \caption{SCM deformable soil model parameters used in simulation.}
    \label{tab:scm_params}
    \begin{tabular}{l c c}
    \toprule
    Parameter & Symbol & Value \\
    \midrule
    Frictional modulus & $K_\phi$ & $0.05\times10^{6}\,\text{N/m}^{2+n}$ \\
    Cohesive modulus & $K_c$ & $25\,\text{N/m}^{1+n}$ \\
    Sinkage exponent & $n$ & $0.6$ \\
    Cohesion & $c$ & $25\,\text{Pa}$ \\
    Internal friction angle & $\phi$ & $28^\circ$ \\
    Janosi shear coefficient & $k$ & $0.04\,\text{m}$ \\
    Elastic foundation stiffness & $K$ & $2.5\times10^{6}\,\text{Pa/m}$ \\
    Vertical damping coefficient & $R$ & $5\times10^{3}\,\text{Pa}\cdot\text{s/m}$ \\
    \bottomrule
    \end{tabular}
\end{table}

The frictional modulus $K_\phi$ and cohesive modulus $K_c$ appear in the
Bekker–Wong pressure–sinkage law and together control how strongly pressure
rises with increasing sinkage. A larger $K_\phi$ increases stiffness associated
with internal friction between grains, while $K_c$ represents the part of the
stiffness that can be attributed to cohesive forces. The exponent $n = 0.6$
determines the nonlinearity of the pressure–sinkage relationship: a value less
than one corresponds to a material that stiffens gradually and is consistent
with loose dry sand.

The cohesion $c = 25\,\text{Pa}$ is relatively small, which reflects the
near cohesionless nature of dry granular material. The internal friction angle
$\phi = 28^\circ$ sets the slope of the Mohr–Coulomb failure envelope and
captures how much shear resistance increases with normal pressure. The Janosi
shear coefficient $k = 0.04\,\text{m}$ defines the characteristic shear
displacement over which shear stress grows from zero to its maximum value. A
smaller $k$ would cause shear stress to saturate rapidly with very little slip,
whereas a larger $k$ would require more slip before yielding occurs. The chosen
value was selected so that the amount of slip before failure matches the
observed motion of the sand under the robot in physical experiments.

The elastic foundation stiffness $K = 2.5\times10^{6}\,\text{Pa/m}$ acts as an
additional vertical stiffness term that stabilizes the terrain when deformation
is small and avoids overly soft response at negligible sinkage. This parameter
does not originate directly from classical terramechanics, but is used
practically in Chrono to maintain numerical robustness. The vertical damping
coefficient $R = 5\times10^{3}\,\text{Pa}\cdot\text{s/m}$ governs the rate
dependent component of the vertical soil reaction. It introduces viscous-like
damping in the vertical direction, which suppresses oscillations in surface
height when the robot steps on and off the terrain and helps the simulation
remain stable when gait frequencies increase. These values were tuned so that
the simulated sinkage, recovery speed, and damping of the terrain matched the
qualitative appearance and time scales observed in video recordings of the
physical experiments.

Beyond the pressure–sinkage and shear laws, SCM includes several parameters
that control lateral redistribution of soil due to bulldozing. The erosion angle
is set to $25^\circ$. This angle defines the slope at which soil is no longer
able to support itself and begins to slide sideways, forming berms along the
edges of ruts. A larger erosion angle would permit steeper walls in the
deformed terrain, while a smaller one would cause material to spread more
readily. The chosen value was selected to produce berm profiles that visually
matched the side piles of sand observed in the experiments.

The flow factor is assigned a value of $1.0$. This factor scales the amount of
material redistributed laterally during each erosion update. A flow factor less
than one would reduce the amount of soil that moves outward in each step,
leading to sharper, more localized depressions, while a value greater than one
would overemphasize lateral spreading and flatten the terrain too quickly. By
setting the flow factor to unity, the simulation preserves a direct
correspondence between the computed deformation and the empirical terramechanics
model, and we avoid artificially amplifying or suppressing lateral flow.

The erosion iterations per time step are set to $2$. Each erosion iteration
applies the bulldozing update rules once over the currently affected nodes.
Running more iterations per time step allows soil to evolve toward a smoother,
more equilibrated shape, whereas a single iteration yields a more jagged
surface. We found that two iterations strike a good compromise between visual
smoothness and computational cost, while still maintaining features such as berm
edges and rut depth that closely resemble the experimental terrain profiles.

The erosion propagation rings are chosen as $3$. This parameter controls how far
the bulldozing effect can extend beyond the immediate contact area. A ring
corresponds to one layer of neighboring nodes in the terrain grid. With three
rings, soil displaced under the robot can influence nodes up to three cells away
in each direction. This allows berms to grow laterally beyond the exact contact
footprint but prevents material from spreading unrealistically far. We selected
three rings based on visual comparison with experiment, ensuring that berm
widths and overall disturbed regions match the footprint of the robot and the
observed spread of sand.

\subsection{Robot Geometry and Contact Material}

The COBRA robot bodies and passive rollers are modeled in Chrono using convex
collision shapes that approximate the actual link geometry. The contact material
parameters are chosen to emulate the behavior of 3D printed polymer components
interacting with sand. The Young’s modulus of the robot material is set to
$E = 3.5\times10^9\,\text{Pa}$, which corresponds to a relatively stiff plastic.
The Poisson ratio $\nu = 0.36$ reflects the near-incompressibility typical of
many polymers. These two parameters determine the effective contact stiffness
when combined with the soil response.

The coefficient of friction between the robot and the soil is set to
$\mu = 0.45$. This value was selected such that the ratio of tangential to
normal forces obtained in simulation yields slip patterns comparable to those
seen on sand, without causing unrealistic sticking or excessive sliding. The
coefficient of restitution $e = 0.02$ is very small, which reflects the highly
inelastic nature of impacts when a heavy robot interacts with granular soil.
Kinetic energy is rapidly dissipated during contact, and the robot does not
bounce noticeably, as also observed in experiments.

Each link is assigned an effective contact area
\[
A_{\text{eff}} = 5.5\times10^{-3}\,\text{m}^2.
\]
This area acts as a representative footprint used when estimating the contact
patch width $b$ for the Bekker–Wong formulation. The value was chosen based on
the projected area of the link regions that most consistently interact with the
terrain during sidewinding, and it aligns the simulated sinkage depth with what
is measured in the physical tests.

\subsection{Motion Control and Gait Execution}

The robot model is imported into Chrono from a URDF description using
\texttt{ChParserURDF}. Joint motion is driven in position control mode using
precomputed sinusoidal trajectories stored in a time-indexed CSV file. This
ensures that the same gait parameters used in other simulation platforms and in
experiments are precisely reproduced.

Before locomotion begins, the simulation is run for a $2\,\text{s}$ settling
period with the gait either at zero amplitude or slowly ramped. This allows the
robot to come to rest on the deformable terrain, ensuring that initial
conditions do not contain spurious transients or unrealistically large
oscillations in sinkage. The alternating sign convention in the joint commands
enforces a lateral rolling constraint that replicates the desired sidewinding
pattern: neighboring joints are driven with phase offsets and sign changes so
that the body forms the characteristic sideways-traveling wave with vertical
and horizontal components.

\subsection{Numerical Solver and Visualization}

The Chrono simulations are integrated with a fixed time step of
$\Delta t = 0.001\,\text{s}$. This time step is small enough to resolve the
contact dynamics between the robot and the deformable terrain while remaining
large enough to keep overall simulation runtimes manageable. Contact
constraints are solved using a Barzilai–Borwein non-smooth iterative method,
with a maximum of $200$ solver iterations per time step. The Barzilai–Borwein
scheme accelerates convergence for the non-smooth complementarity problems that
arise in contact and friction, and the iteration cap prevents excessive
computation if convergence slows in difficult configurations.

Visualization is handled through the Vulkan Scene Graph backend. This backend
provides hardware-accelerated rendering suitable for detailed terrain meshes
and for visualizing node-level deformation and contact forces during locomotion.
During all simulations, full spatial wrenches for each body are logged at every
time step. These time series contain forces and torques at the body frame and
are later used for offline analysis of load distribution, traction patterns, and
energy transfer between the robot and the terrain.

Together, these empirical parameters and numerical settings have been validated
by comparing Chrono SCM results against experimental sidewinding data. The
configuration documented here defines the baseline used for all deformable
terrain simulations in this thesis and provides a reproducible reference for
future studies.

\section{Simulation Loop}
\label{sec:scm_sim_loop}

The Chrono--SCM simulation operates through a sequential update loop that advances the multibody dynamics, computes terrain deformation, and applies contact forces at each timestep. The pseudo code below summarizes the execution structure implemented in this work:

\begin{algorithm}[H]
\caption{Project Chrono SCM Simulation Loop}
\label{alg:scm_sim}
\begin{algorithmic}[1]

\State \textbf{Initialize Chrono system} and set gravity: $\mathbf{g} = (0, 0, -9.81)\,\text{m/s}^2$
\State Load robot multibody model from URDF using \texttt{ChParserURDF}
\State Assign material properties and contact geometry to all link bodies
\State Create SCM terrain with dimensions $(9 \times 8)\,\text{m}$ and grid resolution $\Delta=0.02\,\text{m}$
\State Set soil mechanical parameters $\{K_c, K_\phi, n, c, \phi, k, K, R\}$
\State Configure bulldozing model and terrain visualization options
\State Load joint trajectory input (time-indexed sinusoidal gait)
\State Initialize time $t \gets 0$

\While{$t < T_{\text{final}}$}
    \State Read current pose of each link body
    \State Apply joint position commands at time $t$ (position-controlled actuation)
    \State Perform ray-casting from terrain grid nodes to collision shapes
    \State Identify active contact patches and compute:
    \State \quad Sinkage $z$
    \State \quad Contact area $A$ and perimeter $L$
    \State \quad Normal stress $p$
    \State \quad Shear stress $\tau$
    \State Update terrain vertical deformation and bulldozing side flows
    \State Accumulate nodal contact forces and apply to rigid bodies
    \State Integrate multibody dynamics for timestep $\Delta t$
    \State Log contact forces, joint states, and body kinematics
    \If{visualization enabled}
        \State Update VSG render scene
    \EndIf
    \State $t \gets t + \Delta t$
\EndWhile

\State \textbf{Terminate} simulation and export datasets for analysis

\end{algorithmic}
\end{algorithm}

This loop ensures consistent coupling between deformable terrain response and sidewinding locomotion kinematics. Terrain deflection evolves only where contact occurs, while high-frequency wrench logging enables post-processing of distributed contact mechanics along the \ac{COBRA} body.

\section{Simulation Details}
\label{sec:chrono_scm_details}

\begin{figure}[ht]
    \centering
    \includegraphics[width=0.90\textwidth]{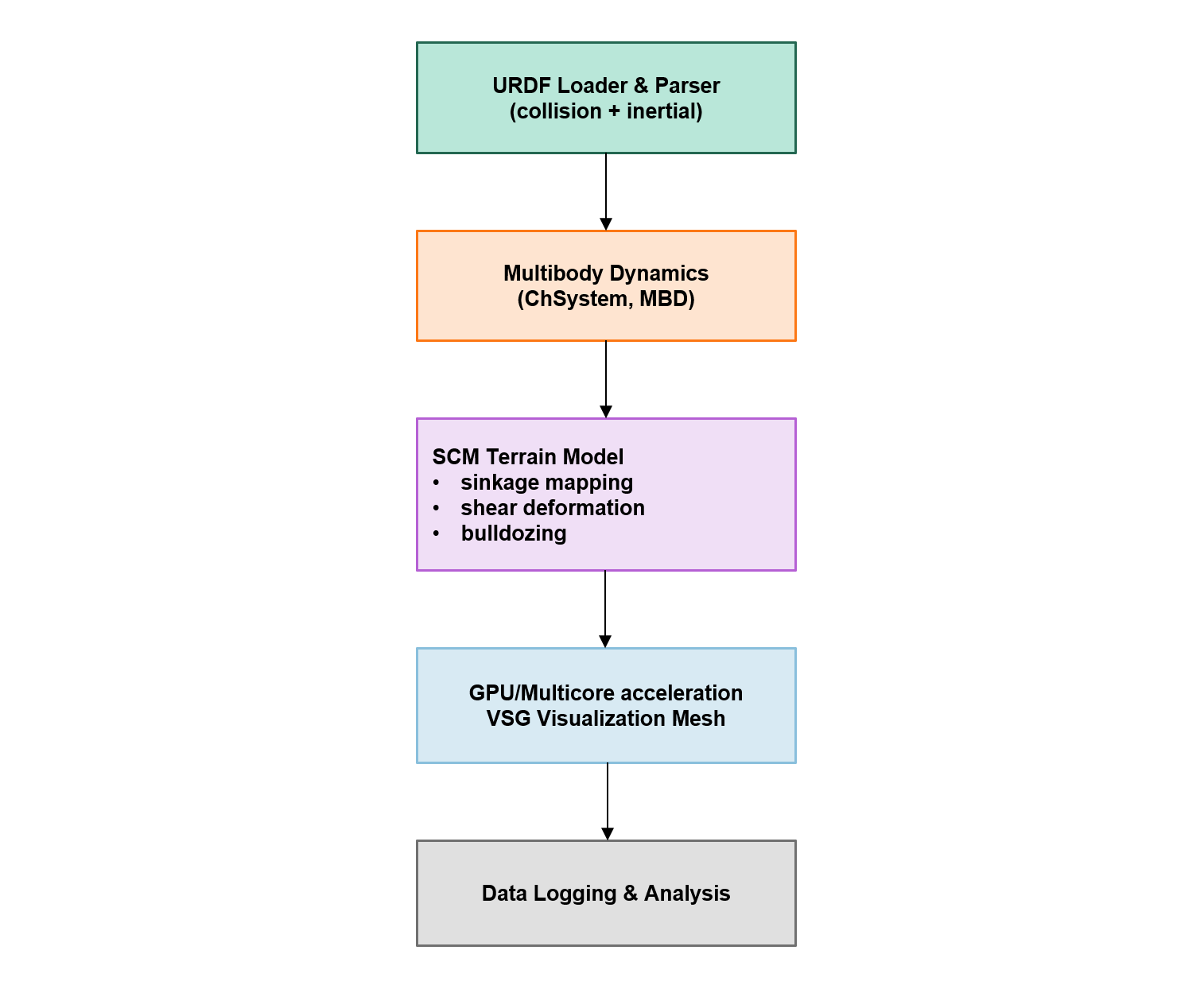}
    \caption{Software architecture of the Chrono-based SCM simulation environment for COBRA.}
    \label{fig:chrono_architecture}
\end{figure}

The Project Chrono SCM environment developed in this thesis integrates custom snake robot modeling, deformable soil mechanics, GPU-supported multibody simulation, and high-frequency contact logging. The overall architecture in Figure~\ref{fig:chrono_architecture} organizes these elements into clean modules that support consistent data flow and reproducible experiments.

\subsection{Core Components}

Chrono’s Multibody Dynamics engine computes rigid-body motion, joint constraints, and system updates at each timestep. Robot geometry, inertial values, and joint information are imported directly from the URDF file, ensuring consistency with ROS and Gazebo without maintaining separate geometry.

The SCM subsystem models deformable terrain by applying pressure–sinkage relations, shear deformation, bulldozing behavior, and damping effects. Terrain states such as sinkage and shear displacement are updated continuously and influence future contacts. GPU and multicore acceleration are used to handle the large number of terrain nodes. Visualization is provided by the Vulkan Scene Graph, which renders the robot and evolving terrain fields in real time.

\subsection{Required Input Assets}

The simulation uses the COBRA URDF model for link hierarchy, inertial tensors, actuator frames, and simplified convex collision meshes. Joint trajectories are loaded as time-indexed CSV files containing the sinusoidal commands used for sidewinding. SCM parameters specify the soil patch size, grid resolution, and soil constants such as pressure–sinkage coefficients, shear properties, bulldozing factors, and damping terms. These inputs define the mechanical environment experienced by the robot.

\subsection{Simulation Execution Pipeline}

The simulation begins with initialization of the multibody system, materials, and parsers. The robot is loaded from URDF, its joints are linked to the control subsystem, and collision shapes are added to the scene. The SCM grid is then created and assigned all soil and bulldozing parameters, which determine how displaced material spreads.

Logging channels are initialized, and the main loop advances the robot and terrain states together. SCM updates sinkage, shear deformation, and plastic displacement, while the solver computes forces and integrates the multibody motion. Outputs are written continuously to disk for later analysis.

\subsection{Logging and Output Data}

The system records link trajectories, orientations, and joint states at roughly five hundred hertz. SCM logs nodal forces, sinkage depths, shear displacement, and vertical deformation, creating a detailed record of terrain response. Additional metadata such as the number of active nodes, ray casting counts, and solver statistics is stored for diagnostics.

This modular design runs reliably on both local workstations and GPU clusters. All configuration files and build scripts used in this thesis are available through the SSLab GitHub repository, providing a stable platform for studying locomotion on deformable terrain using Chrono and SCM.

\section{Hardware Tests}
\label{sec:hardware_tests}

To evaluate the locomotion performance of the COBRA robot on real terrain,
two controlled sets of hardware experiments were conducted. The first involved
rigid laboratory flooring equipped with high-precision external motion capture,
while the second involved an outdoor loose-sand environment in which state
estimation relied entirely on onboard visual–inertial odometry. The objective
was to compare physical behavior against the predictions of the rigid-ground
Simulink model in Section~\ref{sec:simulink} and the deformable SCM
terrain simulations in Section~\ref{subsec:scm_terrain}. These experiments
provide the empirical basis needed to judge how well the simulation pipeline
reproduces both locomotion efficiency and terrain–robot interaction effects.

\begin{figure}[ht]
    \centering
    \includegraphics[width=0.85\linewidth]{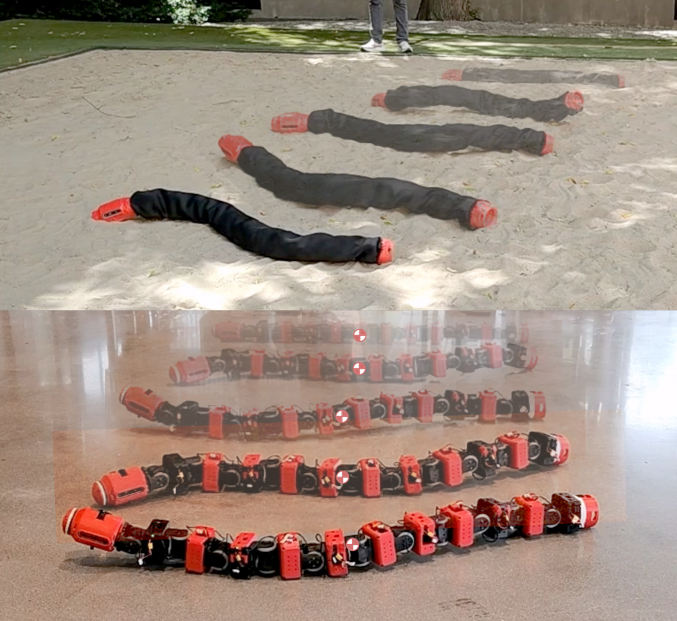}
    \caption{Snapshots of the snake robot operating in rigid laboratory
    conditions using motion capture and on loose outdoor sand using onboard
    visual–inertial odometry. Both setups are used to benchmark the predicted
    behavior from rigid and deformable terrain simulations. \textit{Image courtesy of Adarsh Salagame}. }
    \label{fig:exp-snapshots}
\end{figure}

\subsection{Laboratory Rigid-Ground Experiments}

The rigid-ground experiments were conducted inside the motion-capture space of
the Northeastern University Silicon Synapse Lab. The environment provides a
flat, hard floor with negligible compliance, making it suitable for validating
the rigid-ground assumptions used in the Simulink model. A twelve-camera
OptiTrack system, sampling at one hundred and twenty hertz, tracked the robot’s
head link using reflective markers. This setup supplies highly accurate
six-degree-of-freedom ground-truth trajectories and enables direct comparison
between measured and simulated forward progression, lateral drift, and body
heading changes.

The commanded joint motions were identical to those used in simulation. They
were produced from the same sinusoidal trajectory generator and executed by the
robot’s onboard motor drivers. Each experimental condition was repeated multiple
times to mitigate motion-capture dropouts, brief occlusions, or minor variations
in actuator response. Repeated trials help ensure that deviations between
simulation and experiment reflect modeling limitations rather than random
measurement noise.

\subsection{Loose-Sand Experiments with Onboard VIO}

The loose-sand tests were performed in an outdoor environment consisting of dry,
unconsolidated sand with grain sizes in the approximate range of zero point one
to one millimeter. Such terrain is continuously reshaped by the robot’s motion,
producing sinkage, shear trenches, berm build-up, and other topographical
changes that make external markers or fiducial-based tracking unreliable.
Further, a laboratory-style motion-capture system cannot be deployed outdoors
under these conditions.

For this reason, trajectory estimation was achieved using onboard monocular
visual–inertial odometry based on ORB-SLAM3 \cite{campos_orb-slam3_2021}.
A RealSense D435i camera supplied synchronized RGB frames and inertial
measurements at thirty hertz. The system constructs a locally consistent
trajectory by combining image features and accelerometer readings with a state
estimator that manages drift over short time windows. This method provides
reliable motion estimates when visual features remain stable in the field of
view and when the robot’s motion does not produce excessive blur.

\begin{figure}[ht]
    \centering
    \includegraphics[width=0.85\linewidth]{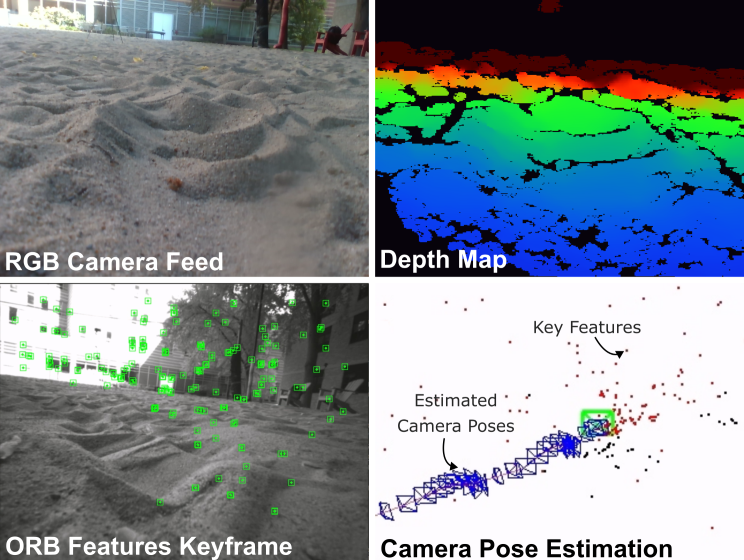}
    \caption{Onboard sensing during loose-sand tests. The RGB stream, depth
    reconstruction, and the visual–inertial odometry pose estimation illustrate
    how robot motion is tracked in environments where external motion capture is
    not feasible. \textit{Image courtesy of Adarsh Salagame}. }
    \label{fig:camera}
\end{figure}

Visual–inertial odometry performance was sensitive to image quality, terrain
appearance, and the robot’s instantaneous motion. Large roll or pitch movements
increased motion blur and could reduce the number of trackable features in the
scene. In addition, self-occlusion from nearby links occasionally disrupted the
feature-tracking pipeline. Trials in which the estimator drifted excessively or
lost track altogether were removed from the dataset to prevent corrupted
trajectories from influencing the comparison with simulated predictions.

\subsection{Gaits and Test Conditions}

Four combinations of gait and frequency were commanded across the two
environments. These correspond to the two gait shape parameters used throughout
this work, each executed at two distinct frequencies:
\[
\text{Gait 1: } \{0.4,\, 0.2\}\,\text{Hz}
\qquad
\text{Gait 2: } \{0.3,\, 0.1\}\,\text{Hz}.
\]
These frequencies were chosen by balancing two competing needs. Higher
frequencies promote faster forward travel and more pronounced sidewinding
deformations, increasing the stress applied to the soil. Lower frequencies
increase the likelihood of stable visual–inertial odometry performance because
image blur is reduced and features remain more easily trackable between frames.
The lowest-frequency case of Gait 2 at zero point one hertz consistently
produced unstable odometry, often leading to rapid drift; such trials were
excluded from the final analysis.

\subsection{Dataset Summary}

In each environment, a consistent suite of measurements was gathered to support
comparison with simulation. For the rigid laboratory surface, the head-link
position and orientation were recorded throughout each trial. Commanded joint
trajectories and encoder-reported joint measurements were stored in parallel to
diagnose discrepancies between intended and realized motion. In the loose-sand
environment, the onboard visual–inertial odometry provided relative motion
estimates, which were stored together with joint data and timestamps. In both
environments, qualitative observations were documented, including visible
sinkage, slip patterns, link–soil interaction, and notable terrain disturbances.
These records also include notes about visual–inertial odometry reliability in
order to distinguish between estimator degradation and genuine differences in
locomotion behavior.

The combined hardware dataset serves as both a validation reference for the
simulation models and an empirical foundation for understanding how gait
amplitude, frequency, and terrain compliance shape the robot’s overall
locomotion performance.

\section{Tumbling Gait Analysis}
\label{sec:tumbling_intro}

Tumbling is a high-energy locomotion mode used by the COBRA robot to descend
steep slopes where sidewinding becomes ineffective or unstable. In this mode,
the robot mechanically links its head and tail to form a closed-loop barrel
configuration. This shape converts the body into a protective rolling shell
that shields sensitive components and allows the vehicle to descend slopes
primarily through gravitational potential energy rather than through
precision-controlled shape changes. The closed-loop structure stabilizes the
center of mass inside the body cavity and distributes external impacts over the
outer links, making tumbling particularly well suited for rapid descent on
slippery, steep, or yielding terrain.

To understand the physics of downhill tumbling, this thesis uses two modeling
approaches that complement one another. The first is a rigid-ground model built
in MATLAB Simulink. This representation isolates the pure rotational dynamics
of the barrel-like structure, allowing analysis of angular acceleration, normal
impact forces, and slip-induced motion when the robot rolls over a hard
inclined plane. By neglecting soil deformation, this model emphasizes the
effects of gravity, contact stiffness, and the placement of the center of mass.
It provides a lightweight environment in which impacts, rebound behavior, and
momentum transfer can be studied without the numerical complexity of
deformable media.

The second modeling approach uses the Chrono DEM Engine to simulate tumbling on
granular terrain. DEM represents the soil as an ensemble of discrete particles
that interact through collision laws, friction, and cohesive or cohesionless
force models. During tumbling, each body link interacts with hundreds or
thousands of particles, causing localized sinkage, particle rearrangement,
drag, and mass-wasting effects. Soil yields beneath the robot, redistributes
under its weight, and may even avalanche downslope depending on slope angle,
grain size, and local compaction. This method captures the detailed mechanics
that cannot be approximated with rigid-surface or simple compliance models,
including complex stick–slip transitions, intermittent burial of the rolling
shell, and velocity-dependent drag forces generated by particle impacts.

\begin{figure}[ht]
    \centering
    \includegraphics[width=0.80\linewidth]{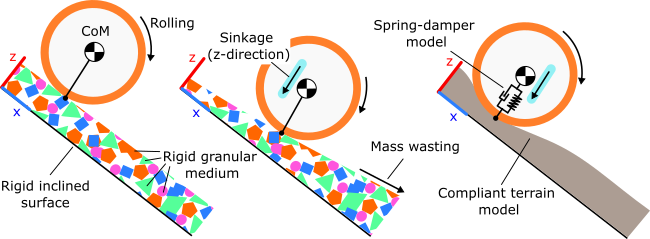}
    \caption{Comparison of terrain–interaction models during downhill tumbling.
    Left: Rigid inclined surface dominated by rotational inertia and hard impact.
    Center: Granular medium with depth-dependent sinkage and slip.
    Right: Compliant deformable material exhibiting spring–damper response and
    terrain reshaping due to mass wasting. \textit{Image courtesy of Adarsh Salagame}. \cite{salagame_crater_2025}}
    \label{fig:interaction_downhill}
\end{figure}

The rigid-ground and granular-terrain models together provide a validation
framework that spans simple to highly detailed physical representations. The
rigid model reveals how tumbling initiates, how body shape influences
rotational stability, and how gravity and impact govern descent speed. The DEM
model extends this analysis to real soils, uncovering the influence of particle
size, compaction, yield strength, and depth-dependent resistance on overall
mobility. By comparing results across the two approaches, it becomes possible
to separate the roles of pure rigid-body dynamics from terrain-mediated effects
such as sinkage, drag, and energy dissipation through soil deformation.

This combined analysis enables high-fidelity assessment of downhill mobility
for snake robots and provides the physical foundation needed to evaluate their
suitability for future off-road or planetary exploration scenarios, where
complex terrain and unreliable footing can restrict the effectiveness of more
traditional gaits.

\section{MATLAB Simulink Tumbling Model for Rigid Ground}
\label{sec:simulink_tumbling}

To establish a fast and interpretable reference for the COBRA robot’s rolling
locomotion, a rigid-ground tumbling simulation was developed in MATLAB Simscape
Multibody. This model provides a simplified environment in which tumbling
dynamics can be analyzed without the additional complexity of soil deformation.
It allows direct examination of how the robot’s geometry, mass distribution, and
contact forces govern its rotational motion during downhill travel. The model
functions as a counterpart to the granular simulations presented later in
Section~\ref{sec:chrono_dem_intro}, and together the two frameworks separate
pure rigid-body effects from terrain-mediated phenomena.

In this Simulink setting, the robot is configured into its circular barrel
shape, produced by mechanically linking the head and tail modules. This
closed-loop configuration relocates the center of mass toward the interior and
distributes impacts over the outer surface of the body. The terrain is modeled
as a perfectly rigid plane, meaning no soil deformation, sinkage, or granular
flow occurs during contact. Instead, forces arise solely from elastic and
dissipative interactions between the shell of the robot and the rigid surface.
These interactions are captured through a nonlinear spring–damper formulation in
the normal direction and a velocity-dependent Coulomb friction law in the
tangential direction. The same mathematical structure appears in the rigid
contact model used for sidewinding in Section~\ref{sec:simulink}, ensuring
consistency between the two locomotion modes.

Normal forces are governed by a spring–damper contact law of the form
\[
f_n = s(d,w)\,(k\,d + b\,\dot d),
\]
in which the penetration depth \(d\) and its time derivative \(\dot d\) are used
to compute the instantaneous force transmitted through the ground. The constant
\(k\) represents the normal stiffness of the contact, which determines how
strongly the robot resists penetration. Higher stiffness values cause the
contact to behave more like a hard collision, leading to more abrupt changes in
angular velocity during tumbling. The coefficient \(b\) provides dissipation
that converts impact energy into heat, reducing rebound and smoothing the
transient response. The smoothing function \(s(d,w)\), which transitions between
non-contact and contact states over a characteristic width \(w\), ensures that
the normal force evolves continuously as the robot rolls, preventing numerical
instabilities associated with sharp contact transitions.

Tangential forces follow a velocity-dependent Coulomb friction law,
\[
\|f_t\| = \mu_{\text{eff}}(u_t)\,\|f_n\|,
\]
where \(u_t\) is the relative slip velocity at the contact point. The effective
friction coefficient \(\mu_{\text{eff}}(u_t)\) interpolates smoothly between
static and dynamic friction. When the tangential slip is small, the contact acts
in a nearly sticking regime with \(\mu_{\text{eff}} \approx \mu_s\). As slip
increases, friction decreases toward the dynamic coefficient \(\mu_d\). This
transition is important for modeling the interplay between sticking, slipping,
and momentary micro-impacts that occur when the barrel rotates over its contact
edge. The ability to capture these slip transitions is essential for predicting
how rotation accelerates or decays during descent.

The rigid-ground Simulink model therefore provides a computationally efficient
framework for evaluating tumbling behavior when terrain does not deform. By
removing soil mechanics, the model exposes the influence of rigid-body inertia,
geometry, and contact parameters on the rolling trajectory. It predicts nominal
rolling speed, angular acceleration, and the stability of the descent path when
the robot travels over firm ground. These predictions establish a baseline that
can be directly compared with the more complex responses observed in granular
terrain.

The outputs of the model include time histories of angular velocity, forward
translational velocity, and the normal and tangential contact forces that govern
energy exchange during impacts. The rolling trajectory on inclined rigid
surfaces is recorded to assess whether the robot maintains a stable descent or
exhibits lateral drift. These results provide a clear reference for evaluating
the degree to which granular processes, such as sinkage and particle
rearrangement, alter the expected motion.

In summary, the rigid-ground Simscape tumbling model acts as a fast predictive
tool for rolling locomotion on hard terrain. Granular DEM simulations extend the
same gait analysis to realistic soil conditions by adding mass wasting,
depth-dependent drag, and particle-scale deformation effects. Together, these
models form the comparative foundation used throughout the thesis to understand
the mechanics of tumbling on rigid and deformable terrain.

\section{Introduction to Project Chrono DEM-Engine}
\label{sec:chrono_dem_intro}

To investigate the highly dynamic interaction between the COBRA robot and loose
granular terrain during tumbling, this thesis incorporates simulations based on
the Discrete Element Method (DEM). DEM provides a particle-level representation
of soil mechanics and allows the terrain to deform, collapse, compact, and
redistribute in response to robot motion. Although the primary emphasis of this
thesis is sidewinding on deformable but continuum-style terrain models presented
earlier in Sections~\ref{subsec:scm_terrain} through~\ref{sec:chrono_scm_details},
analysis of tumbling in granular media is included due to its strong relevance
to understanding contact mechanics under high energy, large deformation, and
partially submerged conditions.

Prior to engaging with the granular model, a rigid-ground baseline for tumbling
was established using MATLAB Simulink, as described in
Section~\ref{sec:simulink_tumbling}. This baseline isolates the geometric and
inertial effects of tumbling on hard slopes and provides a clean reference
against which the additional challenges posed by loose soil may be interpreted.
These challenges include increased slip as particles rearrange beneath the
rolling shell, deeper sinkage that changes the effective rolling radius, and
momentum loss due to drag forces generated by particle impacts.

The Chrono DEM-Engine is a GPU-accelerated granular simulation system designed
for high-throughput computation with large particle counts
\cite{maurel_chronodeme_2023}. In DEM, the soil is represented as a collection
of discrete spherical or polyhedral particles that interact through explicit
contact laws. These interactions include normal and tangential collision forces,
frictional resistance, cohesive behavior if specified, and velocity-dependent
damping. Because the method resolves particle contacts directly rather than
through continuum approximations, DEM naturally captures phenomena such as
localized shear planes, non-uniform sinkage, intermittent yielding, and
particle flow that forms around bodies undergoing rapid motion.

This level of modeling detail is essential for tumbling because the robot
operates in a regime where its outer shell repeatedly strikes, compresses, and
shears through layers of granular material. As the closed-loop body descends a
slope, particles are displaced both vertically and laterally, forming temporary
force chains that bear load until they collapse under stress. These force chains
strongly influence stability by either supporting the robot briefly or allowing
rapid loss of support when they fail. DEM is capable of reproducing these
effects because each grain contributes individually to the evolving stress
network within the soil.

Further, DEM captures depth-dependent reaction forces that arise when parts of
the robot become partially buried during tumbling. The resistance experienced by
the shell depends on the number, depth, and compaction of particles in contact,
all of which evolve dynamically as the body rolls. The rigid-ground model cannot
reproduce these effects, since it assumes uniform contact with no penetration or
granular flow. DEM therefore serves as the high-fidelity tool required to
evaluate how loose sand modifies the descent trajectory, reduces overall
momentum, and shapes the stability of rolling on steep inclines.

Including the DEM analysis in this thesis provides a complementary perspective
on terrain–robot interaction. While the SCM model describes soft soil behavior
using continuum-inspired pressure–sinkage and shear laws appropriate for
sidewinding, DEM extends the modeling fidelity to situations involving deep
penetration, large soil displacements, and rapid particle rearrangement. This
multi-resolution approach, combining rigid, compliant, and particle-resolved
terrain models, supports a holistic understanding of COBRA’s mobility across its
full range of gaits.

Validated DEM simulation results are summarized in
Chapter~\ref{chap:results}, where they are compared with experimental behavior
to assess how well particle-level terrain modeling reproduces the contact forces
and motion characteristics observed during physical tumbling trials.

\section{DEM Terrain Workflow}
\label{sec:chrono_dem_workflow}

To support the contact modeling perspective of this thesis, this section
presents the GPU-accelerated Discrete Element Method (DEM) simulations used to
analyze tumbling locomotion in granular terrain. These simulations were
originally developed and validated in the AISJ tumbling study \cite{salagame_crater_2025}, and although no
new DEM computations were generated specifically for this thesis, the results
remain essential because tumbling involves deep penetration, rapid momentum
exchange, and large soil displacements that cannot be approximated by any
continuum model such as SCM. DEM’s particle-resolved formulation is uniquely
capable of capturing these behaviors, making it the correct tool for analyzing
high-energy rolling motion in loose sand.

DEM models the terrain as an ensemble of discrete grains whose positions,
velocities, and contact forces evolve according to history-dependent interaction
laws. This representation is critical for tumbling, because rolling motion
causes soil to compact, dilate, shear, and collapse beneath the robot, forming
force chains that intermittently support the body before failing under load. As
the robot rotates downslope, particles are displaced both vertically and
laterally, creating drag forces, sinkage gradients, and slip episodes that
strongly influence stability and descent speed.

\begin{figure}[ht]
    \centering
    \includegraphics[width=0.95\textwidth]{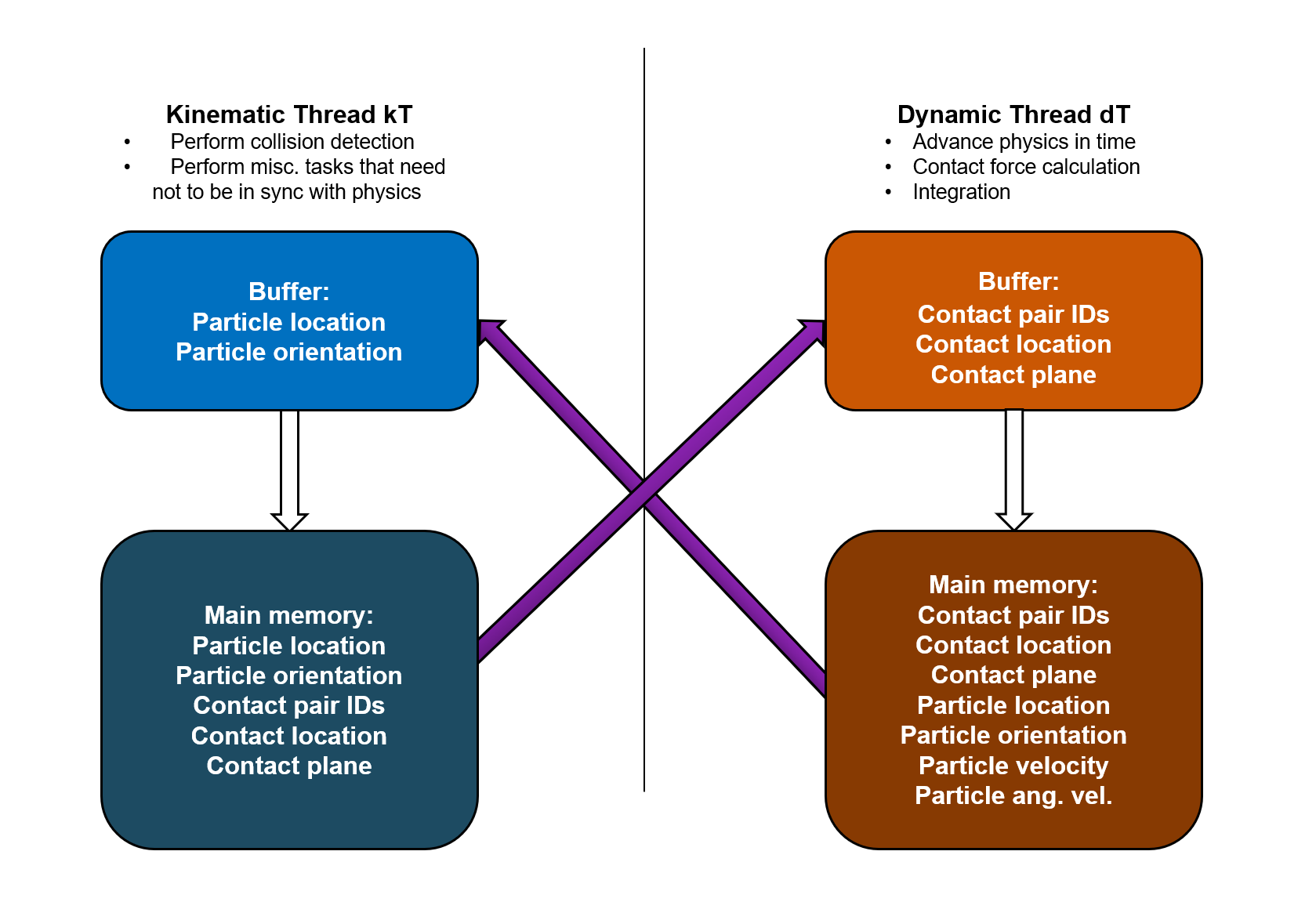}
    \caption{Conceptual dual-thread GPU architecture used by the DEM Engine.
    The Kinematic Thread supplies robot geometry and motion, while the Dynamic
    Thread resolves particle contacts and integrates the granular system.
    Inspired by \cite{maurel_chronodeme_2023}.}
    \label{fig:dem_dual_threads}
\end{figure}

The DEM Engine used in this work implements a dual-thread GPU architecture that
separates geometry updates from granular physics evaluation. Two asynchronous
threads operate in parallel: the Kinematic Thread (kT) and the Dynamic Thread
(dT). The kT is responsible for updating the robot’s rigid-body motion and
feeding the geometry into the granular solver, while the dT handles particle
collisions, contact forces, and time integration. A conceptual overview of this
architecture is shown in Figure~\ref{fig:dem_dual_threads}. The separation of
these responsibilities increases throughput and allows the simulation to handle
millions of particles efficiently, which is essential for capturing deep
penetration and large soil deformation during tumbling.

\subsection{Simulation Workflow Overview}

\begin{figure}[ht]
    \centering
    \includegraphics[width=0.85\textwidth]{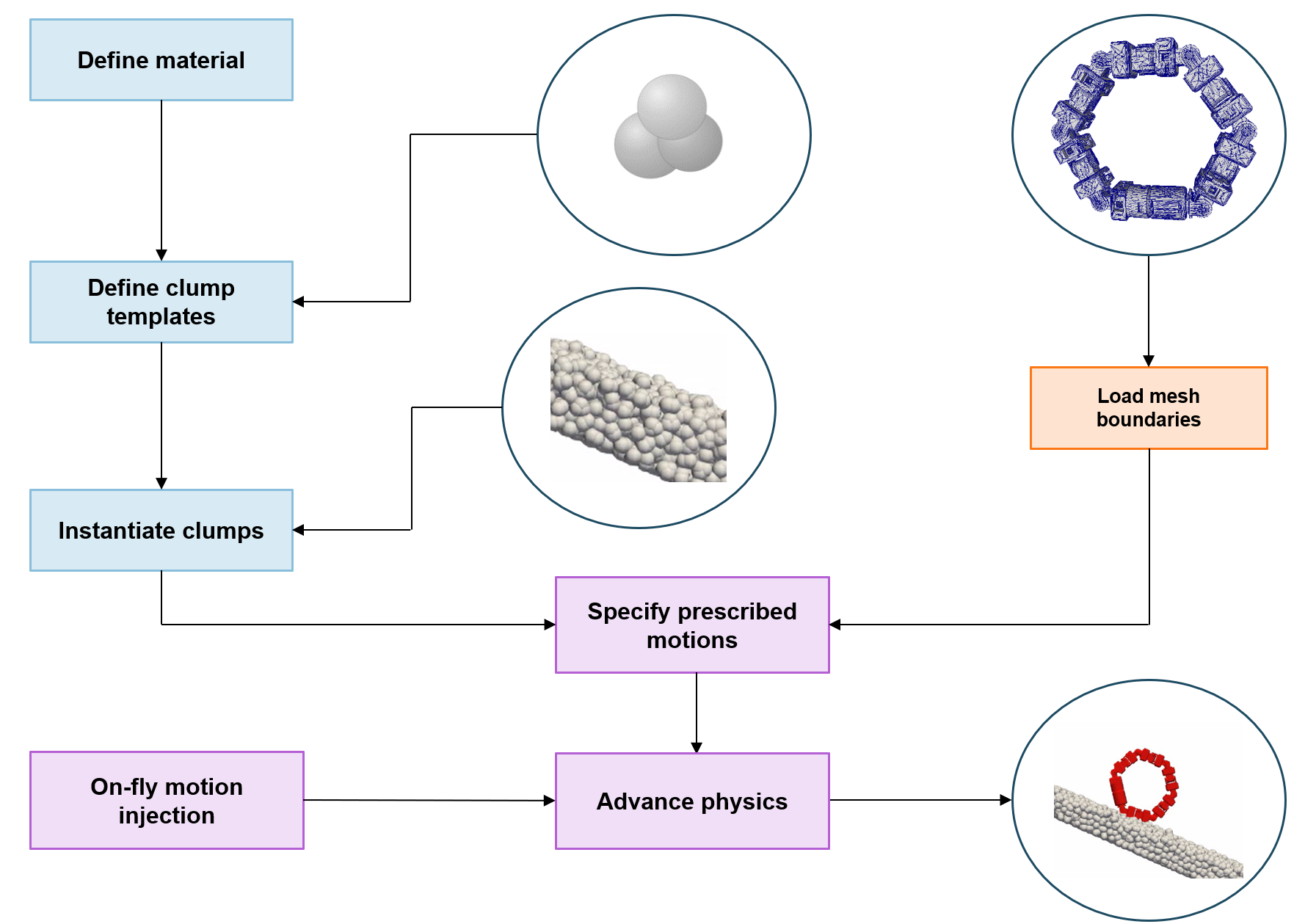}
    \caption{Conceptual workflow of a DEM simulation including grain generation,
    clump instantiation, boundary settling, and motion prescription.
    Adapted from \cite{maurel_chronodeme_2023}.}
    \label{fig:dem_workflow}
\end{figure}

The DEM simulations used in the AISJ study \cite{salagame_crater_2025} follow a structured workflow that
separates material definition, terrain generation, and robot motion input. This
pipeline, illustrated in Figure~\ref{fig:dem_workflow}, proceeds in several
stages. First, the material properties of the granular medium are defined,
including particle density, Young’s modulus, Poisson ratio, and friction. These
parameters control how stiff or compliant grain–grain collisions are and govern
the angle of repose of the resulting granular bed. A Young’s modulus that is too
large causes excessively rigid collisions and unrealistic bouncing. A modulus
that is too small produces overly soft grains that absorb too much energy and
flatten under load. Poisson ratio influences the lateral expansion of particles
under compression and therefore affects packing behavior and bulk density.
Friction determines the resistance to sliding between grains and directly
controls the angle of repose and shear stability of the slope.

Next, clump templates are created. A clump is a rigid aggregate of spheres that
moves as a single body. Clumps capture the irregularity of real sand grains
better than single spheres do. Real grains are angular and interlock under load,
and clumps approximate this behavior by presenting multiple overlapping contact
patches. Using clumps increases computational cost slightly but dramatically
improves bulk soil behavior, especially during shear and slope failure.

The granular domain is then populated by randomly depositing clumps into a
virtual container and allowing them to settle under gravity. This procedure
generates a realistic sand bed whose density and porosity depend on friction,
grain shape, and the deposition protocol. Once settled, the container boundaries
are removed or reconfigured to form the experimental terrain.

The robot geometry is imported next. The COBRA rolling configuration is added as
a collision shape. Because the tumbling motion is prescribed in these
simulations, actuation is not necessary. Instead, a motion input is applied at
the head link to rotate the connected circular structure. This prescribed motion
ensures that the robot follows a consistent tumbling pattern and allows direct
comparison across different soil conditions.

As the simulation proceeds, the kinematic thread updates the robot’s position and
orientation according to the prescribed tumbling motion. The dynamic thread
computes particle contacts and integrates all grain trajectories. Forces arising
from grain impacts, drag, sinkage, and mass redistribution are recorded and
mapped onto the robot. Output particle trajectories, grain velocities,
collisions, and reaction forces form the basis for later analysis.

\subsection{Particle–Particle Contact Mechanics}

\begin{figure}[ht]
    \centering
    \includegraphics[width=0.75\textwidth]{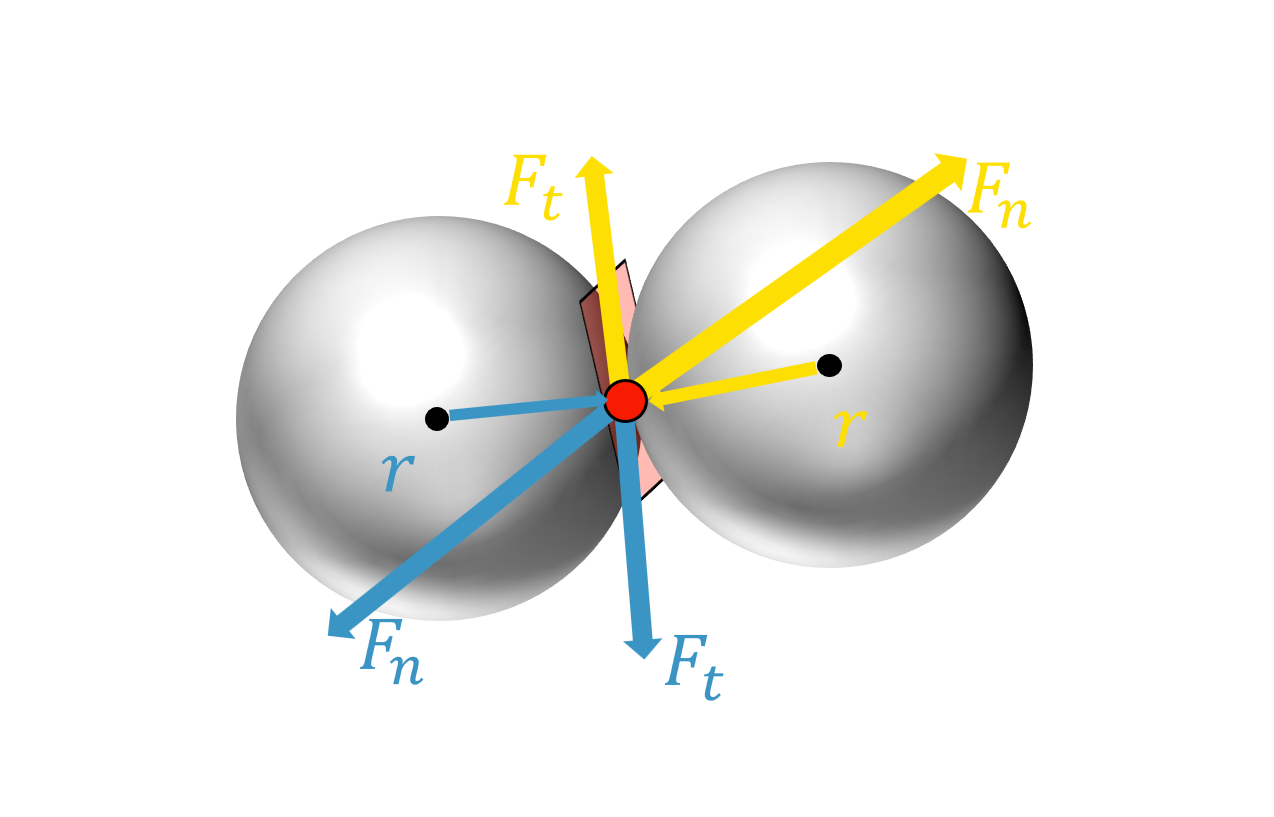}
    \caption{Conceptual representation of DEM particle contact mechanics,
    showing normal and tangential components of the interaction. Inspired by
    \cite{maurel_chronodeme_2023}.}
    \label{fig:dem_contact_forces}
\end{figure}

DEM represents the interaction between particles using nonlinear contact laws in
both the normal and tangential directions. Figure~\ref{fig:dem_contact_forces}
illustrates the conceptual model. When two particles overlap, the penetration
depth \(\delta_n\) represents how far their surfaces intersect. This overlap is
a numerical construct used to compute elastic and damping forces. The normal
force is given by
\[
F_n = k_n \delta_n - d_n \dot{\delta}_n,
\]
where \(k_n\) is the normal stiffness and determines how strongly particles push
back as they compress. A larger stiffness produces sharper collisions and higher
energy transfer. The damping parameter \(d_n\) dissipates kinetic energy and
reduces rebound, ensuring that particles do not oscillate unrealistically during
settling or impacts.

Tangential forces arise from shear displacement. The DEM engine maintains a
history variable \(j\), which stores tangential slip accumulated over the
contact duration. The tangential force is computed as
\[
F_t = \min(\mu F_n,\; k_t j),
\]
where \(k_t\) is the tangential stiffness and determines how shear resistance
builds up as grains attempt to slide. The friction coefficient \(\mu\) provides a
Coulomb limit. When \(k_t j < \mu F_n\), grains stick and shear elastically. When
the threshold is exceeded, grains slide, producing plastic shear and rearranging
the granular structure. This formulation is essential for modeling slope
collapse, intermittent jamming, and the highly discontinuous nature of rolling
motion through sand.

\subsection{DEM Clump Types}

Chrono DEM supports different clump styles for representing grains. Single-sphere
clumps compute quickly but unrealistically smooth out shear resistance and
underestimate angle-of-repose behavior. Irregular multi-sphere clumps, used in
the AISJ work \cite{salagame_crater_2025}, offer a closer approximation to natural sand grains by producing
multiple contact surfaces and greater interlocking. More advanced grain
templates based on measured sand geometry have also been studied in
\cite{maurel_chronodeme_2023}. For the purposes of the tumbling analysis,
three-sphere clump templates were chosen because they balance computational
efficiency with realistic bulk shear strength and sinkage properties. These
templates were tuned so that the resulting granular bed matches observed sinkage
depths and resistance forces encountered during hardware tumbling tests of the
COBRA robot.

Although the DEM results used in this thesis originate from earlier work rather
than newly executed simulations, their inclusion is essential because they
provide detailed insight into extreme granular interactions that complement the
SCM-based sidewinding analysis. Together, SCM and DEM demonstrate how soil
mechanics influence the spectrum of COBRA’s mobility modes, from smooth,
low-energy sidewinding on soft terrain to vigorous rolling through deep,
yielding sand.

\section{DEM Terrain Mathematical Formulation}
\label{sec:chrono_dem_formulation}

The granular terrain model used in this thesis follows the same Discrete Element
Method (DEM) formulation that was developed for the COBRA tumbling study. It is
included here because the particle-resolved physics directly support the contact
modeling viewpoint of this work, particularly in situations involving deep
penetration, rolling impacts, and nonlinear soil resistance. Although no new DEM
experiments were generated specifically for this thesis, the validated
formulation is reproduced here for clarity and reproducibility.

DEM treats the soil as a collection of discrete particles whose interactions are
resolved using history-dependent contact mechanics. Each particle may represent
a single grain or a clump of spheres, and its motion evolves according to
Newton’s laws subject to contact forces, damping, and friction. This approach is
crucial for modeling tumbling, where the robot rolls into and through loose
sand, causing grains to rearrange, compact, and flow in ways that continuum
models such as SCM cannot capture. DEM naturally reproduces jamming, force-chain
formation, dilatancy, and yielding, all of which strongly influence tumbling
behavior.

\paragraph{Normal Contact Force.}

Consider two spheres with radii \(r_i\) and \(r_j\), positions
\(\mathbf{x}_i\) and \(\mathbf{x}_j\), and relative velocity
\(\mathbf{v}_{ij}\). The unit normal vector at the point of contact is

\[
\hat{\mathbf{n}} = \frac{\mathbf{x}_i - \mathbf{x}_j}
{\|\mathbf{x}_i - \mathbf{x}_j\|},
\]

and the normal overlap, or penetration depth, is

\[
\delta_n = r_i + r_j - \|\mathbf{x}_i - \mathbf{x}_j\|.
\]

The overlap \(\delta_n\) is not a physical interpenetration but a numerical
device used to compute the elastic repulsion between grains. When
\(\delta_n > 0\), a force in the normal direction is applied:

\begin{equation}
\label{eq:dem_normal_update}
\mathbf{F}_n =
k_n\,\delta_n\,\hat{\mathbf{n}}
- d_n\,(\hat{\mathbf{n}} \cdot \mathbf{v}_{ij})\,\hat{\mathbf{n}}
\end{equation}

where \(k_n\) is the normal stiffness and determines how strongly grains
push back against compression. Higher stiffness produces sharper, more
impact-like collisions and increases the effective bulk stiffness of the granular
bed. The damping coefficient \(d_n\) controls the rate at which kinetic energy is
lost during collisions. A sufficiently large \(d_n\) prevents unrealistic bouncing
and ensures stable settling of particles under gravity.

The quantity \(\hat{\mathbf{n}} \cdot \mathbf{v}_{ij}\) represents the approach
speed between particles along the line of action. The damping term opposes this
approach velocity, absorbing energy and dissipating oscillations. Together, the
elastic and damping components define the normal repulsive behavior and play a
critical role in the formation and collapse of force chains beneath the robot.

\paragraph{Tangential Contact Force with Shear Accumulation.}

Tangential forces are derived from the history of shear deformation. The DEM
engine maintains a shear displacement vector \(\boldsymbol{\xi}\) at every active
contact, which evolves according to the tangential slip velocity:

\[
\dot{\boldsymbol{\xi}} = \mathbf{v}_t, \qquad
\mathbf{v}_t = \mathbf{v}_{ij}
- (\hat{\mathbf{n}} \cdot \mathbf{v}_{ij})\,\hat{\mathbf{n}}.
\]

The vector \(\mathbf{v}_t\) captures the sliding motion that occurs as particles
move past each other, and \(\boldsymbol{\xi}\) accumulates over time, representing
how far the grains have sheared since contact was initiated. This shear history
enables DEM to capture displacement-dependent stick–slip transitions and
nonlinear tangential resistance.

The proposed tangential force before friction limiting is

\[
\mathbf{F}_{t}^{\text{prop}} = -k_t\,\boldsymbol{\xi} - d_t\,\mathbf{v}_t,
\]

where \(k_t\) is the tangential stiffness. This parameter determines how
rapidly shear resistance builds up as grains resist sliding. The tangential
damping \(d_t\) dissipates energy associated with lateral motion and prevents
high-frequency oscillations. Together, these terms model the elastic and viscous
components of shear interaction.

The actual tangential force must satisfy Coulomb friction:

\begin{equation}
\label{eq:dem_friction_constraint}
\mathbf{F}_t =
\begin{cases}
\mathbf{F}_t^{\text{prop}}, &
\|\mathbf{F}_t^{\text{prop}}\| \le \mu\,\|\mathbf{F}_n\|,
\\[3pt]
-\mu\,\|\mathbf{F}_n\|\,
\dfrac{\mathbf{v}_t}{\|\mathbf{v}_t\| + \epsilon_v}, &
\text{otherwise},
\end{cases}
\end{equation}

where \(\mu\) is the friction coefficient, controlling the maximum tangential
force before sliding occurs. The small constant \(\epsilon_v\) ensures numerical
stability when \(\mathbf{v}_t\) is close to zero. When
\(\|\mathbf{F}_t^{\text{prop}}\| < \mu \|\mathbf{F}_n\|\), particles remain in a
sticking regime and move together elastically. When the bound is exceeded,
particles enter sliding, producing granular flow, shear bands, and avalanche-type
behavior. These transitions are fundamental to tumbling, where rolling motion
induces rapid shear stresses that propagate through the terrain.

\paragraph{Clump-Based Terrain Representation.}

Each DEM particle is modeled as a rigid clump composed of overlapping spheres.
This representation allows the grains to exhibit angularity, local interlocking,
and realistic dilatancy without the cost of polygonal mesh collisions. Clumps
maintain a fixed internal structure, while individual sphere–sphere interactions
between clumps generate distributed contact forces and torques. As the robot
rolls, these interactions create complex load paths and transient support
patterns. Clumps also enhance the angle-of-repose behavior of the granular bed,
ensuring that slopes collapse and reform in ways consistent with real sand.

Clump-based terrain is particularly important for modeling tumbling, because the
robot experiences momentary partial burial and must push through grains that
compact and shear around its body. The ability to represent force chains and
localized yielding enables DEM to capture the intricate terrain responses that
arise during high-energy, multi-contact rolling motion.

\vspace{6pt}
This DEM formulation underlies all tumbling simulations referenced in this
thesis. Although the simulations originate from prior published work, they are
included here because the particle-level mechanics they capture provide
high-fidelity insight into soil–robot interactions that cannot be reproduced
using continuum approximations alone.

\section{Empirical Parameters from Simulation Code}
\label{sec:dem_params}

The DEM-based tumbling parameters presented here correspond exactly to the
validated configuration used in the previously published COBRA tumbling study.
Although no new granular simulations were produced specifically for this thesis,
the same parameter set is reproduced to maintain transparency, enable
reproducibility, and support the broader contact--interaction analysis developed
throughout this work. These parameters govern particle stiffness, contact
damping, friction, wheel dynamics, terrain generation, and solver behavior, all
of which collectively determine the fidelity of the DEM environment.

\subsection{Material Parameters}

Two material classes are used: one to represent the robot link (approximated as
a rigid wheel during tumbling) and one to describe the granular bed. These
parameters regulate how stiff the interacting bodies are, how much energy is
lost during collisions, and how strongly particles resist sliding.

\paragraph{Wheel Material.}
\[
E = 1\times10^{9}~\text{Pa},\quad
\nu = 0.3,\quad
\mathrm{CoR} = 0.4,\quad
\mu = 0.6,\quad
C_{rr} = 0.02
\]

Here, \(E\) is the Young’s modulus providing normal stiffness of the wheel,
\(\nu\) is the Poisson ratio determining lateral contraction under load,
\(\mathrm{CoR}\) is the coefficient of restitution governing post-impact energy
retention, \(\mu\) is the friction coefficient controlling shear resistance, and
\(C_{rr}\) is the rolling--resistance coefficient modeling micro-slip and
internal losses during rotation.

\paragraph{Terrain Material.}
\[
E = 1\times10^{8}~\text{Pa},\quad
\nu = 0.3,\quad
\mathrm{CoR} = 0.1,\quad
\mu = 0.67,\quad
C_{rr} = 0.05
\]

For the granular bed, \(E\) is softer to represent yielding regolith,
\(\mathrm{CoR}=0.1\) reflects highly dissipative grain impacts, \(\mu=0.67\)
matches typical dry sand friction, and the larger \(C_{rr}\) captures energy
lost to grain rearrangement and void collapse.

Pairwise wheel--terrain interaction parameters are
\[
\mu_{\text{wheel/terrain}} = 0.67,\qquad
\mathrm{CoR}_{\text{wheel/terrain}} = 0.2.
\]
Here, \(\mu_{\text{wheel/terrain}}\) is the friction coefficient at the
interface, and \(\mathrm{CoR}_{\text{wheel/terrain}}\) dictates impact
dissipation between rigid and granular bodies.

\subsection{Gravitational and World Settings}

The simulation uses Earth gravity
\[
g = 9.8~\text{m/s}^2,
\]
where \(g\) denotes gravitational acceleration. The domain dimensions are
\[
L_x = 5~\text{m},\quad
L_y = 1~\text{m},\quad
L_z = 3~\text{m},
\]
with \(L_x, L_y, L_z\) representing the longitudinal, lateral, and vertical
extents of the granular bed. The inclined plane angle
\[
\theta = 24^\circ
\]
defines the slope on which tumbling occurs and determines the gravitational
component driving downhill motion.

\subsection*{Wheel (Tumbling Body) Properties}

The tumbling body is modeled as a rigid cylindrical wheel with
\[
r = 0.20~\text{m},\qquad
w = 0.12~\text{m},\qquad
m = 6.0~\text{kg},
\]
where \(r\) is radius, \(w\) is width, and \(m\) is total mass.

The principal moments of inertia are
\[
I_{yy} = 0.75\,m r^2,\qquad
I_{xx} = I_{zz} = m\left(0.4 r^2 + \frac{w^2}{12}\right),
\]
where \(I_{yy}\) corresponds to rotation about the axis of rolling, and
\(I_{xx}, I_{zz}\) correspond to lateral and vertical rotational axes. These
inertia values dictate angular acceleration, rotational stability, and momentum
exchange with the terrain.

\subsection{Particle-Based Terrain Settings}

The granular bed uses particle density
\[
\rho = 2600~\text{kg/m}^3,
\]
where \(\rho\) is the intrinsic mass density of sand-like grains. Before scaling,
each clump template has volume
\[
V = 4.2520508~\text{m}^3,
\]
with \(V\) denoting the multisphere aggregate volume. The scale factor
\[
s = 0.02
\]
uniformly shrinks particle dimensions so that actual DEM grains reach
millimeter-scale sizes. The triangular--flat multisphere clump shape introduces
angularity to model shear resistance, and \texttt{HCPSampler} initializes a
hexagonal close-packed distribution to minimize voids.

\subsection{Time Integration Settings}

The timestep used is
\[
h = 5\times10^{-6}~\text{s},
\]
where \(h\) ensures stability during stiff collisions. Velocity caps
\[
|v|_{\max} = 20~\text{m/s},\qquad
|v|_{\text{error}} = 35~\text{m/s},
\]
limit excessive particle acceleration; \( |v|_{\max} \) is the physical bound,
while \( |v|_{\text{error}} \) is a safety threshold for unstable events.
Collision detection updates occur every 40 solver iterations, and adaptive
frequency is disabled to ensure consistent numerical behavior.

\subsection{Wheel Actuation in Tumbling Mode}

The prescribed angular velocity is
\[
\omega_r = \frac{\pi}{4}~\text{rad/s},\qquad
v_\text{ref} = \omega_r r,
\]
where \(\omega_r\) sets the rolling rate and \(v_{\mathrm{ref}}\) is the
tangential speed at the wheel's rim. The wheel is constrained to translate only
along the \(y\)-axis and rotate only about the \(y\)-axis, ensuring motion
remains planar.

\subsection{Output Logging}

Simulation outputs are captured at
\[
10~\text{FPS},
\]
where FPS denotes frames per second. Logged data include particle positions
(\texttt{CSV}), contact mesh states (\texttt{VTK}), and wheel force and
kinematic histories. The simulation runs until
\[
t_\text{end} = 14~\text{s},
\]
where \(t_\text{end}\) specifies total simulated duration.

\bigskip
This parameter set provides the complete specification necessary to reproduce
the granular tumbling simulations and is included here because it characterizes
the extreme terrain--interaction regimes that complement the primary
sidewinding analysis presented in this thesis.

\section{Simulation Loop}
\label{sec:dem_sim_loop}

The DEM tumbling simulation follows a high-frequency update loop where particle
dynamics, collision resolution, and wheel motion are integrated in parallel on
the GPU. Algorithm~\ref{alg:dem_loop} summarizes the simulation workflow used
for COBRA rolling locomotion.

\begin{algorithm}[H]
\caption{DEM Simulation Loop (GPU-Accelerated)}
\label{alg:dem_loop}
\begin{algorithmic}[1]
\Require
    Clump templates, particle distribution, robot wheel geometry,\\
    soil material parameters $(k_n, k_t, \mu, d_n)$, gravity $g$,\\
    simulation time step $h$, total duration $T$
\Statex

\State Initialize Chrono DEM system and spatial hashing grid
\State Load granular bed (multi-sphere clumps) into domain
\State Assign wheel rigid-body properties and initial pose
\State Apply inclined boundary plane at slope $\theta$
\State Configure GPU contact kernels and logging channels

\For{$t = 0 \rightarrow T$ \textbf{with step} $h$}
    \State Update wheel kinematic constraint: $\omega_r \gets \pi/4$ rad/s
    \State Advance wheel motion DOF (translation constrained, rotation free)
    \State Kinematic thread (kT) updates geometry positions in GPU buffers

    \State \textbf{Dynamic thread (dT):}
    \Statex \hspace{0.5cm}Detect particle--wheel and particle--particle neighbors
    \Statex \hspace{0.5cm}Compute overlaps $\delta_n$ and shear increments $\dot{\boldsymbol{\xi}}$
    \Statex \hspace{0.5cm}Resolve normal/tangential forces via
        Eqs.~\eqref{eq:dem_normal_update}--\eqref{eq:dem_friction_constraint}
    \Statex \hspace{0.5cm}Accumulate per-body contact wrench on wheel

    \State Velocity and position update using explicit integration
    \State Enforce domain bounds and delete escaped particles

    \If{frame\_output triggered}
        \State Export particle states (\texttt{CSV}) and terrain mesh (\texttt{VTK})
        \State Log wheel COM trajectory, forces, and slip state
    \EndIf
\EndFor

\State Save final dataset and performance statistics
\end{algorithmic}
\end{algorithm}

\noindent The dual-thread GPU execution (kinematic vs. dynamic updates) ensures
efficient neighbor search and force computation even for large particle counts.
With a timestep of $h = 5 \times 10^{-6}$~s and optimized broad-phase culling,
the solver maintains stable dynamics under transient impact and deep soil
penetration conditions, which are characteristic of COBRA’s tumbling mode.

\section{Simulation Details}
\label{sec:dem_sim_details}

The DEM-based tumbling simulation is implemented in a modular Chrono setup that allows granular physics to interact with a simplified rigid-body representation of the robot in its circular tumbling configuration. The modular design keeps the system easy to configure, reproduce, and extend, while also ensuring that the granular terrain and the wheel-shaped robot body exchange forces in a physically meaningful way.

The core components of the simulation include the DEM physics engine, the rigid-body wheel model, the broad-phase collision system, and the logging interface. The DEM engine runs entirely on the GPU and computes particle interactions, collision forces, and momentum transfer at every timestep. The wheel model represents the closed-loop shape formed when the head and tail of the robot are connected, and it experiences the same soil resistance and impact forces as it would during real tumbling. A spatial hashing grid is used for broad-phase collision detection so that only nearby particles are considered for contact evaluation, which keeps computational cost manageable. The logging interface continuously records particle states, wheel kinematics, and force histories so that the resulting data can be analyzed afterward.

Several input assets are required to initialize the simulation. These include multi-sphere clump templates that approximate the shape and frictional behavior of granular particles, a terrain initialization routine that distributes particles using a hexagonal close-packing pattern, and wheel geometry with mass and inertia properties matching the tumbling configuration. Material parameters for both particles and the wheel reflect loose sand behavior and are selected to match real sinkage and shear responses. A control script defines the slope angle, the permitted degrees of freedom for the wheel, and the format of the output data.

The execution pipeline begins with the creation of the granular domain, assignment of gravity, and definition of the inclined plane. Motion constraints are then imposed on the wheel so that it rolls down the slope in a controlled manner. Once the solver begins running, particle contacts and wheel interactions are evaluated at every timestep using the GPU kernels. All relevant outputs, including particle rearrangement, wheel motion, and contact forces, are written to log files. After the simulation completes, the data are processed to extract sinkage trends, rolling distance, and other behavioral metrics.

The output data describe the evolution of wheel position, velocity, and orientation, along with detailed measurements of particle flow patterns and mound formation around the rolling body. The contact force history provides insight into how the wheel interacts with loose soil during each phase of the tumble, while solver performance statistics indicate how efficiently the GPU-based system handled the simulation.

These simulations illustrate the particle-scale mechanisms that govern tumbling on loose terrain. They complement the SCM-based analysis of sidewinding by providing a higher fidelity view of soil resistance, flow, and momentum exchange that cannot be captured with a continuum model. The combination of both approaches establishes a coherent and consistent understanding of how COBRA interacts with soft terrain across its different locomotion modes.

\chapter{Results}
\label{chap:results}

This chapter presents a comprehensive evaluation of the COBRA robot’s locomotion across two distinct gait modalities, namely sidewinding and tumbling. The analyses combine rigid-ground modeling, deformable-terrain simulations, and physical experiments conducted on real sand. Taken together, these studies illustrate how changes in terrain mechanics, from fully rigid support to compliant continuum deformation and finally to fully particulate flow, influence the robot’s ability to generate traction, maintain stability, and move effectively in unstructured environments.

\vspace{6pt}

The sidewinding results compare Simscape Multibody rigid-ground simulations, Project Chrono SCM deformable-terrain simulations, and hardware experiments on both laboratory flooring and loose sand. These comparisons highlight how compliance alters load transfer, contact persistence, and stride efficiency. The tumbling results combine Simscape rigid-ground models with high-fidelity Chrono DEM simulations, enabling detailed study of rolling stability, granular drag, and mass-wasting behavior on steep slopes. Together, these results span the transition from structured frictional contact to fully granular yielding, offering a unified perspective on COBRA’s locomotion mechanics.

The overarching objective of this chapter is to determine how different terrain models influence predicted kinematics, contact forces, and overall locomotion performance. By quantifying the divergence between rigid, continuum, and particle-based models—and validating each against experiment—this chapter establishes when simplified contact models suffice for predictive reasoning and when high-fidelity granular simulation becomes necessary.

\section{Qualitative Gait Replication in Simulation}

\begin{figure*}[h]
    \centering
    \includegraphics[width=0.9\textwidth]{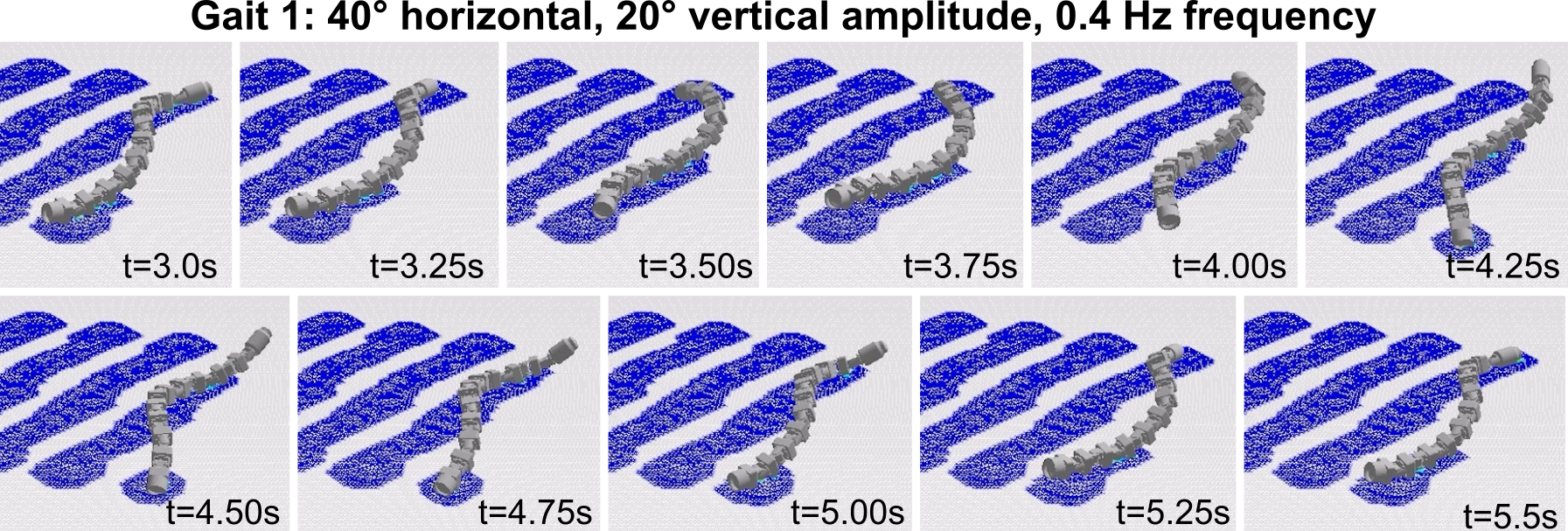}
    \caption{SCM simulation snapshots for Gait~1 (0.4~Hz). Blue shading highlights terrain depression caused by link loading. \textit{Image courtesy of Adarsh Salagame}.}
    \label{fig:chrono-gait1}
\end{figure*}

\begin{figure*}[h]
    \centering
    \includegraphics[width=0.9\textwidth]{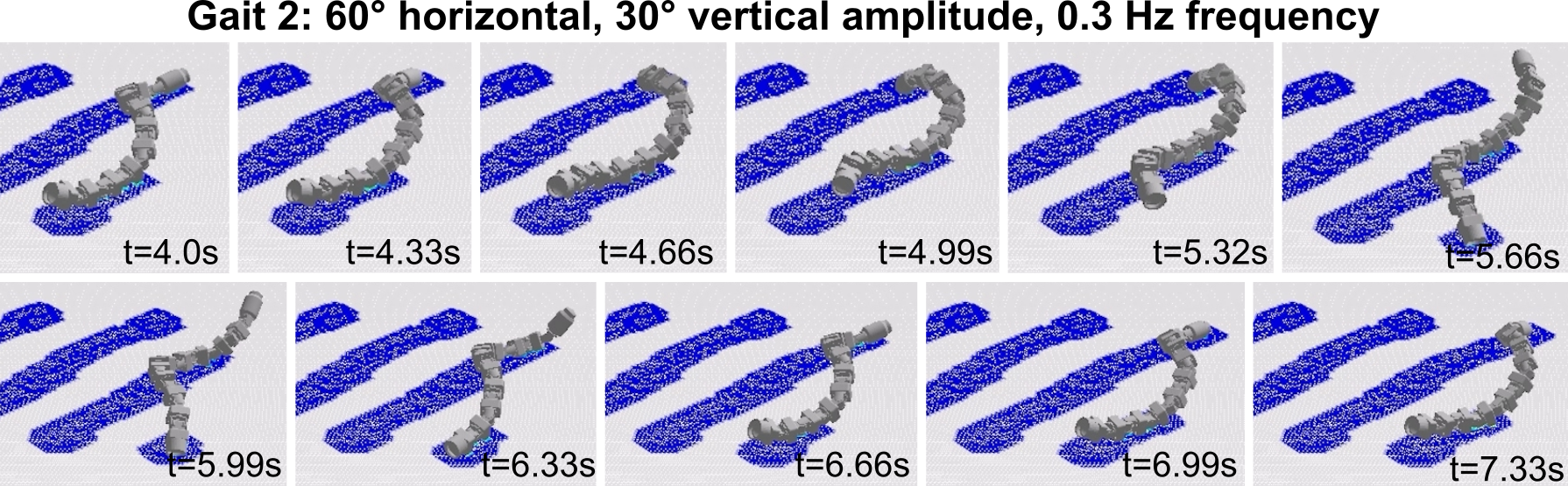}
    \caption{SCM simulation snapshots for Gait~2 (0.3~Hz). Increased amplitudes lead to greater sinkage and lateral soil displacement. \textit{Image courtesy of Adarsh Salagame}.}
    \label{fig:chrono-gait2}
\end{figure*}

Figures~\ref{fig:chrono-gait1} and~\ref{fig:chrono-gait2} illustrate sidewinding snapshots generated using Chrono SCM for two gait frequencies. In both cases, the traveling lateral wave is clearly reproduced, and the phase relationship between head and tail motion matches experimental observations. On deformable terrain, however, the robot’s links leave visible permanent depressions as they press into the soil. These depressions deepen when amplitudes increase and produce detectable lateral berms along the swept path. 

This phenomenon is entirely absent in rigid-terrain simulations, where ground reaction forces act instantaneously without modifying the terrain. The SCM results therefore capture a key physical mechanism present in real sand: compliant ground reduces the firmness of the anchoring points that sidewinding relies upon, leading to reduced forward progression per cycle.

Despite the presence of sinkage and drag, the SCM model faithfully reproduces the high-level kinematic structure of the gait. The head, mid-body, and tail segments maintain the alternating tripod-like contact structure characteristic of efficient sidewinding. This indicates that although terrain compliance changes local force magnitudes, the global kinematic template remains stable, demonstrating that the gait is robust to moderate substrate deformability.

\FloatBarrier      
\section{Contact Mechanics and Load Redistribution}
\FloatBarrier      

\begin{figure}[ht]
    \centering
    \includegraphics[width=0.70\linewidth]{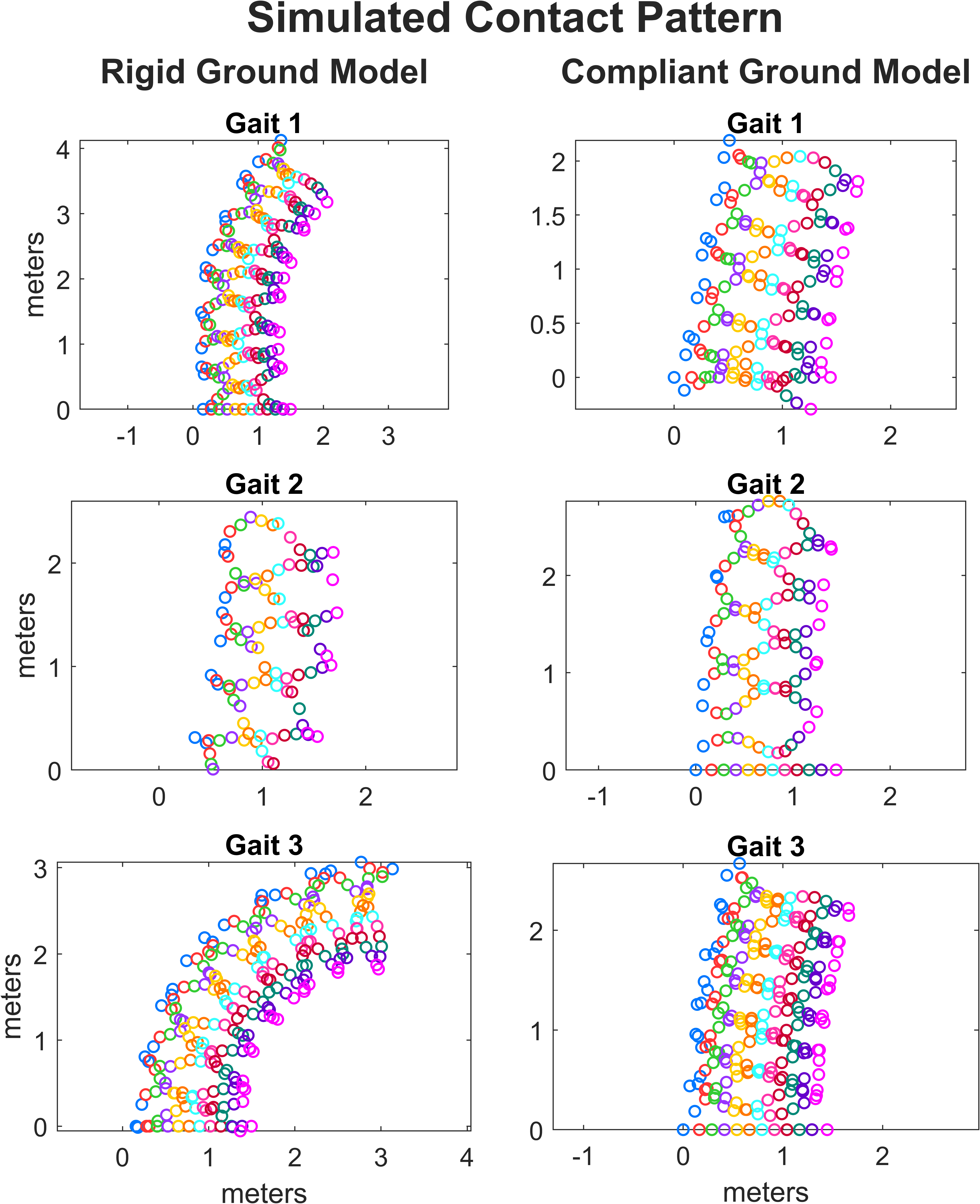}
    \caption{Contact distribution over one gait cycle. Deformable terrain results in wider and more diffused loading regions.}
    \label{fig:contact}
\end{figure}

The contact-location comparison in Fig.~\ref{fig:contact} reveals a notable contrast between rigid and deformable terrain. On rigid ground, contact points are discrete and localized. Each link establishes a short-duration, high-stiffness contact that cleanly transfers load to the surface. In contrast, SCM produces spatially broader and temporally longer loading patches. As the body sinks, neighboring nodes share the load, distributing contact over a larger region. This diffused interaction reduces the sharp force spikes observed in rigid-ground simulations and experiments on hard surfaces.

\FloatBarrier  
\begin{figure}[h]
    \centering
    \includegraphics[width=0.70\linewidth]{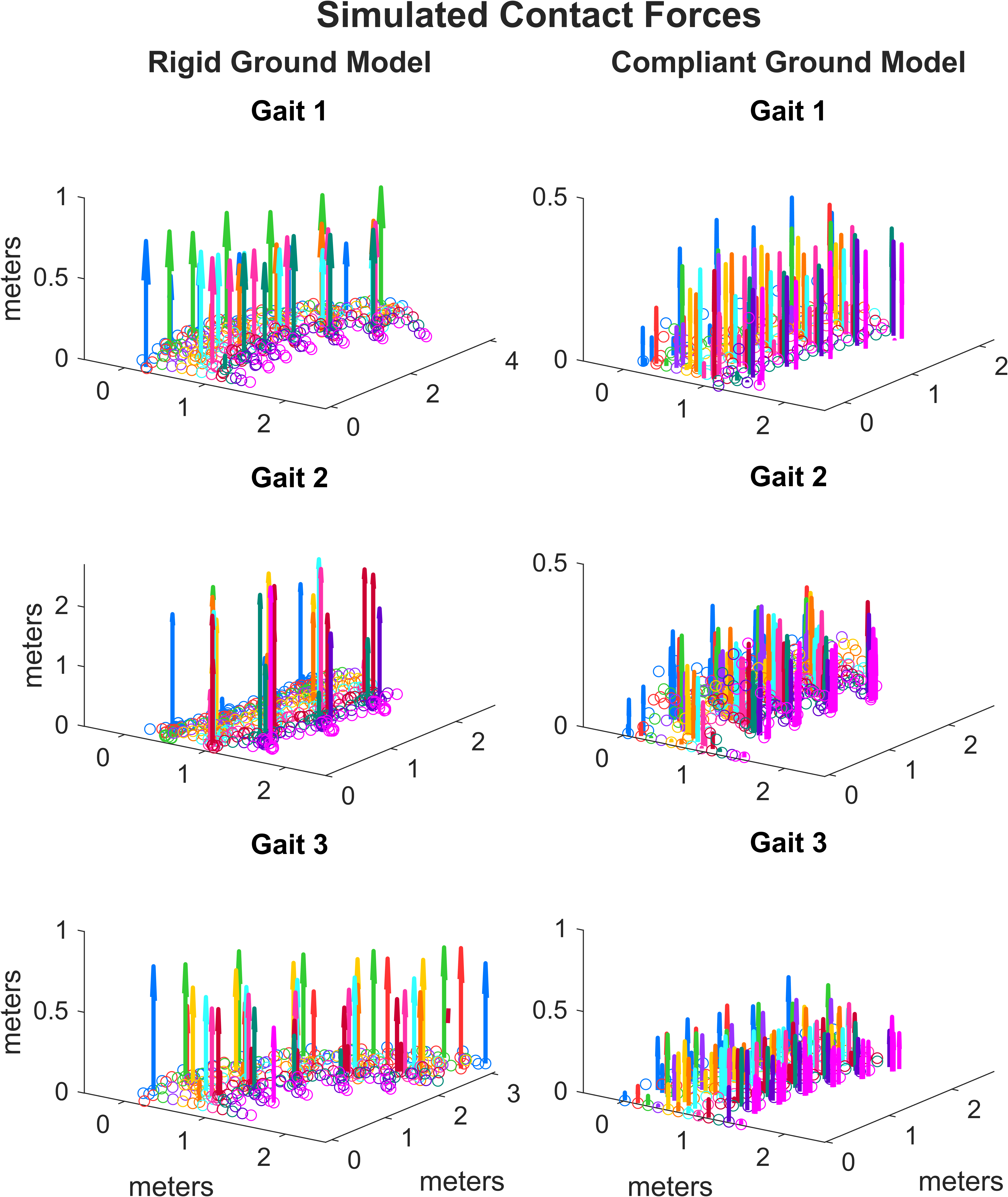}
    \caption{Comparison of ground reaction forces. Deformable terrain decreases peak forces and spreads loading temporally.}
    \label{fig:grf}
\end{figure}

Normal force profiles in Fig.~\ref{fig:grf} further highlight this behavior. While all models exhibit periodic force oscillations aligned with the gait cycle, deformable terrain consistently lowers peak force magnitudes. This reduction arises because the soil deforms plastically under compression, effectively attenuating high-frequency impacts and providing a cushioned interaction. Hardware trials on loose sand show similar signatures, validating the SCM predictions.

\FloatBarrier
\begin{figure*}[ht]
    \centering
    \includegraphics[width=1\textwidth]{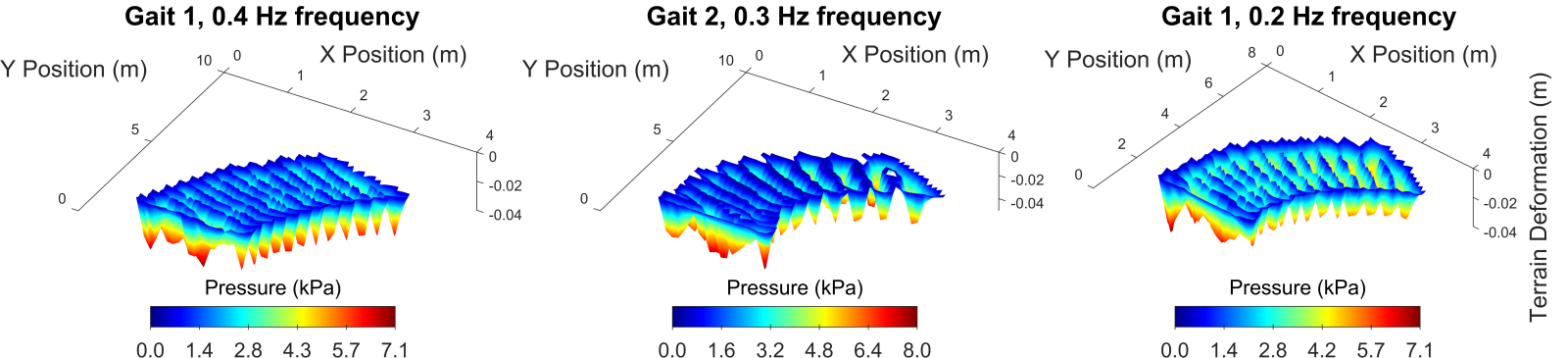}
    \caption{Contact pressure and permanent terrain depression over a gait cycle in Chrono SCM. Node-level sinking and smoothing are visible.}
    \label{fig:pressure}
\end{figure*}

The pressure-field visualization in Fig.~\ref{fig:pressure} gives additional insight into how local sinkage redistributes loading. Regions of high pressure cluster near inflection points in the body wave, where curvature is greatest and load transfer is most pronounced. Over time, these regions blend into smooth depressions due to plastic soil yielding. This permanent deformation mirrors experimental footage in which the robot leaves a characteristic sinusoidal trench along its path. The simulation therefore captures the essential terramechanics of sidewinding: contact patches expand, anchor quality decreases, and forward displacement per cycle diminishes.

\FloatBarrier

\section{Head Trajectory Comparisons}

\begin{figure}[ht]
    \centering
    \includegraphics[width=0.75\linewidth]{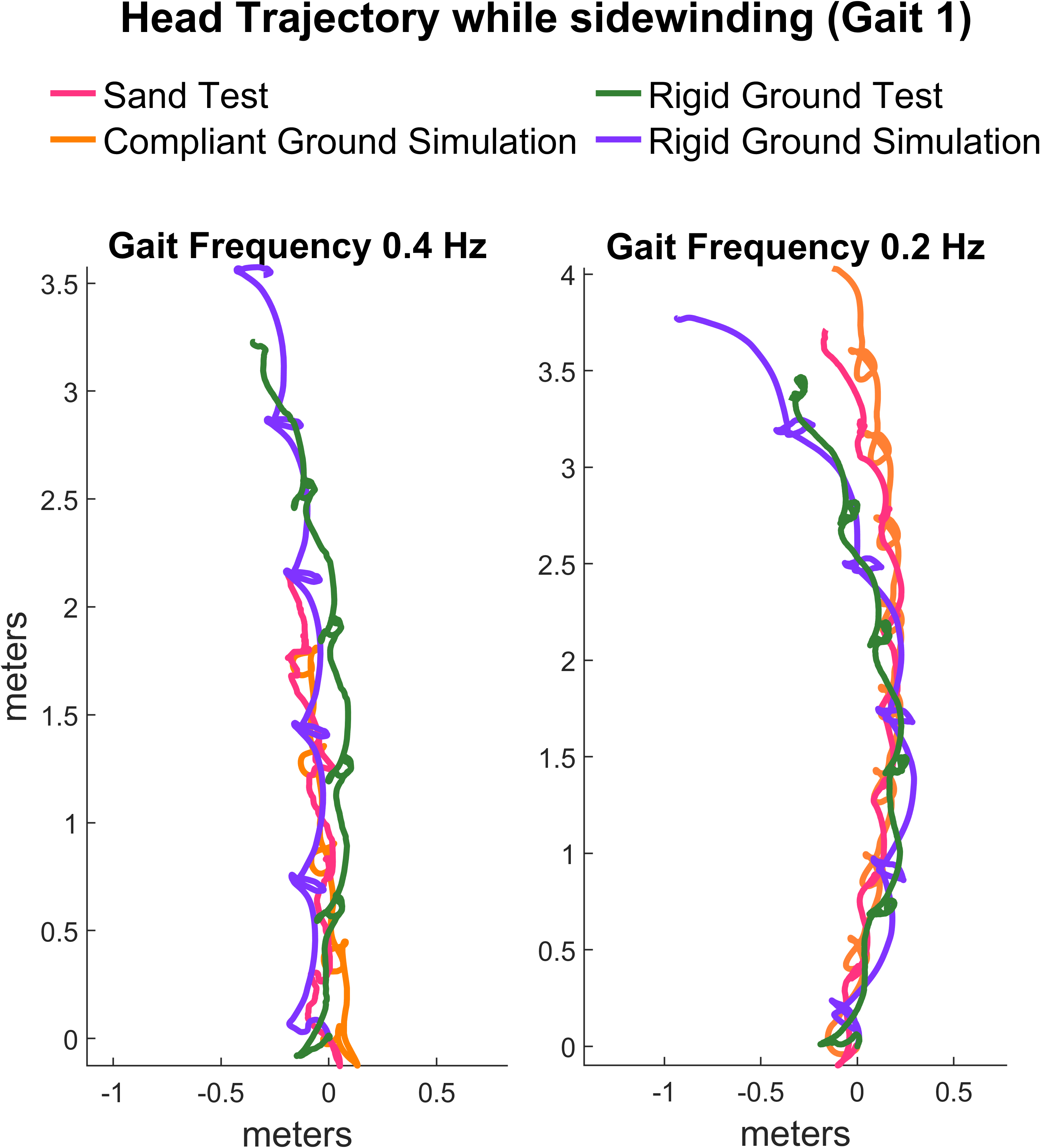}
    \caption{Comparison of head trajectories for Gait~1. Rigid-ground simulations closely match rigid-ground experiments; SCM captures compliant-ground trends.}
    \label{fig:head_traj}
\end{figure}

\begin{figure}[ht]
    \centering
    \includegraphics[width=0.9\linewidth]{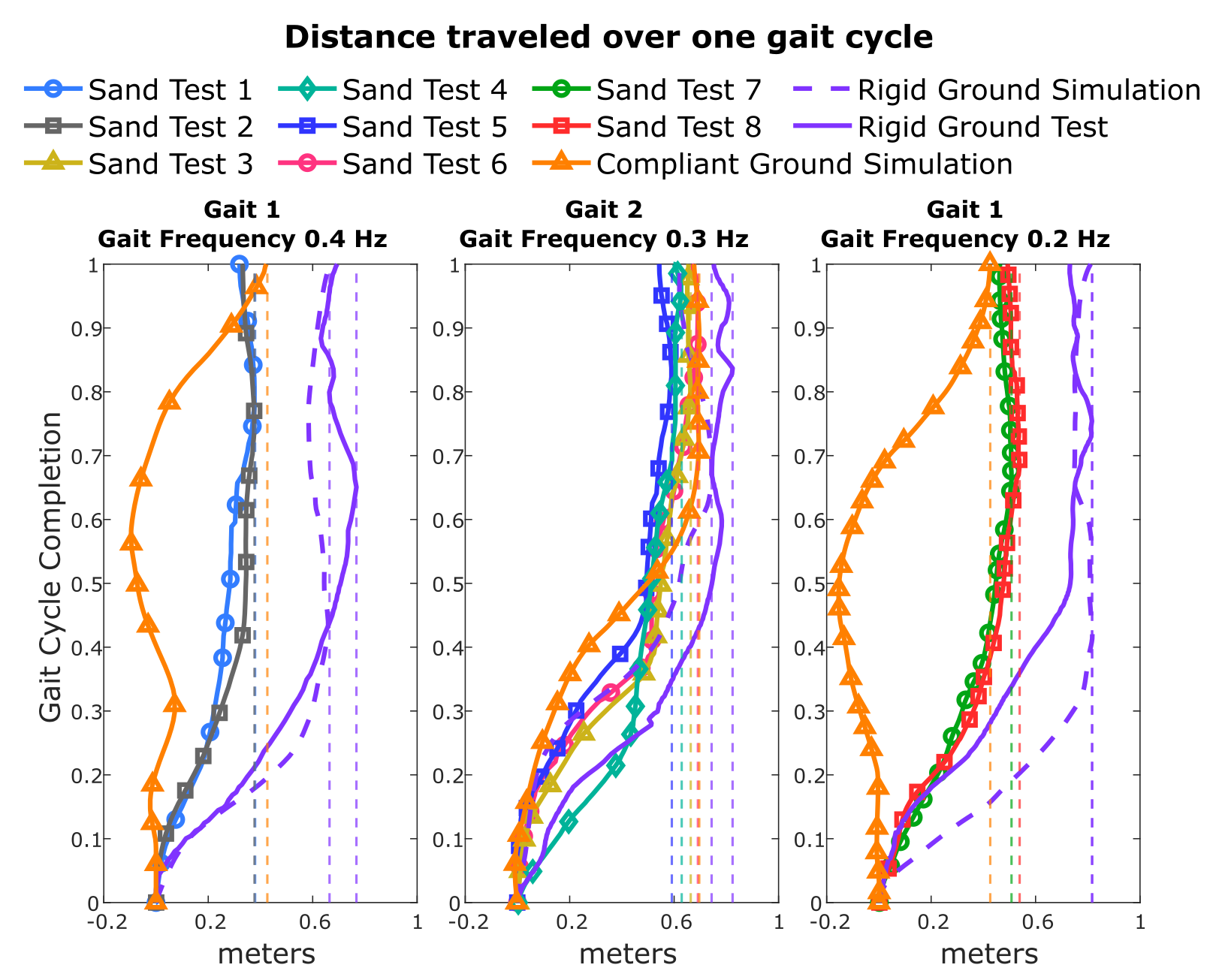}
    \caption{Normalized head trajectories over a gait cycle showing similar stationary and sweeping phases across all environments.}
    \label{fig:head_traj_norm}
\end{figure}

Figures~\ref{fig:head_traj} and~\ref{fig:head_traj_norm} compare experimental and simulated head trajectories, revealing how terrain compliance alters both the magnitude and temporal structure of motion. The rigid-ground simulation closely matches laboratory rigid-surface experiments, demonstrating that frictional interaction alone is sufficient to predict high-level kinematics on hard floors. In contrast, SCM and real sand data show shorter forward displacement per cycle due to energy lost to soil deformation.

The normalized trajectories highlight that all cases preserve the alternating sweep-and-pause behavior intrinsic to sidewinding. However, compliant terrain introduces slight temporal delays in the sweeping phase as the body drags through displaced material. Experimental results show small deviations caused by reduced VIO reliability at low frequencies; nonetheless, the overall motion pattern remains consistent across environments. These comparisons establish that while terrain compliance alters locomotor efficiency, it does not destabilize the fundamental geometry of the gait.

\section{Summary of Findings}

The collective evidence from all simulations and experiments demonstrates that sidewinding is highly robust to substrate compliance. Directionality remains stable, the gait structure is preserved, and contact patterns follow the expected multi-anchor locomotion template. However, deformable terrain consistently reduces stride length due to slip, sinkage, and energy lost to soil rearrangement. These findings justify a hierarchical modeling strategy: rigid-ground simplifications provide fast and reliable predictions for short-horizon control, while compliant models such as SCM are necessary when efficiency, traction limits, or long-duration behaviors are of interest.

\FloatBarrier
\section{Tumbling Gait Results: Chrono DEM Engine}
\label{sec:dem_results}

\begin{figure}[ht]
    \centering
    \includegraphics[width=0.95\linewidth]{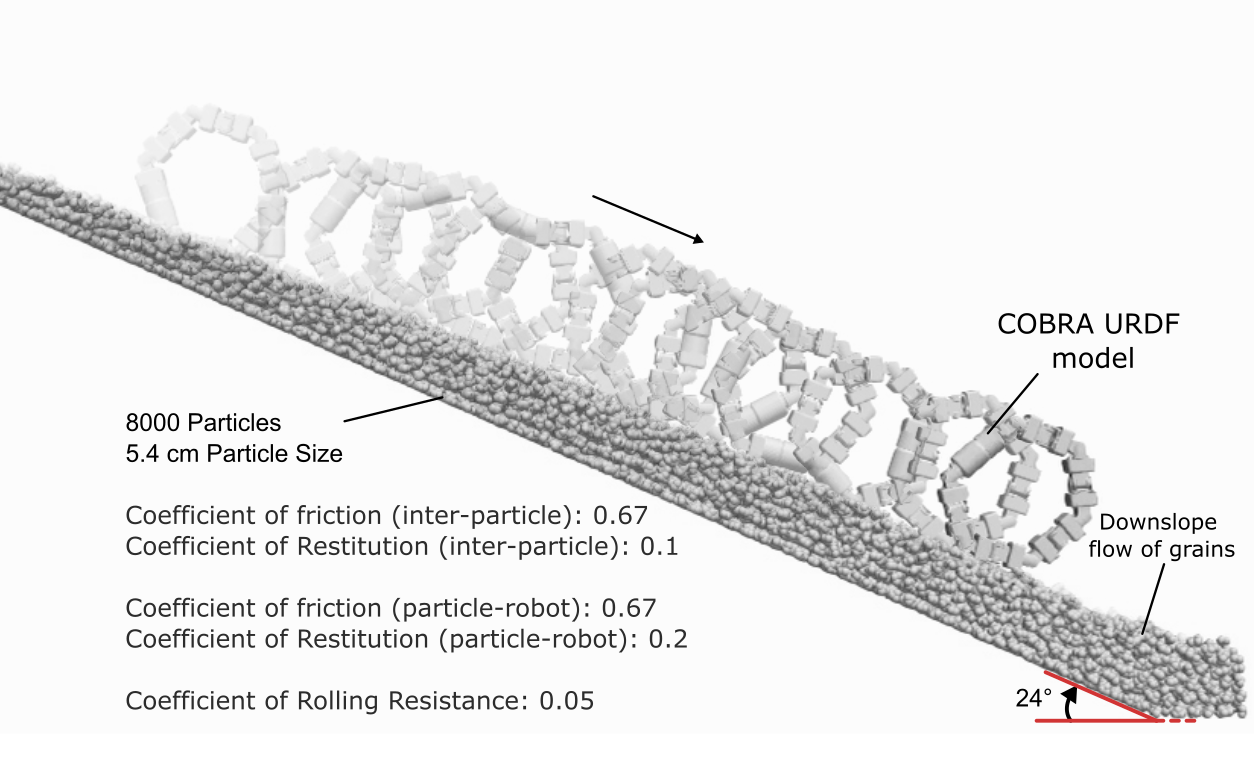}
    \caption{Chrono DEM simulation of COBRA tumbling downslope on loose granular
    terrain ($24^\circ$ inclination). Terrain deformation, mass–wasting, and
    body sinkage are visible.\textit{Image courtesy of Adarsh Salagame}. \cite{salagame_crater_2025}
    \label{fig:dem_snapshot}}
\end{figure}

The tumbling analysis examines COBRA’s behavior on steep granular slopes, where rolling replaces sidewinding as the primary locomotion strategy. The DEM simulation in Fig.~\ref{fig:dem_snapshot} reveals the complex particle–robot interactions that arise during this high-energy mode. As the robot descends a $24^\circ$ slope, grains accumulate at the front surface, slide beneath the body, and disperse downslope as the shell rotates. These flows are hallmarks of mass-wasting phenomena observed in natural granular systems. The robot experiences partial burial during some phases, and local hills of displaced material form around its trajectory—behaviors impossible to capture with rigid or continuum soil models.

\begin{figure*}[htbp]
    \centering
    \includegraphics[width=0.95\textwidth]{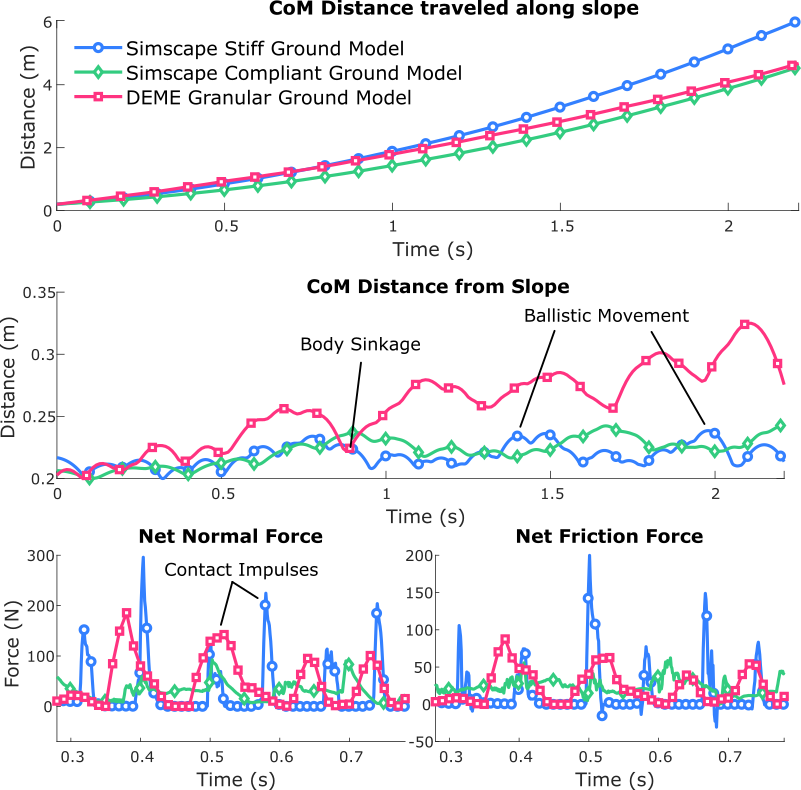}
    \caption{Comparison of tumbling dynamics across three models:
    Simscape stiff ground (blue), Simscape compliant ground (green), and Chrono
    DEM granular terrain (magenta). DEM results replicate body sinkage,
    dissipative energy loss, and reduced impact impulses observed in real sand.\textit{Image courtesy of Adarsh Salagame}. \cite{salagame_crater_2025}}
    \label{fig:dem_output_plots}
\end{figure*}

The quantitative comparison in Fig.~\ref{fig:dem_output_plots} contrasts DEM with two Simscape models: a stiff rigid-ground case and a compliant-surface approximation. The rigid model predicts the fastest descent because no energy is dissipated into the substrate. In contrast, the DEM simulation exhibits slower progression due to grain collision losses, slip at the contact interface, and drag arising from particle rearrangement. The compliant Simscape case is slowest because its damping forces are artificially large compared to true granular resistance, causing significant energy absorption that does not fully resemble physical sand behavior.

Vertical CoM motion offers further insight. In the DEM simulation, the CoM trajectory oscillates as the robot intermittently loses support while riding over sparse particle contact patches. These short ballistic phases correspond to airborne micro-transitions caused by local voids and grain slippage. Rigid models cannot replicate these effects because their support surface never fails under load. Compliant Simscape exhibits vertical oscillations but lacks the sharp discontinuities characteristic of true granular yielding.

Normal and friction forces reinforce this interpretation. On stiff ground, impacts generate very high-magnitude impulse spikes because collisions are perfectly localized and instantaneous. DEM attenuates these forces substantially, as grains absorb impact momentum by deforming, sliding, and scattering. The compliant Simscape model dampens impacts even further but does so through a purely numerical mechanism, rather than physically meaningful grain-scale processes. DEM therefore provides a better qualitative and quantitative representation of the reaction loads expected during real rolling on sand.

Overall, the DEM results demonstrate that granular yielding plays a dominant role in moderating downhill acceleration and shaping the robot’s rolling trajectory. The combination of reduced peak forces, intermittent lift, and drag-induced deceleration yields a physically realistic representation of tumbling mechanics on loose sand. Compared to Simscape models, DEM provides the only framework capable of capturing mass-wasting, energy dissipation through particle flow, and the emergent rheological behavior of real granular slopes.

\FloatBarrier

 \chapter{Conclusion}
\label{chap:conclusion}

This thesis presented a unified modeling and simulation framework for analyzing
the locomotion of the \ac{COBRA} snake robot across rigid, compliant, and
granular terrains. Two modes of mobility were studied: sidewinding for efficient
planar traversal and tumbling for rapid descent on steep slopes. Together, these
analyses demonstrated how terrain mechanics influence stability, slip,
propulsion efficiency, and net displacement in contact-rich locomotion.

For sidewinding locomotion, MATLAB Simscape simulations on rigid ground
accurately reproduced the primary kinematic patterns and directional movement of
the robot. These results highlight that rigid-ground models are often sufficient
for predictive control, where the objective is to estimate the net direction and
magnitude of motion over short horizons. In contrast, simulations and
experiments on soft sand showed that terrain yielding introduces slip and
sinkage, reducing displacement per cycle. Project Chrono SCM simulations
captured these effects by modeling normal and shear soil deformation, providing
a closer representation of locomotion behavior on deformable ground.

In the high-energy tumbling locomotion mode, impacts and bulk soil flow become
the dominant contributors to motion. While a rigid-ground Simulink model
provides a clean baseline for motion prediction, the Chrono \ac{DEM} engine
revealed transient lift-off, grain rearrangements, and mass-wasting phenomena
that strongly influence the descent rate and contact forces. These dynamics
demonstrate that particle-level physics are required for accurate locomotion
analysis during extreme terrain interactions and deep penetration events.

Across both locomotion modes, this work shows that the level of terrain fidelity
required for simulation depends on the underlying mobility strategy:

\begin{itemize}
    \item \textbf{Rigid-ground models} remain well-suited for real-time planning
    and control design due to their computational efficiency and strong
    directional accuracy.
    \item \textbf{Continuum soil models} such as SCM enable correct prediction
    of slip, sinkage, and load redistribution effects that drive changes in
    locomotion efficiency.
    \item \textbf{DISCRETE particle-based models} such as DEM are necessary for
    high-energy and worst-case interactions where soil failure and bulk
    displacement dominate.
\end{itemize}

This hierarchical approach to locomotion modeling establishes a complete
analysis pipeline that spans real-time motion prediction to high-fidelity
terrain mechanics, forming a robust basis for future field deployment in
uncertain environments.

\subsection{Future Work}

The research presented in this thesis opens multiple promising directions:

\begin{itemize}
    \item \textbf{Closed-loop control on deformable terrain:}
    Integrating compliant-ground modeling into predictive controllers to
    improve slip compensation and stability.

    \item \textbf{Online terrain parameter estimation:}
    Using sensing and contact feedback to automatically adapt soil models for
    different substrates, improving robustness to environmental variability.

    \item \textbf{Multi-modal locomotion transitions:}
    Developing autonomous strategies for switching between sidewinding,
    tumbling, and other locomotion modes based on situational awareness.

    \item \textbf{GPU-accelerated granular simulation at larger scales:}
    Improving DEM resolution for more accurate representation of grain-scale
    failure and deep submersion events.

    \item \textbf{Field testing in planetary-analogue environments:}
    Extending validation beyond laboratory settings to steep slopes and
    crater-like sites with mixed and loose soil conditions.
\end{itemize}

\subsection{Final Remarks}

Through detailed modeling, rigorous simulation, and hardware validation, this
thesis advances understanding of how a snake robot generates motion under
varying terrain mechanics. The contributions support the long-term goal of
deploying resilient, adaptive robotic systems capable of navigating challenging,
unstructured environments where traditional mobility platforms fail. This work
lays the foundation for reliable and versatile locomotion in robotics missions
ranging from extraterrestrial exploration to search and rescue in
disaster-stricken areas.

\printbibliography



\appendix


\end{document}